\documentclass[twoside,11pt]{article}

\PassOptionsToPackage{hyperfootnotes=false}{hyperref}
\usepackage{subcaption}
\usepackage[abbrvbib, preprint]{jmlr2e}
\usepackage[utf8]{inputenc}
\usepackage[T1]{fontenc}
\usepackage{amsmath,amsfonts}
\allowdisplaybreaks
\usepackage{aliascnt}
\usepackage{array}
\usepackage{textcomp}
\usepackage{verbatim}

\usepackage{amssymb}
\usepackage{mathtools}
\usepackage{amsthm}
\usepackage{mathrsfs}
\usepackage{bbm}
\usepackage{tasks}
\usepackage{wrapfig}
\usepackage{placeins}
\usepackage[capitalise]{cleveref}
\usepackage{enumitem}
\usepackage{booktabs}
\usepackage{xurl}
\usepackage{pifont}% http://ctan.org/pkg/pifont
\usepackage{makecell}
\usepackage{multicol}
\usepackage{tikz-cd}
\usepackage{units}
\usepackage[ruled]{algorithm2e}
\usepackage{titletoc}
\usepackage{multirow}
\usepackage[dvipsnames]{xcolor}
\usepackage{colortbl}
\usepackage{diagbox}
\usepackage{crossreftools}
\usepackage{listings}
\usepackage{needspace}

\newcommand{\BlackBox}{\rule{1.5ex}{1.5ex}}
\ifdefined\proof
    \renewenvironment{proof}{\par\noindent{\bf Proof\ }}{\hfill\BlackBox\\[2mm]}
\else
    \newenvironment{proof}{\par\noindent{\bf Proof\ }}{\hfill\BlackBox\\[2mm]}
\fi

\theoremstyle{plain}

\theoremstyle{definition}

\theoremstyle{remark}

\usepackage[most]{tcolorbox}

\definecolor{thesisParisBlue}{HTML}{3D91C6}
\definecolor{thesisParisRed}{HTML}{E47B72}
\definecolor{thesisParisGreen}{HTML}{16805D}
\definecolor{thesisParisGold}{HTML}{A88A45}

\definecolor{thesisParisBrown}{RGB}{190,148,110} % Light walnut.
\definecolor{thesisParisBackground}{RGB}{247,249,251} % Current cool blue-gray.
\newtcolorbox{parismath}[1][thesisParisBlue]{
  enhanced jigsaw,
  breakable,
  colback=thesisParisBackground,
  colframe=black!18,
  boxrule=0.35pt,
  sharp corners,
  left=8pt,
  right=8pt,
  top=6pt,
  bottom=6pt,
  before skip=6pt,
  after skip=6pt,
  drop fuzzy shadow southeast={black!22},
  overlay unbroken and first={%
    \draw[color=#1,line width=2.2pt]
      (frame.north west) -- ([xshift=2.4cm]frame.north west);%
  }
}

\newcommand{\thesisParisBeginTheorem}[3]{%
  \begin{parismath}[#1]%
  \IfNoValueTF{#3}{\begin{#2}}{\begin{#2}[#3]}%
}

\NewDocumentEnvironment{parisdefinition}{o}
  {\thesisParisBeginTheorem{thesisParisGreen}{definition}{#1}}
  {\end{definition}\end{parismath}}

\NewDocumentEnvironment{parisremark}{o}
  {\thesisParisBeginTheorem{thesisParisRed}{remark}{#1}}
  {\end{remark}\end{parismath}}

\NewDocumentEnvironment{paristheorem}{o}
  {\thesisParisBeginTheorem{thesisParisBlue}{theorem}{#1}}
  {\end{theorem}\end{parismath}}

\NewDocumentEnvironment{paristhm}{o}
  {\thesisParisBeginTheorem{thesisParisBlue}{thm}{#1}}
  {\end{thm}\end{parismath}}

\NewDocumentEnvironment{parisproposition}{o}
  {\thesisParisBeginTheorem{thesisParisBlue}{proposition}{#1}}
  {\end{proposition}\end{parismath}}

\NewDocumentEnvironment{parisprop}{o}
  {\thesisParisBeginTheorem{thesisParisBlue}{prop}{#1}}
  {\end{prop}\end{parismath}}

\NewDocumentEnvironment{parisduplicate}{o}
  {\thesisParisBeginTheorem{thesisParisBlue}{duplicate}{#1}}
  {\end{duplicate}\end{parismath}}

\NewDocumentEnvironment{parislemma}{o}
  {\thesisParisBeginTheorem{thesisParisBlue}{lemma}{#1}}
  {\end{lemma}\end{parismath}}

\NewDocumentEnvironment{parisdup}{o}
  {\thesisParisBeginTheorem{thesisParisBlue}{dup}{#1}}
  {\end{dup}\end{parismath}}

\NewDocumentEnvironment{pariscorollary}{o}
  {\thesisParisBeginTheorem{thesisParisBlue}{corollary}{#1}}
  {\end{corollary}\end{parismath}}

\NewDocumentEnvironment{parisassumption}{o}
  {\thesisParisBeginTheorem{thesisParisBlue}{assumption}{#1}}
  {\end{assumption}\end{parismath}}

\NewDocumentEnvironment{parisexample}{o}
  {\thesisParisBeginTheorem{thesisParisBrown}{example}{#1}}
  {\end{example}\end{parismath}}

\providecommand{\calM}{\mathcal{M}}
\providecommand{\calN}{\mathcal{N}}

\providecommand{\calY}{\mathcal{Y}}

\providecommand{\bbK}{\mathbb{K}}
\providecommand{\bbD}{\mathbb{D}}
\providecommand{\bbL}{\mathbb{L}}

\providecommand{\bbX}{\mathbb{X}}
\providecommand{\bbY}{\mathbb{Y}}
\providecommand{\sym}[1]{\mathcal{S}^{#1}}
\providecommand{\spd}[1]{\mathcal{S}^{#1}_{++}}
\providecommand{\bbR}[1]{\mathbb {R}^{#1}}

\providecommand{\bbRscalar}{\mathbb {R}}

\providecommand{\orth}[1]{\mathrm{O}({#1})}
\providecommand{\bfst}{\mathbf{ST}}

\providecommand{\dist}{\operatorname{d}}
\providecommand{\rieexp}{\operatorname{Exp}}
\providecommand{\rielog}{\operatorname{Log}}
\providecommand{\diff}{\operatorname{d}}

\providecommand{\pt}[2]{\operatorname{PT}_{#1 \rightarrow #2}}

\providecommand{\bbD}{\mathbb {D}}

\providecommand{\diag}{\operatorname{diag}}

\providecommand{\diff}{\operatorname{d}}
\providecommand{\tr}{\operatorname{tr}}

\providecommand{\geuc}{g^{\mathrm{E}}}

\providecommand{\fnorm}[1]{\left\|{#1} \right\|_\mathrm{F}}
\providecommand{\tr}{\operatorname{tr}}
\providecommand{\chol}{\operatorname{Chol}}

\providecommand{\dlog}{\operatorname{Dlog}}
\providecommand{\pow}{\operatorname{P}}
\providecommand{\inner}[2]{\left\langle #1, #2\right \rangle}

\providecommand{\rmF}{\mathrm{F}}

\providecommand{\clog}{\psi_{\mathrm{LC}}}
\providecommand{\fm}{\operatorname{FM}}
\providecommand{\wfm}{\operatorname{WFM}}
\providecommand{\argmin}{\operatorname{argmin}}
\providecommand{\argmax}{\operatorname{argmax}}
\providecommand{\liebn}{\operatorname{LieBN}}
\providecommand{\mlog}{\operatorname{log}}
\providecommand{\mexp}{\operatorname{exp}}
\providecommand{\spdtrans}[2]{\Gamma_{#1 \rightarrow #2}}

\providecommand{\odotai}{\odot^{\mathrm{AI}}}
\providecommand{\odotle}{\odot^{\mathrm{LE}}}
\providecommand{\odotlc}{\odot^{\mathrm{LC}}}
\providecommand{\odotpai}{\odot^{\theta\text{-AI}}}

\providecommand{\odotplc}{\odot^{\theta\text{-LC}}}

\providecommand{\alphabeta}{(\alpha,\beta)}

\providecommand{\biparamAIM}{(\alpha,\beta)\text{-AIM}}
\providecommand{\biparamLEM}{(\alpha,\beta)\text{-LEM}}
\providecommand{\triparamLEM}{(\theta,\alpha,\beta)\text{-LEM}}
\providecommand{\triparamAIM}{(\theta,\alpha,\beta)\text{-AIM}}
\providecommand{\gtriparamAI}{g^{(\theta,\alpha,\beta)\text{-AI}}}

\providecommand{\paramLCM}{\theta\text{-LCM}}

\providecommand{\triparamLEM}{(\theta,\alpha,\beta)\text{-LEM}}
\providecommand{\gbiparamLE}{g^{(\alpha,\beta)\text{-LE}}}
\providecommand{\gtriparamLE}{g^{(\theta,\alpha,\beta)\text{-LE}}}

\providecommand{\gbiparamai}{g^{(\alpha,\beta)\text{-AI}}}

\providecommand{\mle}{\mathrm{MLE}}

\providecommand{\calB}{\mathcal{B}}

\providecommand{\so}[1]{\mathrm{SO}(#1)}
\providecommand{\soprod}[2]{\mathrm{SO}^{#1}(#2)}
\providecommand{\soLieAlgebra}[1]{\mathfrak{so}(#1)}

\providecommand{\gleft}{g^{\mathrm{L}}}
\providecommand{\gright}{g^{\mathrm{R}}}
\providecommand{\ltrans}{\operatorname{L}}
\providecommand{\rtrans}{\operatorname{R}}
\providecommand{\inv}[1]{{#1}^{-1}_{\odot}}

\providecommand{\gcri}{g^{\mathrm{CRI}}}
\providecommand{\gdefcri}{g^{\theta\text{-CRI}}}
\providecommand{\symmetrize}[1]{\left(#1 \right)_{\mathrm{Sym}}}

\providecommand{\mrL}{\mathrm{L}}

\providecommand{\bfzero}{\mathbf{0}}
\providecommand{\cor}[1]{\mathcal{C}^{#1}_{++}}
\providecommand{\coropt}{\operatorname{Cor}}
\providecommand{\hol}[1]{\mathrm{Hol}^{#1}}
\providecommand{\chospace}[1]{\mathrm{LT}_{++}^{#1}}
\providecommand{\trilspace}[1]{\mathrm{LT}^{#1}}

\providecommand{\LTone}[1]{\mathrm{LT}_1^{#1}}
\providecommand{\LTzero}[1]{\mathrm{LT}_0^{#1}}

\providecommand{\bbDspace}[1]{\mathbb{D}^{#1}}
\providecommand{\bbDplus}[1]{\mathbb{D}_{+}^{#1}}
\providecommand{\covtocor}{\operatorname{Cor}}

\providecommand{\diag}{\operatorname{diag}}
\providecommand{\vecone}{\boldsymbol{1}}

\providecommand{\singperm}[1]{\mathfrak{S}^{\pm}(n)}
\providecommand{\perm}[1]{\mathfrak{S}^{#1}}

\providecommand{\off}{\mathrm{Off}}
\providecommand{\holinner}[2]{\left \langle #1, #2 \right \rangle^{(\alpha,\beta,\gamma)}}
\providecommand{\holnorm}[1]{\left \| #1 \right \|^{(\alpha,\beta,\gamma)}}
\providecommand{\offlog}{\operatorname{Log}^{\circ}}
\providecommand{\offexp}{\operatorname{Exp}^{\circ}}
\providecommand{\Sum}{\operatorname{Sum}}

\providecommand{\rzero}[1]{\mathrm{Row}_0^{#1}}
\providecommand{\rone}[1]{\mathrm{Row}_1 ^{#1}}
\providecommand{\dstar}{\mathcal{D}^\star}
\providecommand{\dplus}{\mathcal{D}^+}

\providecommand{\rzeroinner}[2]{\left \langle #1, #2 \right \rangle^{(\alpha,\delta,\zeta)}}
\providecommand{\rzeronorm}[1]{\left \| #1 \right \|^{(\alpha,\delta,\zeta)}}
\providecommand{\logscaled}{\operatorname{Log}^{\star}}
\providecommand{\expscaled}{\operatorname{Exp}^{\star}}

\providecommand{\pball}[1]{\mathbb{P}^{#1}}

\providecommand{\hs}[1]{\mathrm{H}\mathbb{S}^{#1}}

\newcommand{\cmark}{\textcolor{green}{\ding{51}}}
\newcommand{\xmark}{\textcolor{red}{\ding{55}}}
\newcommand{\na}{\textcolor{gray}{N/A}}%

\providecommand{\ie}{\emph{i.e.}}

\providecommand{\mypara}[1]{\textit{#1}}

\definecolor{HilightColor}{RGB}{240, 235, 255}  % 更淡紫色

\crefname{equation}{Equation}{Equations}
\Crefname{equation}{Equation}{Equations}
\crefname{figure}{Figure}{Figures}
\Crefname{figure}{Figure}{Figures}
\crefname{table}{Table}{Tables}
\Crefname{table}{Table}{Tables}
\crefname{algocf}{Algorithm}{Algorithms}
\Crefname{algocf}{Algorithm}{Algorithms}
\crefname{section}{Section}{Sections}
\Crefname{section}{Section}{Sections}
\crefname{appendix}{Appendix}{Appendices}
\Crefname{appendix}{Appendix}{Appendices}

\crefname{theorem}{Theorem}{Theorems}
\Crefname{theorem}{Theorem}{Theorems}
\crefname{lemma}{Lemma}{Lemmas}
\Crefname{lemma}{Lemma}{Lemmas}
\crefname{definition}{Definition}{Definitions}
\Crefname{definition}{Definition}{Definitions}
\crefname{corollary}{Corollary}{Corollaries}
\Crefname{corollary}{Corollary}{Corollaries}
\crefname{remark}{Remark}{Remarks}
\Crefname{remark}{Remark}{Remarks}
\crefname{proposition}{Proposition}{Propositions}
\Crefname{proposition}{Proposition}{Propositions}
\crefname{proof}{Proof}{Proofs}
\Crefname{proof}{Proof}{Proofs}
\crefname{assumption}{Assumption}{Assumptions}
\Crefname{assumption}{Assumption}{Assumptions}
\crefname{enumi}{Property}{Properties}
\Crefname{enumi}{Property}{Properties}

\input{link_proof}
\usepackage{lastpage}

\jmlrheading{XX}{2026}{1--\pageref{LastPage}}{7/26}{}{XX-XXXX}{Ziheng Chen, Yue Song, Rui Wang, Xiao-Jun Wu, and Nicu Sebe}
\ShortHeadings{LieBN: Batch Normalization Over Lie Groups}{Chen, Song, Wang, Wu, and Sebe}
\firstpageno{1}

\begin{document}

\title{LieBN: Batch Normalization Over Lie Groups}

\author{%
\name Ziheng Chen \email ziheng\_ch@163.com \\
\addr Department of Information Engineering and Computer Science,\\
University of Trento, Via Sommarive 9, I-38123 Povo (TN), Italy
\AND
\name Yue Song \email yue-song@mail.tsinghua.edu.cn \\
\addr College of AI,\\
Tsinghua University, No. 45 Chengfu Road, Haidian District, Beijing 100083, China
\AND
\name Rui Wang \email cs\_wr@jiangnan.edu.cn \\
\addr School of Artificial Intelligence and Computer Science,\\
Jiangnan University, No. 1800 Lihu Avenue, Wuxi, Jiangsu 214122, China
\AND
\name Xiao-Jun Wu \email wu\_xiaojun@jiangnan.edu.cn \\
\addr School of Artificial Intelligence and Computer Science,\\
Jiangnan University, No. 1800 Lihu Avenue, Wuxi, Jiangsu 214122, China
\AND
\name Nicu Sebe \email niculae.sebe@unitn.it \\
\addr Department of Information Engineering and Computer Science,\\
University of Trento, Via Sommarive 9, I-38123 Povo (TN), Italy
}

\editor{To be assigned}
\maketitle



\begin{abstract}%
Manifold-valued measurements are prevalent in various machine learning tasks. Recent advances have extended Deep Neural Networks (DNNs) to operate on manifolds. These extensions have been accompanied by normalization techniques tailored to different geometries, collectively referred to as Riemannian normalization. However, most existing Riemannian normalization methods are either designed for specific manifolds or fail to effectively normalize manifold-valued sample distributions. To address these limitations, we propose LieBN, a framework for Riemannian Batch Normalization (RBN) over Lie groups. Our approach leverages the theoretically convenient left- and right-invariant metrics, which naturally exist in every Lie group, and provides theoretical guarantees for controlling the Riemannian mean and variance. We instantiate LieBN across nine distinct geometries: four on the Symmetric Positive Definite (SPD) manifold, one on the group of rotation matrices, and four on the manifold of full-rank correlation matrices. Notably, among the SPD metrics, we introduce a novel right-invariant metric and extend three existing Lie group structures via matrix power deformation. Experiments on different manifolds validate the effectiveness of our framework. The code is available at \url{https://github.com/GitZH-Chen/LieBN.git}.
\end{abstract}


\section{Introduction}
\label{sec:intro}

\begin{figure}[t]
\centering
\includegraphics[width=\linewidth,trim={1cm 1cm 1cm 0cm}]{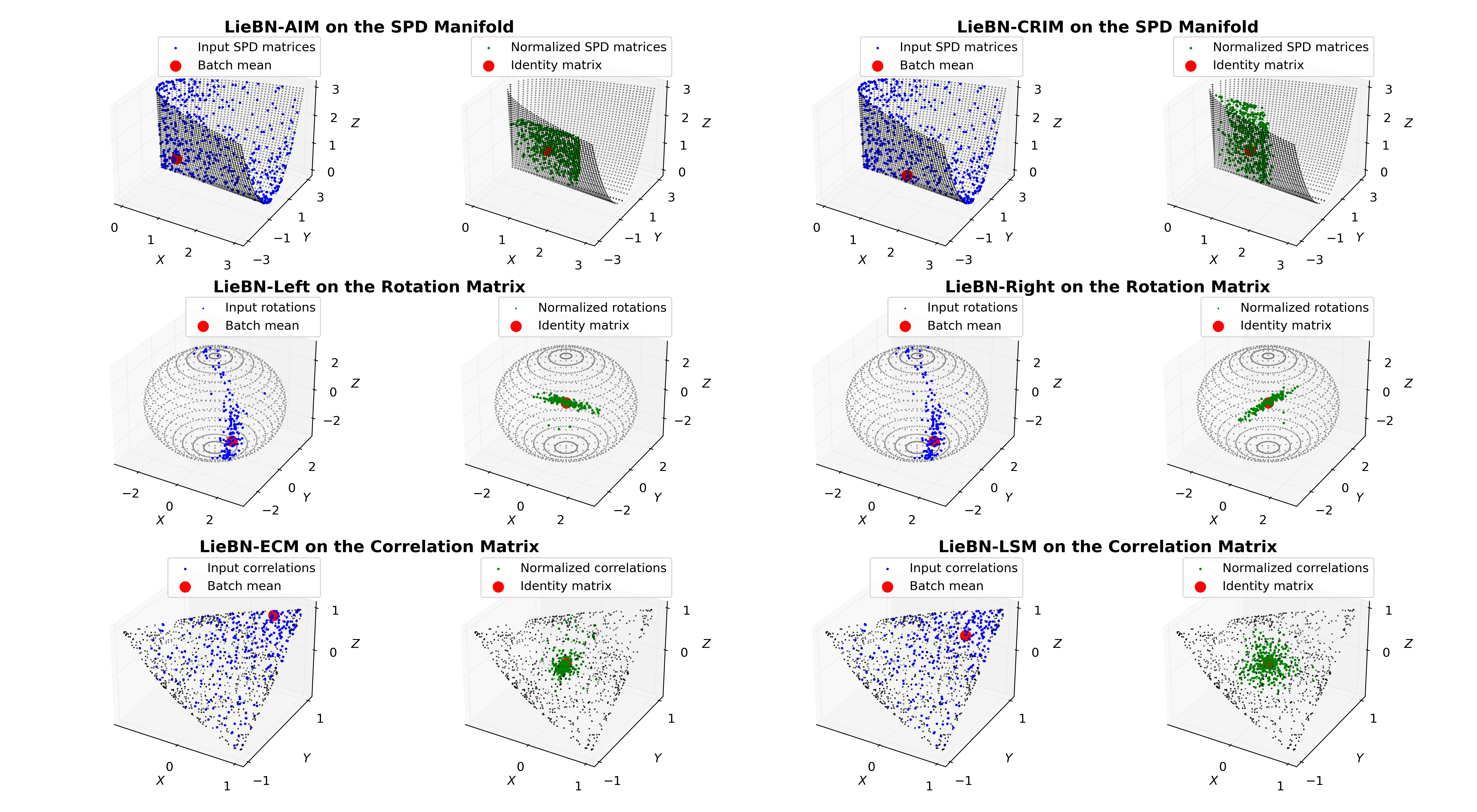}
% trim={left bottom right top}
\caption{Illustration of LieBN on the SPD, rotation, and correlation Lie groups. The $2 \times 2$ SPD, $3 \times 3$ rotation, and $3 \times 3$ correlation manifolds can be embedded into $\mathbb{R}^3$ as an open cone \citep{yair2019parallel}, a closed ball with antipodal points identified \citep{hartley2013rotation}, and an open elliptope \citep{thanwerdas2022theoretically}, respectively. LieBN is illustrated by (1) the left-invariant AIM and the proposed right-invariant CRIM geometry on the SPD manifold, (2) left or right translation under a bi-invariant metric on the rotation manifold, and (3) the bi-invariant ECM and LSM geometry on the correlation manifold. On the SPD and correlation manifolds, the batch mean and variance of the same input samples differ under different geometries. In all subfigures, the black, blue, green, and red dots denote the boundary of the space, the input Lie group samples, the normalized samples, and the batch mean, respectively. As illustrated, our LieBN effectively normalizes the Lie group distribution.
}
\label{fig:illustration}
\vspace{-4mm}
\end{figure}

Over the past decade or so, Deep Neural Networks (DNNs) have achieved significant progress across various scientific fields \citep{hochreiter1997long, krizhevsky2012imagenet, he2016deep, vaswani2017attention}. Traditionally, DNNs have been developed under the assumption that the latent space of the input data is Euclidean. However, many applications involve non-Euclidean structures, such as manifolds \citep{bronstein2017geometric}. To address this challenge, researchers have extended various types of DNNs to manifolds, based on the theories of Riemannian geometry \citep{huang2017riemannian,huang2017deep,huang2018building,ganea2018hyperbolic,chakraborty2018statistical,chakraborty2022manifoldnet,wang2022learning,de2022riemannian,wang2022dreamnet,chen2023riemannian,chen2023distribution,yim2023se,chen2026product,chen2024spdmlr,wangspdmetric2024,wang2024grassatt,chen2024rmlr,chen2025understanding,hu2026riemannian}.

Motivated by the great success of normalization techniques \citep{ioffe2015batch,ba2016layer,ulyanov2016instance,wu2018group}, researchers have sought to devise normalization layers tailored for manifold-valued data. \citet{brooks2019riemannian} introduced Riemannian Batch Normalization (RBN) designed specifically for the Symmetric Positive Definite (SPD) manifold, with the ability to normalize the Riemannian mean. \citet{kobler2022spd} extended this approach to further control the Riemannian variance. \textit{However, the above methods are constrained to the Affine-Invariant Metric (AIM) on the SPD manifold, limiting their applicability.} On the other hand, \citet{chakraborty2020manifoldnorm} proposed two distinct Riemannian normalization frameworks: one for Riemannian homogeneous spaces \citep[Algorithms~1--2]{chakraborty2020manifoldnorm} and another for matrix Lie groups \citep[Algorithms~3--4]{chakraborty2020manifoldnorm}. \textit{Nonetheless, the normalization designed for Riemannian homogeneous spaces cannot normalize either the mean or the variance, while the one for matrix Lie groups is confined to a specific type of distance \citep[Section~3.2]{chakraborty2020manifoldnorm}.} Meanwhile, \citet[Algorithm~2]{lou2020differentiating} proposed an RBN layer for general geometries. However, similar to the methods in \citet[Algorithms~1--2]{chakraborty2020manifoldnorm}, it lacks theoretical guarantees for normalizing sample statistics. Therefore, a principled Riemannian normalization framework capable of controlling both Riemannian mean and variance remains unexplored.

Given that Batch Normalization (BN)~\citep{ioffe2015batch} serves as the foundational prototype for various types of normalization, our paper focuses on RBN, with the potential to be extended to other normalization variants. Since several manifold-valued measurements form Lie groups, such as Symmetric Positive Definite (SPD) manifolds \citep{arsigny2005fast,lin2019riemannian,thanwerdas2022theoretically}, special orthogonal groups $\so{n}$ \citep{boumal2011discrete}, and full-rank correlation matrices \citep{thanwerdas2022theoretically,thanwerdas2024permutation}, we further direct our attention to Lie groups. As each Lie group naturally admits left- and right-invariant metrics \citep[Chapter~1.2]{do1992riemannian}, we propose a principled framework for RBN over Lie groups under invariant metrics, referred to as LieBN. Compared to previous work, our framework provides a theoretical guarantee for normalizing the Riemannian sample mean and variance across general Lie groups.

Empirically, we focus on the SPD, special orthogonal, and full-rank correlation manifolds. On SPD manifolds, we generalize three existing Lie group structures into parameterized ones by matrix power deformation. Additionally, we propose a novel right-invariant metric, referred to as Cholesky Right Invariant Metric (CRIM). We then instantiate our LieBN framework on SPD manifolds under these four Lie group structures. For rotation matrices, we adopt the popular bi-invariant metric \citep{boumal2011discrete}, which will induce two types of LieBN: one w.r.t. left-invariance and another w.r.t. right-invariance. On the correlation manifold, we manifest our LieBN under four recently developed correlation geometries \citep{thanwerdas2022theoretically,thanwerdas2024permutation}. Besides, we discuss the optimization of the involved correlation-valued parameters. To facilitate usage, we provide a LieBN toolbox compatible with PyTorch, which can be used as a drop-in module. \cref{fig:illustration} illustrates our LieBN on different geometries, while \cref{fig:liebn_examples} illustrates a minimal demo. Extensive experiments on SPD, rotation, and correlation manifolds across three tasks---radar recognition, human action recognition, and electroencephalography (EEG) classification---demonstrate the effectiveness of our methods.

\begin{figure}[tbp]
\centering
\begin{lstlisting}[language=Python]
from LieBN import LieBNSPD, LieBNRot, LieBNCor
from LieBN.Geometry.SPD import SPDMatrices
from LieBN.Geometry.Rotations import RotMatrices
from LieBN.Geometry.Correlation import Correlation

# ==== SPD matrices ====
P_spd = SPDMatrices(n=5).random(4, 2, 5, 5)
# Implemented metrics: LEM,ALEM,LCM,AIM,CRIM
liebn_spd = LieBNSPD([2, 5, 5], metric="LEM", batchdim=[0])
output_spd = liebn_spd(P_spd)

# ==== SO(3) matrices ====
P_so3 = RotMatrices().random(4, 2, 3, 3, 3)
# LieBN-Left if is_left else -Right
liebn_so3 = LieBNRot([3, 3, 3], batchdim=[0, 1], is_left=False)
output_so3 = liebn_so3(P_so3)

# ==== Correlation matrices ====
P_cor = Correlation(n=5).random(4, 2, 5, 5)
# Implemented metrics: ECM,LECM,OLM,LSM
liebn_cor = LieBNCor([2, 5, 5], metric="ECM", batchdim=[0])
output_cor = liebn_cor(P_cor)
\end{lstlisting}
\caption{Minimal examples of applying LieBN.}
\label{fig:liebn_examples}
\end{figure}

We emphasize that our work is fundamentally distinct in theory from previous RBN methods \citep{brooks2019riemannian,kobler2022spd,lou2020differentiating} and is more general than ManifoldNorm \citep{chakraborty2020manifoldnorm}. Previous RBN methods are either designed for specific geometries \citep{brooks2019riemannian,kobler2022spd,chakraborty2020manifoldnorm} or fail to control both the mean and variance \citep{lou2020differentiating}. In contrast, our LieBN ensures the normalization of both the mean and variance across general Lie groups. In summary, our main contributions are:
\begin{itemize}
    \item
    A general Lie group batch normalization framework with controllable first- and second-order moments;
    \item
    A novel right-invariant metric on the SPD manifold, which is the first non-trivial right-invariant SPD metric;
    \item
    Concrete instantiations of our LieBN framework on different geometries: four on SPD manifolds, one on the rotation manifold, and four on the correlation manifold;
    \item
    Validation of the effectiveness of our LieBN framework by extensive experiments on different geometries.
\end{itemize}

\mypara{Outline.}
\cref{sec:preliminary} reviews the necessary background on Lie groups and the SPD, rotation, and full-rank correlation Lie groups considered in this work. \cref{sec:revisit_normalization} revisits Euclidean and Riemannian batch normalization, while \cref{sec:liebn} develops the general LieBN framework under invariant metrics. \cref{sec:manifestations} instantiates LieBN on SPD, rotation, and full-rank correlation manifolds. \cref{sec:experiments} reports experiments that evaluate LieBN across these geometries. \cref{sec:conclusions} concludes the paper. All proofs are deferred to \cref{app:sec:proofs}.

This paper extends our previous conference paper~\citep{chen2024liebn} in both theory and implementation. Theoretically, the original LieBN framework was restricted to Lie groups under a left-invariant metric. However, Lie groups also naturally admit right-invariant metrics, which share many theoretical properties with left-invariant ones. Therefore, we generalize LieBN to all natural invariant metrics, including left-, right-, and bi-invariant metrics, providing a more comprehensive framework for RBN. Additionally, we propose a novel non-trivial right-invariant metric on the SPD manifold. In terms of implementation, beyond the original applications to SPD and rotation matrices, we further manifest LieBN on four correlation geometries. Besides, the previous LieBN on $\so{3}$ was based solely on left-invariance and validated on a small data set. In contrast, this journal submission expands the implementation to both left- and right-invariance and conducts extensive experiments across multiple data sets.

\section{Preliminaries}
\label{sec:preliminary}

This section briefly reviews Lie groups, as well as the concrete Lie groups of SPD, rotation, and full-rank correlation matrices. For more in-depth discussions, we refer the reader to \citet{loring2011introduction} for smooth manifolds, \citet{do1992riemannian,lee2018introduction} for Riemannian manifolds, and \citet{hall2015lie} for Lie groups.

\subsection{Lie Groups}

\begin{parisdefinition}[Lie Groups \citep{loring2011introduction}] \label{def:lie_group}
A manifold $\calM$ is a Lie group if it forms a group with a group operation $\odot$ such that $m(\cdot,\cdot): \calM \times \calM \ni (x,y) \to x \odot y \in \calM$ and the group inverse $i(\cdot): \calM \ni x \to x_{\odot}^{-1} \in \calM$ are both smooth.
\end{parisdefinition}

\begin{parisdefinition}[Invariance \citep{do1992riemannian}] \label{def:left_invariance}
A Riemannian metric $\gleft$ over a Lie group $\{\calM, \odot\}$ is left-invariant if for any $x,y \in \calM$ and $V_1,V_2 \in T_x\calM$, it satisfies $\gleft _y(V_1,V_2) = \gleft_{\ltrans _x(y)} \left(\ltrans _{x*,y}(V_1), \ltrans _{x*,y}(V_2) \right)$, with $\ltrans _x(y) = x \odot y$ as the left translation by $x$, and $\ltrans _{x*,y}$ as the differential map of $\ltrans _x$ at $y$. Similarly, a right-invariant metric $\gright$ satisfies $\gright _y(V_1,V_2) = \gright_{\rtrans _x(y)} \left(\rtrans _{x*,y}(V_1), \rtrans _{x*,y}(V_2) \right)$, with $\rtrans _x(y) = y \odot x$ as the right translation by $x$, and $\rtrans _{x*,y}$ as the differential at $y$.
\end{parisdefinition}

A Lie group is both a group and a manifold. The most natural Riemannian metric on a Lie group is the left- or right-invariant metric.\footnote{An invariant metric always exists for every Lie group \citep[Chapter~1.2]{do1992riemannian}.} A bi-invariant metric has both left- and right-invariance. In this paper, $\{\calM, \odot, g\}$, abbreviated as $\calM$, always denotes a Lie group with an invariant metric.

The idea of pullback is ubiquitous in differential geometry. A diffeomorphism can be viewed as the smooth counterpart of a set-theoretic bijection.

\begin{parisdefinition} [Pullback Metrics \citep{lee2018introduction}]
    Suppose $\calM_1,\calM_2$ are smooth manifolds, $g$ is a Riemannian metric on $\calM_2$, and $f:\calM_1 \rightarrow \calM_2$ is a diffeomorphism. The pullback of $g$ by $f$ is defined pointwise as $(f^*g)_p(V,W) = g_{f(p)}(f_{*,p}(V),f_{*,p}(W))$, where $f_{*,p}(\cdot)$ is the differential map of $f$ at $p \in \calM_1$, and $V,W \in T_p\calM_1$. $f^*g$ is a Riemannian metric on $\calM_1$, called the pullback metric of $g$ by $f$.
\end{parisdefinition}

Although pullback metrics can also be defined by a smooth map \citep{lee2018introduction}, we focus on the case where the map is a diffeomorphism. Additionally, if $\{\calM_2, \odot_2 \}$ forms a Lie group, the diffeomorphism $f$ can pull back the group operation $\odot_2$ to $\odot_1$ on $\calM_1$:
\begin{equation} \label{eq:group_iso}
    P \odot_1 Q = f^{-1}(f(P) \odot_2 f(Q)), \forall P, Q \in \calM_1.
\end{equation}

\begin{parisdefinition} [Weighted Fréchet Mean \& Variance \citep{frechet1948elements}] \label{def:wfm}
Let $\{w_{1 \ldots N}\}$ be weights satisfying a convexity constraint, \ie, $\forall i, w_i>0$ and $\sum_i w_i=1$. The weighted Fréchet mean (WFM) of a set of manifold-valued points $\{ P_{i \ldots N} \in \mathcal{M} \}$ is defined as
\begin{equation} \label{eq:wfm_spd}
    \wfm \left( \{w_i\}, \{P_{i} \} \right) = \underset{M \in \calM}{\argmin} \sum\nolimits_{i=1}^N  w_i\dist^2\left(P_i, M \right),
\end{equation}
where $\dist(\cdot,\cdot)$ denotes the geodesic distance. When $w_i=\nicefrac{1}{N}$ for all $i$, \cref{eq:wfm_spd} is reduced to the Fréchet mean, denoted as $\fm (\{P_{i} \})$. The Fréchet variance $v^2$ is the attained value at the minimizer of the Fréchet mean.
\end{parisdefinition}

On Riemannian manifolds, the WFM exists locally and is unique \citep{afsari2011riemannian}, as detailed in \cref{app:subsec:exist_unique_wfm} for completeness. In this paper, we always assume the Fréchet batch mean exists. As we focus on Riemannian manifolds, we will use the terms ``Riemannian mean'' and ``Fréchet mean'' interchangeably, as well as ``Riemannian variance'' and ``Fréchet variance.''

\mypara{Basic notations.}
For Euclidean spaces, we denote $\inner{\cdot}{\cdot}$ as the canonical inner product over $\bbR{n \times n}$, with $\fnorm{\cdot}$ as the induced norm, \ie, $F$-norm for matrices. For a manifold $\calM$, we denote $\rielog_P$, $\rieexp_P$, and $\inner{\cdot}{\cdot}_{P}=g_P(\cdot,\cdot)$ as the Riemannian logarithm, exponentiation, and metric at $P \in \calM$, respectively. Besides, $\gamma_{(P,Q)}(t)$ denotes the geodesic connecting $P$ and $Q$, and $\dist(\cdot,\cdot)$ denotes the geodesic distance. We provide a complete table of notations in \cref{app:notations}.

\subsection{SPD Lie Groups}

\begin{table}[t]
    \centering
    % \resizebox{0.99\linewidth}{!}{%
    \begin{tabular}{c|ccc}
        \toprule
        Operator & $\biparamAIM$ & $\biparamLEM$ & LCM \\
        \midrule
        $Q \odot P$ & $KPK^\top$ & $\mexp(\mlog(P)+\mlog(Q))$ & $\chol^{-1}(\lfloor L + K \rfloor + \bbK \bbL)$ \\
        $P_{\odot}^{-1}$ & $\chol^{-1}(L^{-1})$ & $\mexp(-\mlog(P))$ & $\chol^{-1}(-\lfloor L\rfloor+\bbL^{-1})$ \\
        Identity & $I_n$ & $I_n$ & $I_n$ \\
        Invariance & Left-invariance & Bi-invariance & Bi-invariance \\
        \bottomrule
    \end{tabular}
    % }
    \caption{SPD Lie groups and invariant metrics.}
    \label{tab:riem_lie_spd}
\end{table}
The SPD manifold has shown great success in diverse applications \citep{huang2017riemannian,chakraborty2022manifoldnet,wang2020deep,lopez2021vector,chen2023hybrid,kobler2022spd}. We denote $n \times n$ SPD matrices as $\spd{n}$ and $n \times n$ real symmetric matrices as $\sym{n}$. As shown by \citet{arsigny2005fast}, $\spd{n}$ forms an open submanifold of the Euclidean space $\sym{n}$, known as the SPD manifold. SPD manifolds exhibit three Lie group structures, each associated with an invariant metric. These metrics include the Log-Euclidean Metric (LEM) \citep{arsigny2005fast}, Affine-Invariant Metric (AIM) \citep{pennec2006riemannian}, and Log-Cholesky Metric (LCM) \citep{lin2019riemannian}. We denote LEM and AIM as $\biparamLEM$ and $\biparamAIM$, as they are induced by the following $\orth{n}$-invariant inner product on the tangent space at the identity matrix \citep{thanwerdas2023n}:
\begin{equation} \label{eq:oim_sym}
    \langle V,W \rangle^{\alphabeta}=\alpha \langle V,W \rangle + \beta \tr(V)\tr(W),
\end{equation}
where $V,W \in T_I\spd{n} \cong \sym{n}$, and $(\alpha,\beta) \in \bfst = \{(\alpha,\beta) \in \mathbb{R}^2 \mid \min (\alpha, \alpha+n \beta)>0\}$. \cref{tab:riem_lie_spd} summarizes the Lie structures on SPD manifolds with the following notations. Let $P, Q \in \spd{n}$ be SPD matrices. We denote the matrix logarithm, exponentiation, and Cholesky decomposition by $\mlog(\cdot)$, $\mexp(\cdot)$, and $\chol(\cdot)$, respectively. The Cholesky factors of $P$ and $Q$ are $L=\chol(P)$ and $K=\chol(Q)$. $\lfloor \cdot \rfloor$ returns the strictly lower triangular part of a square matrix. Other Riemannian operators are summarized in \cref{app:subsec:spd_geometries}.

For $\biparamLEM$ and LCM, the Fréchet mean admits a closed-form expression, as shown in \cref{app:tab:riem_operators_props}. Moreover, since $\biparamAIM$ has non-positive sectional curvature \citep[Table~5]{thanwerdas2023n}, the WFM is a global operation \citep[Chapter~6.1.5]{berger2003panoramic} and can be computed by the Karcher flow algorithm \citep{karcher1977riemannian}.

\subsection{Rotation Lie Groups}
\begin{wraptable}{r}{0.4\linewidth}
    \centering
    % \resizebox{\linewidth}{!}{
    \begin{tabular}{cc}
        \toprule
        Operator & Expression \\
        \midrule
        $R \odot S$ & $RS$ \\
        $R_{\odot}^{-1}$ & $R^{-1}=R^\top$ \\
        Identity & $I_n$ \\
        $g_R(A_1,A_2)$ & $\langle A_1,A_2 \rangle$ \\
        $\dist(R,S)$ & $\left\|\mlog\left(R^\top S\right)\right\|_\rmF$ \\
        $\rielog_R S$ & $R\mlog\left(R^\top S\right)$ \\
        $\rieexp_R(A)$ & $R\mexp\left(R^\top A\right)$ \\
        $\gamma_{(R,S)}(t)$ & $R\mexp\left(t\mlog\left(R^\top S\right)\right)$ \\
        FM & Karcher flow \\
        Invariance & Bi-invariance \\
        \bottomrule
    \end{tabular}
    % }
    \caption{Operators on $\so{n}$.}
    \label{tab:riem_rotation}
    \vspace{-5mm}
\end{wraptable}
The set of $n \times n$ rotation matrices forms a Lie group, known as the special orthogonal group, denoted as $\so{n}$ \citep{loring2011introduction}. Its group operation is the matrix product, with the identity matrix as the neutral element. Any tangent vector $A$ in $T _R \so{n}$ can be represented as $A = RV$, with $V \in \soLieAlgebra{n}$. Here, $\soLieAlgebra{n}$ is the Lie algebra of $\so{n}$, which is the tangent space at the identity matrix, formed by the set of $n \times n$ skew-symmetric matrices. The Fréchet mean can be obtained by Karcher flow \citep{manton2004globally}. Furthermore, if all the rotations lie in a closed ball of radius $r< \nicefrac{\pi}{2}$, then Karcher flow converges to the unique mean \citep[Theorem~5.1]{manton2004globally}. The associated Riemannian operators are summarized in \cref{tab:riem_rotation}, where $R,S \in \so{n}$ are rotation matrices and $A,A_1,A_2 \in T_R\so{n}$ are tangent vectors at $R$.

\subsection{Full-Rank Correlation Lie Groups}
\label{subsec:cor_lie_groups}

\begingroup\tolerance=1000\emergencystretch=10pt
The correlation matrix of a covariance matrix $\Sigma$ is defined as $C =\allowbreak \coropt(\Sigma) =\allowbreak \bbD(\Sigma)^{-\nicefrac{1}{2}} \Sigma \bbD(\Sigma)^{-\nicefrac{1}{2}}$, where $\bbD(\cdot)$ returns a diagonal matrix with diagonal elements of $\Sigma$. The space of $n \times n$ full-rank correlation matrices, denoted as $\cor{n}$, forms a quotient manifold of the SPD manifold $\spd{n}$ \citep[Theorem~1]{david2019riemannian}, referred to as the correlation manifold. This manifold can be interpreted as a compactly normalized SPD manifold that encodes scale-invariant information \citep{thanwerdas2022theoretically}. However, its Riemannian structure has been less studied than SPD matrices. Recently, \citet{thanwerdas2022theoretically} developed two convenient Riemannian metrics: the Euclidean-Cholesky Metric (ECM) and Log-Euclidean-Cholesky Metric (LECM). \citet{thanwerdas2024permutation} further proposed two permutation-invariant metrics, the Off-Log Metric (OLM) and Log-Scaled Metric (LSM). All four geometries above are pullback metrics from simpler Euclidean spaces. We first review the related prototype spaces, followed by an examination of the four Riemannian metrics. \par
\endgroup

\begin{itemize}
    \item $\LTone{n}$: Euclidean space of $n \times n$ lower triangular matrices with unit diagonals.
    \item $\LTzero{n}$: Euclidean space of $n \times n$ lower triangular matrices with null diagonals.
    \item $\hol{n}$: Euclidean space of $n \times n$ symmetric matrices with null diagonals. The tangent space $T_C \cor{n}$ at $C \in \cor{n}$ can be identified with $\hol{n}$.
    \item $\rzero{n}$: Euclidean space of $n \times n$ symmetric matrices with null row sum.
\end{itemize}

\textit{ECM} is derived from $\LTone{n}$ by
\begin{equation}
    \cor{n}\allowbreak \xrightleftharpoons[\Theta^{-1} = \covtocor \circ \chol^{-1}]{\Theta} \LTone{n},
\end{equation}
where $\Theta(C) =\allowbreak \bbD(\chol(C))^{-1} \chol(C)$ for any $C \in \cor{n}$.

\textit{LECM} is defined by further pulling back ECM: $\cor{n}\allowbreak \xrightleftharpoons[(\log \circ \Theta)^{-1} = \covtocor \circ \chol^{-1} \circ \exp]{\log \circ \Theta} \LTzero{n}$. Due to the nilpotency of $\LTzero{n}$, the matrix logarithm $\log(\cdot): \LTone{n} \rightarrow \LTzero{n}$ and its inverse $\exp(\cdot)$ over $\LTzero{n}$ are free from eigendecomposition.

\textit{OLM} is derived from a permutation-invariant inner product on $\hol{n}$. The associated diffeomorphism is $\cor{n}\allowbreak \xrightleftharpoons[\offexp]{\offlog=\off \circ \log} \hol{n}$. Here, $\off(\cdot)$ returns a matrix in $\hol{n}$ consisting of off-diagonal elements. For any symmetric hollow matrix $H \in \hol{n}$, there exists a unique diagonal matrix $\dplus(H)$ such that $\offexp(H) = \exp(\dplus(H) + H) \in \cor{n}$. The map $\offexp$ is a diffeomorphism, and $\dplus(H)$ can be computed by the following exponentially convergent algorithm: $D_{k+1}= D_k -\log (\bbD(\exp (D_k + H)))$ \citep[Section~5]{archakov2021new}.

\textit{LSM} is derived from a permutation-invariant inner product on $\rzero{n}$. The associated diffeomorphism is $\cor{n}\allowbreak \xrightleftharpoons[\expscaled=\coropt \circ \exp]{\logscaled} \rzero{n}$. For any correlation matrix $C \in \cor{n}$, there exists a unique positive diagonal matrix $\dstar (C)$ such that $\logscaled(C) = \log ( \dstar (C) C \dstar (C)) \in \rzero{n}$ is a diffeomorphism, where $\dstar (C)$ could be solved by damped Newton's method \citep[Section~3.5]{thanwerdas2024permutation}.

\begin{table}[t]
    \centering
    \begin{tabular}{c|cccc}
        \toprule
        Operator & ECM & LECM & OLM & LSM \\
        \midrule
        $C \odot C'$ & \multicolumn{4}{c}{$f^{-1}\left(f(C)+f(C')\right)$} \\
        $C_{\odot}^{-1}$ & \multicolumn{4}{c}{$f^{-1}\left(-f(C)\right)$} \\
        Identity & \multicolumn{4}{c}{$f^{-1}(\bfzero_{n\times n})$} \\
        WFM & \multicolumn{4}{c}{$f^{-1}\left(\sum_{i=1}^N w_i f(C_i)\right)$} \\
        Invariance & \multicolumn{4}{c}{Bi-invariance} \\
        \bottomrule
    \end{tabular}
    \caption{Full-rank correlation Lie groups and invariant metrics. Here $f$ denotes $\Theta$, $\log \circ \Theta$, $\offlog$, and $\logscaled$ for ECM, LECM, OLM, and LSM, respectively.}
    \label{tab:riem_lie_cor}
\end{table}

The group operations under these four metrics are defined as in \cref{eq:group_iso}:
\begin{equation}
    P \odot Q = f^{-1}(f(P) + f(Q)), \quad \forall P, Q \in \cor{n},
\end{equation}
with $f$ as $\Theta$, $\log \circ \Theta$, $\offlog$, and $\logscaled$ for ECM, LECM, OLM, and LSM, respectively. The following discusses the invariance and WFM.

\begin{parisproposition}[Invariance \& WFM]
    \label{props:invariance_correlation}
    \linktoproof{props:invariance_correlation}
    ECM, LECM, OLM, and LSM are bi-invariant. Following the notations in \cref{def:wfm}, the WFM for a batch of samples $\{ P_{i \ldots N} \in \cor{n} \}$ is $f^{-1} \left( \sum_{i=1} ^N w_i f(P_i) \right)$, with $f$ as $\Theta$, $\log \circ \Theta$, $\offlog$, and $\logscaled$ for ECM, LECM, OLM, and LSM, respectively.
\end{parisproposition}

The Lie structures under these metrics are summarized in \cref{tab:riem_lie_cor}. More details on Riemannian structures are presented in \cref{app:subsec:cor_geometries}.

\section{Revisiting Normalization}
\label{sec:revisit_normalization}

\subsection{Revisiting Euclidean Normalization}
In Euclidean DNNs, normalization is a significant technique for accelerating network training by mitigating the issue of internal covariate shift \citep{ioffe2015batch}. While various normalization methods have been introduced \citep{ioffe2015batch, ba2016layer, ulyanov2016instance, wu2018group}, they all share a common purpose: the normalization of the first and second moments. This paper focuses on Batch Normalization (BN), the prototype of other normalization variants.

Given a batch of activations $\{x_{i \ldots N } \}$, the core operations in the standard Euclidean BN can be expressed as
\begin{equation} \label{eq:ebn}
    \forall i \leq N, x_i \gets \gamma \frac{x_i-\mu_{b}}{\sqrt{v^2_{b}+\epsilon}} + \beta,
\end{equation}
where $\mu_{b}$ is the batch mean, $v^2_b$ is the batch variance, $\gamma$ is the scaling parameter, $\beta$ is the biasing parameter, and $\epsilon$ is a small scalar for stability.

\subsection{Revisiting Riemannian Batch Normalization}
\label{subsec:revisit_rbn}

Although endeavors have been made to develop Riemannian normalization approaches tailored for manifolds, none of the existing methods effectively handle the first and second moments in a principled manner.

\citet{brooks2019riemannian} introduced RBN over SPD manifolds under AIM. The core operations are defined as follows:
\begin{align}
    \label{eq:spdnetbn_centering}
    \text{Centering from mean } M \in \spd{n}: \bar{P}_i \gets M^{-\frac{1}{2}} P_i M^{-\frac{1}{2}}, \\
    \label{eq:spdnetbn_biasing}
    \text{Biasing towards parameter } B \in \spd{n}: \hat{P}_i \gets B^{\frac{1}{2}} \bar{P}_i B^{\frac{1}{2}},
\end{align}
where $\{P_{i \ldots N}\}$ are SPD matrices, and $M$ is their Fréchet mean under AIM. Let $\spdtrans{P}{Q}(S)=\rieexp_{Q} \left[\pt{P}{Q} \left(\rielog_P (S)\right)\right]$, where $P, Q, S \in \spd{n}$. Under AIM, \cref{eq:spdnetbn_centering,eq:spdnetbn_biasing} can be more generally expressed as
\begin{equation} \label{eq:brooksbn_prototype}
    \spdtrans{I}{B} [\spdtrans{M}{I}(P_i)].
\end{equation}
However, \cref{eq:spdnetbn_centering,eq:spdnetbn_biasing} only consider the Riemannian mean and do not consider the Riemannian variance.\footnote{Although not discussed in \citep{brooks2019riemannian}, \cref{eq:spdnetbn_centering,eq:spdnetbn_biasing} as congruent actions can transfer the batch mean into a desirable one under AIM.} To remedy this limitation, \citet{kobler2022spd} further extended the RBN to involve the second-order statistics. The key operation is formulated as
\begin{equation} \label{eq:kobler_rbn}
    \forall i \leq N, \bar{P}_i \gets \spdtrans{I}{B}[(\spdtrans{M}{I}(P_i))^{\frac{s}{v}}],
\end{equation}
where $v^2$ is the Fréchet variance and $s \in \bbRscalar$ is a scaling factor. However, this method is still limited to SPD manifolds under AIM. In parallel, \citet[Algorithms~1--2]{chakraborty2020manifoldnorm} proposed a general framework for Riemannian homogeneous spaces based on \cref{eq:brooksbn_prototype}, which involves both first and second moments. However, \cref{eq:brooksbn_prototype} does not generally guarantee control of the Riemannian mean, resulting in agnostic Riemannian statistics \citep[Section~3.1]{chakraborty2020manifoldnorm}. To mitigate this limitation, \citet[Algorithms~3--4]{chakraborty2020manifoldnorm} further proposed normalization over the matrix Lie group. However, the discussion is limited to a certain distance, limiting the applicability of their method. On the other hand, \citet[Algorithm~2]{lou2020differentiating} proposed an RBN based on a variant of \cref{eq:brooksbn_prototype}. Similarly, their approach suffers from the same problem of agnostic Riemannian statistics of the output samples.

In summary, prevailing Riemannian normalization approaches lack a principled guarantee for controlling the first- and second-order statistics. In contrast, our method can normalize first- and second-order statistics over Lie groups. \cref{tab:sum_rbn} summarizes the above RBN methods.

\begin{table}[t]
    \centering
    \resizebox{0.99\linewidth}{!}{
    \begin{tabular}{ccccc}
         \toprule
         Methods & \makecell{Involved \\ Statistics} & \makecell{Controllable \\ Mean} & \makecell{Controllable \\ Variance}  & Geometries\\
         \midrule
         SPDBN \citep[Algorithm~1]{brooks2019riemannian} & Mean & \cmark & \na & SPD manifolds under AIM \\
         SPDBN \citep[Algorithm~1]{kobler2022controlling} & Mean+Variance & \cmark & \cmark & SPD manifolds under AIM \\
         SPDDSMBN \citep{kobler2022spd} & Mean+Variance & \cmark & \cmark & SPD manifolds under AIM \\
         ManifoldNorm \citep[Algorithms~1--2]{chakraborty2020manifoldnorm} & Mean+Variance & \xmark & \xmark & Riemannian homogeneous space \\
        ManifoldNorm \citep[Algorithms~3--4]{chakraborty2020manifoldnorm} & Mean+Variance & \cmark & \cmark & A specific Lie group structure and distance  \\
         RBN \citep[Algorithm~2]{lou2020differentiating} & Mean+Variance & \xmark & \xmark & Geodesically complete manifolds \\
         \midrule
         \rowcolor{HilightColor} LieBN (Ours) & Mean+Variance & \cmark & \cmark & Lie groups\\
         \bottomrule
    \end{tabular}
    }
    \vspace{-3mm}
    \caption{Summary of some representative RBN methods.}
    \label{tab:sum_rbn}
\end{table}

\section{Lie Group Batch Normalization}
\label{sec:liebn}
Since every Lie group naturally admits invariant metrics, we propose batch normalization over Lie groups based on invariant metrics, referred to as LieBN. We first introduce the core operations under left-invariant metrics and then extend them to right-invariant metrics. Finally, we present the theoretical LieBN framework. In the following, we denote the neutral element in the Lie group $\calM$ as $E$.\footnote{The neutral element $E$ is not necessarily the identity matrix.}

\subsection{Ingredients Under Left-Invariant Metrics}

In this subsection, we always assume that the Lie group $\calM$ admits a left-invariant metric $\gleft$. Recalling the standard Euclidean BN \citep{ioffe2015batch} in \cref{eq:ebn}, two key points are noteworthy: (a) the Euclidean BN implicitly assumes a Gaussian distribution and can effectively normalize the latent Gaussian distribution; (b) the centering and biasing operations control the mean, while the scaling controls the variance. Therefore, extending BN into Lie groups requires the counterparts of Gaussian distribution, centering, biasing, and scaling.

There are several notions of Gaussian distribution over manifolds \citep{pennec2004probabilities,wang2006error,chakraborty2019statistics,barbaresco2021gaussian}. We adopt the intrinsic definition from \citet{chakraborty2019statistics}, which characterizes a Gaussian distribution on the Lie group $\calM$ with a mean parameter $M \in \calM$ and variance $\sigma^2$. This distribution is denoted as $\calN(M,\sigma^2)$, and its Probability Density Function (P.D.F.) is
\begin{equation} \label{eq:lie_gaussian}
    p\left(X \mid M, \sigma^2\right)=k(\sigma) \exp \left(-\frac{\dist(X, M)^2}{2 \sigma^2}\right),
\end{equation}
where $k(\sigma)$ is the normalizing constant and $\dist(\cdot,\cdot)$ is the geodesic distance. When $\calM$ is $\bbRscalar$ with the standard Euclidean metric, \cref{eq:lie_gaussian} reduces to the Euclidean Gaussian.

On Lie groups, the natural counterparts of addition and subtraction in \cref{eq:ebn} are group operations. Therefore, centering and biasing on Lie groups can be defined by the left translation. Additionally, we define scaling via the tangent space. Specifically, for a batch of activations $\{P_{i \ldots N} \in \calM \}$, we define the key operations of LieBN as follows:
\begin{align}
    \label{eq:liebn_centering}
    \text{Centering from mean } M \in \calM: \bar{P}_i \gets L_{M_{\odot}^{-1}}(P_i), \\
    \label{eq:liebn_scaling}
    \text{Scaling: } \hat{P}_i \gets \rieexp_{E} \left [ \frac{s}{\sqrt{v^2+\epsilon}} \rielog_{E}(\bar{P}_i) \right], \\
    \label{eq:liebn_biasing}
    \text{Biasing towards parameter } B \in \calM: \tilde{P}_i \leftarrow L_B\left(\hat{P}_i\right),
\end{align}
where $M$ is the Fréchet mean, $v^2$ is the Fréchet variance, $M_{\odot}^{-1} \in \calM$ is the group inverse of $M$, $L_{M_{\odot}^{-1}}$ and $L_{B}$ are left translations ($L_{B}(P_i)= B \odot P_i$), and $s \in \bbRscalar / \{0\}$ is a scaling parameter. The following two propositions demonstrate how the above operations normalize the mean and variance: one related to population statistics and the other related to sample statistics.

\begin{parisproposition}[Population]
    \label{props:population_gaussian}
    \linktoproof{props:population_gaussian}
    Given a random point $X \sim \calN(M,v^2)$ on the Lie group $\{ \calM, \gleft \}$, where $\calN(M,v^2)$ is defined in \cref{eq:lie_gaussian}, we have the following for the population statistics:
    \begin{enumerate}
        \item \label{pro:mle_m}
        (MLE of $M$) Given i.i.d. samples $\{P_{i \ldots N} \in \calM \}$ from $\calN(M,v^2)$, the maximum likelihood estimator (MLE) of $M$ is the sample Fréchet mean.
        \item (Gaussian homogeneity) \label{pro:hom}
        Given $X \sim \calN(M,v^2)$ and $B \in \calM$, $\ltrans _{B}(X) \sim \calN(\ltrans_B(M),v^2)$.
    \end{enumerate}
\end{parisproposition}
\begin{parisproposition} [Sample]
    \label{props:samples}
    \linktoproof{props:samples}
    Given $N$ samples $\{P_{i \ldots N} \}$ over the Lie group $\{ \calM, \gleft \}$, denoting $\phi_{s}(P_i)=\rieexp_{E} \left [ s \rielog_{E}(P_i) \right]$, we have the following for the sample statistics.
    \begin{itemize}
        \item Sample mean homogeneity:
        \begin{equation}
            \label{eq:hom_fm_lie_group}
            \fm\{\ltrans _{B} (P_i) \} = \ltrans _{B} (\fm\{ P_i \}), \forall B \in \calM.
        \end{equation}

        \item Controllable dispersion from $E$:
        \begin{equation}
            \label{eq:variance_lie_group}
            \sum\nolimits_{i=1}^N  w_i \dist^2(\phi_{s}(P_i), E) = s^2 \sum\nolimits_{i=1}^N w_i \dist^2(P_i, E),
        \end{equation}
    \end{itemize}
    where $\{w_{1 \ldots N}\}$ are weights satisfying a convexity constraint, \ie, $\forall i, w_i>0$ and $\sum_i w_i=1$.
\end{parisproposition}

\cref{props:population_gaussian} and \cref{eq:hom_fm_lie_group} imply that our centering and biasing in \cref{eq:liebn_centering,eq:liebn_biasing} can transfer both the sample mean and the population mean. As the post-centering mean is $E$, \cref{eq:variance_lie_group} implies that \cref{eq:liebn_scaling} can control the sample variance. More interestingly, the latent Gaussian distribution can be transferred under some geometries, such as SPD manifolds under LEM and LCM, which are discussed in \cref{app:sec:result_scaling}.

\begin{parisremark}
The MLE of the mean of the Gaussian distribution has been examined in several previous works \citep{said2017riemannian,chakraborty2019statistics,chakraborty2020manifoldnorm}. However, these studies primarily focus on particular manifolds or specific metrics. In contrast, our contribution lies in presenting a general result for Lie groups.
\end{parisremark}
\begin{parisremark}
    While the Gaussian model in \cref{eq:lie_gaussian} was used by \citet{kobler2022controlling}, the authors only focus on SPD manifolds under AIM. The transformation of the population under their proposed RBN remains unexplored as well. Besides, while \citet{chakraborty2020manifoldnorm} analyzed the population properties for their RBN over matrix Lie groups, their results were confined to a specific distance. In contrast, our work provides a more extensive examination, encompassing both population and sample properties of our LieBN in a general manner.
\end{parisremark}

\subsection{Ingredients Under Right-Invariant Metrics}
The key insight underlying \cref{eq:liebn_centering,eq:liebn_biasing,props:population_gaussian,props:samples} is that left translation is an isometry under left-invariant metrics. Similarly, right translation is an isometry under right-invariant metrics. Therefore, it can be used for centering and biasing under right-invariant metrics. Following the previous notations, we define the centering and biasing under a right-invariant metric $\gright$ as
\begin{align}
    & \text{ centering to $E$: }
    \bar{P}_i \gets \rtrans _{M_{\odot}^{-1}}(P_i),\\
    & \text{ biasing towards $B$: }
    \tilde{P}_i \gets \rtrans _{B}(\hat{P}_i).
\end{align}

Similar to the case under left-invariant metrics, \cref{props:population_gaussian,props:samples} can be easily extended into the ones under right-invariant metrics. Notably, the proofs for MLE of $M$ in \cref{props:population_gaussian} and controllable dispersion in \cref{props:samples} can be directly applied to the right-invariant metric. Therefore, we only show the homogeneity in the following proposition.

\begin{parisproposition}
    \label{props:core_operation_right}
    \linktoproof{props:core_operation_right}
    Given a random point $X \sim \calN(M,v^2)$ over $\{ \calM, \gright \}$, $B \in \calM$, and $N$ samples $\{P_{i \ldots N} \}$ over $\calM$, we have
    \begin{enumerate}
        \item \label{pro:hom_gauss_right}
        Gaussian homogeneity: $\rtrans_{B}(X) \sim \calN(\rtrans_B(M),v^2)$;
        \item
        Sample homogeneity: $\fm\{\rtrans_{B} (P_i) \} = \rtrans_{B} (\fm\{ P_i \})$.
    \end{enumerate}
\end{parisproposition}

\begin{table}[t]
    \centering
    \resizebox{0.7\linewidth}{!}{
    \begin{tabular}{c|ccc|c}
        \toprule
        Commutativity & \multicolumn{3}{c|}{Non-commutative} & Commutative \\
        \midrule
        Invariance & Left & Right & Bi & Left = Right = Bi  \\
        \midrule
        LieBN Types & Left & Right & Left \& Right & Left = Right \\
        \bottomrule
    \end{tabular}
    }
    \caption{Summary of LieBN types.}
    \label{tab:liebn_types}
\end{table}%

\subsection{LieBN Under Invariant Metrics}

\begin{algorithm}[!t] \SetKwInOut{Input}{Input}\SetKwInOut{Output}{Output}\SetKwInOut{Parameters}{Parameters}
\caption{Lie Group Batch Normalization (LieBN)}
\label{alg:liebn}
\Input{
A batch of activations $\{P_{1 \ldots N} \}$ over Lie groups $\{ \calM, \odot, g\}$, a small positive constant $\epsilon$, momentum $\gamma \in [0,1]$, running mean $M_r=E$, running variance $v^2_r=1$, biasing parameter $B \in \calM$, and scaling parameter $s \in \bbRscalar/\{0\}$.
}

\Output{Normalized activations $\{\tilde{P}_{1 \ldots N} \}$.}
\BlankLine
\uIf{training}{
    Compute batch mean $M_b$ and variance $v_b^2$\\
    Update running statistics:
    $M_r \gets \wfm(\{1-\gamma,\gamma\},\{M_r,M_b\})$
    $v^2_r \gets (1-\gamma)v^2_r + \gamma v^2_b$

    Use the batch statistics, $M \gets M_b, v^2 \gets v^2_b$
}
\Else{Use the running statistics, $M \gets M_r, v^2 \gets v^2_r$}

\For{$i \gets 1$ \KwTo $N$}{
    Centering to the neutral element $E$: \\
    \Indp \uIf{$g$ is left-invariant}{$\bar{P}_i \gets \ltrans _{M_{\odot}^{-1}}(P_i)$}
    \Else{$\bar{P}_i \gets \rtrans _{M_{\odot}^{-1}}(P_i)$} \Indm

    Scaling the variance: \\
    \hspace{1.5em} $\hat{P}_i \gets \rieexp_{E} \left[ \frac{s}{\sqrt{v^2+\epsilon}} \rielog_{E}(\bar{P}_i) \right]$

    Biasing towards parameter $B$: \\
    \Indp \uIf{$g$ is left-invariant}{$\tilde{P}_i \gets \ltrans _{B}(\hat{P}_i)$}
    \Else{$\tilde{P}_i \gets \rtrans _{B}(\hat{P}_i)$} \Indm
}
\end{algorithm}

With the above ingredients, \cref{alg:liebn} presents our theoretical LieBN framework. Similar to \citet{ioffe2015batch}, we use the moving average to update the running statistics. For a bi-invariant metric, LieBN can be implemented using either left or right translation. If the Lie group is commutative, the LieBN variants under left and right translations are equivalent. \cref{tab:liebn_types} summarizes the LieBN types under different conditions.

The centering and biasing in Euclidean BN correspond to the group action of $\bbRscalar$. From a geometric perspective, the standard Euclidean metric is invariant under this group operation. Consequently, it is not surprising that our LieBN algorithm naturally generalizes the standard Euclidean BN.

\begin{parisproposition} \label{prop:liebn_natural_extension_ebn}
    \linktoproof{prop:liebn_natural_extension_ebn}
    The LieBN algorithm presented in \cref{alg:liebn} is equivalent to the standard Euclidean BN when $\calM= \bbR{n}$, both during the training and testing phases.
\end{parisproposition}

\section{Manifestations}
\label{sec:manifestations}
This section instantiates our LieBN in \cref{alg:liebn} on nine different Lie groups, including four on the SPD manifold, one on the rotation manifold, and four on the correlation manifold.

\subsection{LieBN on SPD Manifolds}
\label{subsec:sec:liebn_spd}
We first extend the existing Lie group structures on SPD manifolds via matrix power deformation, resulting in three families of parameterized Lie groups. Then, we propose a novel right-invariant metric on the SPD manifold, the first non-trivial right-invariant metric on this manifold. Finally, we construct LieBN layers based on these Lie structures.

\subsubsection{Deformed Lie Structures on SPD Manifolds} \label{subsubsec:spd_param_lie_groups}
As shown in \cref{tab:riem_lie_spd}, there are three Lie groups on SPD manifolds, each with a left-invariant metric. These metrics include $\biparamAIM$, $\biparamLEM$, and LCM. For clarity, we denote the group operations w.r.t. $\biparamAIM$, $\biparamLEM$, and LCM as $\odotai$, $\odotle$, and $\odotlc$, respectively.

Recently, \citet{thanwerdas2019exploration} further extended $(\alpha,\beta)$-AIM to a three-parameter family of metrics by pulling AIM back through the matrix power function $\mathrm{P}_\theta(\cdot)$ and scaling the resulting metric by $\frac{1}{\theta^2}$, denoted as $\triparamAIM$.
The matrix power serves as a deformation, wherein $\triparamAIM$ encompasses $(\alpha,\beta)$-AIM with $\theta = 1$, and becomes $(\alpha,\beta)$-LEM as $\theta$ approaches 0 \citep{thanwerdas2019affine}.
Motivated by the deformation induced by the power function, we define the power-deformed metrics of $(\alpha,\beta)$-LEM and LCM by pulling the metrics back through $\pow_\theta$ and scaling the resulting metrics by $\frac{1}{\theta^2}$.
We denote these two metrics as $\triparamLEM$ and $\paramLCM$, respectively.
We have the following results w.r.t. the deformation.

\begin{parisproposition} [Deformation]\label{prop:spd_param_lem_lcm_deformation}
    \linktoproof{prop:spd_param_lem_lcm_deformation}
    $\triparamLEM$ is equal to $\biparamLEM$.
    $\theta$-LCM interpolates between $\tilde{g}$-LEM ($\theta=0$) and LCM ($\theta=1$). Here, given any $P \in \spd{n}$ and tangent vectors $V,W \in T_P\spd{n}$, $\tilde{g}$-LEM is defined as
    \begin{equation}
        \langle V,W \rangle_P = \tilde{g}(\mlog_{*,P}(V),\mlog_{*,P}(W)),
    \end{equation}
    where $\tilde{g}(V_1,V_2)=\frac{1}{2} \langle V_1, V_2 \rangle -\frac{1}{4} \langle \bbD(V_1), \bbD(V_2) \rangle$, $\bbD(V_i)$ is a diagonal matrix consisting of the diagonal elements of $V_i$, and $\mlog_{*,P}$ is the differential map at $P$.
\end{parisproposition}

As $\triparamLEM$ is equal to $\biparamLEM$, we focus on $\biparamLEM$, $\triparamAIM$, and $\paramLCM$ in the following. As a diffeomorphism, $\pow_{\theta}$ can also pull back the group operations $\odotai$ and $\odotlc$, denoted as $\odotpai$ and $\odotplc$.
We have the following proposition on invariance.

\begin{parisproposition} [Invariance]\label{prop:spd_param_invariance}
    \linktoproof{prop:spd_param_invariance}
    $\triparamAIM$ is left-invariant w.r.t. $\odotpai$, while $\paramLCM$ is bi-invariant w.r.t. $\odotplc$.
\end{parisproposition}

\subsubsection{SPD Right-Invariant Metrics}
\label{subsubsec:right_invariance}

AIM is left-invariant w.r.t. $\odotai$. We can also define a right-invariant metric w.r.t. $\odotai$ by definition \citep[Chapter~1.2]{do1992riemannian}:
\begin{equation}
    \gcri_{P} (V,W) = \left\langle \rtrans _{P_{\odotai}^{-1}*,P} (V) , \rtrans_{P_{\odotai}^{-1}*,P} (W)\right\rangle_{I},
\end{equation}
where $\rtrans_{(\cdot)}$ denotes Lie group right translation, $P_{\odotai}^{-1}$ is the inverse of $P$ under $\odotai$, and $\left\langle \cdot , \cdot \right\rangle_{I}$ denotes an arbitrary inner product on $T_I\spd{n}$. We set $\left\langle \cdot, \cdot \right\rangle_{I}$ to be the same as the AIM at $I$, \ie, $\left\langle \cdot, \cdot \right\rangle^{\alphabeta}$. We call this metric the Cholesky Right Invariant Metric (CRIM), as the group operation is defined by the matrix product of Cholesky factors \citep[Section~3.2]{thanwerdas2022theoretically}.

\begin{paristheorem} \label{thm:crim}
\linktoproof{thm:crim}
Given any SPD matrices $P, Q$ and a tangent vector $V \in T_P\spd{n}$, the Riemannian operators on $\{\spd{n},\gcri\}$ are
\begin{align}
    \gcri_{P}(V,V)
    &= \left( \left \| \symmetrize{ L(L^{-1} V L^{-\top})_{\frac{1}{2}} L^{-1}} \right \|^{\alphabeta} \right)^2, \\
    \label{eq:dist_crim}
    \dist(P,Q)
    &= \left \| \mlog\left( \widetilde{Q}^{-\frac{1}{2}} \widetilde{P} \widetilde{Q}^{-\frac{1}{2}}\right) \right \| ^{\alphabeta}, \\
    \label{eq:exp_crim}
    \rieexp_{P}(V)
    &= \left( \rieexp_{\widetilde{P}}^{\mathrm{AI}} \left(-\bar{V}  \right)\right)_{\odotai}^{-1},\\
    \label{eq:log_crim}
    \rielog_{P}(Q)
    &=-\symmetrize{L L^{\top} \left( L \widetilde{V}L^{\top} \right)_{\frac{1}{2}}^{\top}},
\end{align}
where $L$ is the Cholesky factor of $P=LL^\top$, $(\cdot)_{\odotai}^{-1}$ is the group inverse, $\widetilde{Q}$ and $\widetilde{P}$ are the group inverses of $P$ and $Q$, $\bar{V}=\symmetrize{ \left(L^{-1} V L^{-\top}\right)_{\frac{1}{2}} L^{-1} L^{-\top}}$, and $\widetilde{V}=\rielog^{\mathrm{AI}} _{\widetilde{P}} \left(\widetilde{Q} \right)$. Here, $\symmetrize{X}=X + X^\top, \forall X \in \bbR{n \times n}$ denotes symmetrization, and $(\cdot)_{\frac{1}{2}}$ denotes its inverse map, namely $(X)_{\frac{1}{2}}=\lfloor X \rfloor + \frac{1}{2}\bbX$.
\end{paristheorem}

\begin{pariscorollary} \label{cor:crim_geodesic}
    \linktoproof{cor:crim_geodesic}
    CRIM is geodesically complete, and the associated geodesic connecting SPD matrices $P$ and $Q$ is
    \begin{equation}
    \begin{aligned}
        \gamma _{(P,Q)} (t)
        &= \left\{ \gamma^{\mathrm{AI}}(t; \widetilde{P},\widetilde{Q}) \right\}_{\odotai} ^{-1}\\
        &= \left\{\widetilde{P}^{\frac{1}{2}} \left( \widetilde{P}^{-\frac{1}{2}} \widetilde{Q} \widetilde{P}^{-\frac{1}{2}} \right)^t \widetilde{P}^{\frac{1}{2}} \right\}_{\odotai} ^{-1},
     \end{aligned}
    \end{equation}
    where $\widetilde{P}=P^{-1}_{\odotai}$ and $\widetilde{Q} = Q^{-1}_{\odotai}$ are group inverses, with $\gamma^{\mathrm{AI}}$ as the geodesic under AIM.
\end{pariscorollary}
Similar to the discussion in \cref{subsubsec:spd_param_lie_groups}, we define $\theta$-CRIM by pulling CRIM back through the matrix power function $\mathrm{P}_\theta(\cdot)$ and scaling the resulting metric by $\frac{1}{\theta^2}$. As the pullback of CRIM, $\theta$-CRIM is right-invariant w.r.t. $\odotpai$ by definition.
\begin{parisproposition}\label{prop:spd_deformedcrim_invariance}
    $\theta$-CRIM is right-invariant w.r.t. $\odotpai$.
\end{parisproposition}

\begin{table}[!t]
  \centering
  \resizebox{0.99\linewidth}{!}{
    \begin{tabular}{c|c|cccc}
        \toprule
        \multicolumn{2}{c|}{Metric} & $\triparamAIM$  & $\biparamLEM$ & $\theta$-LCM & $\theta$-CRIM \\
        \midrule
        \multicolumn{2}{c|}{Invariance} & Left-invariance  & \multicolumn{2}{c}{Bi-invariance} & Right-invariance \\
        \midrule
        \multicolumn{2}{c|}{LieBN Type} & LieBN-Left  & \multicolumn{2}{c}{LieBN-Left = LieBN-Right} & LieBN-Right \\
        \midrule
        \multicolumn{2}{c|}{Pullback Map} &  $\pow_{\theta}$ &    $\mlog$ & $\pow_{\theta} \circ \clog$ & $\pow_{\theta}$ \\
        \midrule
        \multicolumn{2}{c|}{Codomain} &  $\{\spd{n},\odotai,\frac{1}{\theta^2}\gbiparamai\}$  & $\{\sym{n}, \langle \cdot , \cdot \rangle^{\alphabeta} \}$ & $\{\trilspace{n}, \frac{1}{\theta^2} \langle \cdot , \cdot \rangle\}$ & $\{\spd{n},\odotai,\frac{1}{\theta^2}\gcri\}$ \\
        \midrule
        \multirow{5}[15]{2.2cm}{\centering Riemannian and Lie group operators in the codomain} &  $\ltrans_{Q}(P)$ or $\rtrans_{Q}(P)$  &  $ KPK^\top $  &    $P+Q$  &  $P+Q$ & $LQL^\top$ \\
        \cmidrule(l){2-6}
        &   $\ltrans_{Q_{\odot}^{-1}}(P)$ or $\rtrans_{Q_{\odot}^{-1}}(P)$   &  $K^{-1}PK^{-\top}$  &  $P-Q$  &  $P-Q$ & $L^{-1}QL^{-\top}$  \\
        \cmidrule(l){2-6}
        &  $ \rieexp_{E} \left [ s \rielog_{E}(P) \right]$   &  $P^{s}$      &    $sP$     &    $sP$ & $\left( \left( P^{-1}_{\odotai} \right)^s \right)^{-1}_{\odotai}$     \\
        \cmidrule(l){2-6}
        &  FM    &    Karcher Flow     &    \makecell{Arithmetic \\ average}  &    \makecell{Arithmetic \\ average}  & Karcher Flow   \\
        \cmidrule(l){2-6}
        &  $\wfm(\{1-\gamma,\gamma\},\{P_1,P_2\})$  &  $P_1^{\frac{1}{2}}\left(P_1^{-\frac{1}{2}} P_2 P_1^{-\frac{1}{2}}\right)^\gamma P_1^{\frac{1}{2}}$ &  \makecell{Arithmetic \\ weighted average} &  \makecell{Arithmetic \\ weighted average} & $\left(\widetilde{P}_1^{\frac{1}{2}}\left(\widetilde{P}_1^{-\frac{1}{2}} \widetilde{P}_2 \widetilde{P}_1^{-\frac{1}{2}}\right)^\gamma \widetilde{P}_1^{\frac{1}{2}} \right)^{-1}_{\odotai}$\\
        \bottomrule
    \end{tabular}
    }
    \vspace{-2mm}
  \caption{Key operators in calculating SPD LieBN. The notations follow \cref{app:subsec:spd_geometries,thm:crim}.}
  \label{tab:ops_liebn_spd}
\end{table}

\begin{table}[!t]
    \centering
    \resizebox{0.99\linewidth}{!}{
    \begin{tabular}{cccccccc}
    \toprule
    Invariance & LieBN Type & $R^{-1}_{\odot}$ & $\ltrans _{R} (S)$ & $\rtrans_{R}(S)$ & $ \rieexp_{I} \left [ t \rielog_{I}(R) \right]$ & FM & $\wfm(\{1-\gamma,\gamma\},\{R,S\})$  \\
    \midrule
    Bi-invariance & LieBN-Left \& LieBN-Right & $R^{-1}$ & $RS$ &  $SR$ & $\mexp \left( t \mlog \left(R\right) \right)$  & \citep[Algorithm~1]{manton2004globally} & $R \mexp (\gamma \mlog(R^\top S))$ \\
    \bottomrule
    \end{tabular}
    }
    \vspace{-3mm}
    \caption{Key operators in calculating rotation LieBN. The notations follow \cref{tab:riem_rotation}.}
    \label{tab:liebn_rotations}
\end{table}

\subsubsection{Manifestations on SPD Manifolds}
\label{subsubsec:liebn_spd}

As discussed in \cref{subsubsec:spd_param_lie_groups,subsubsec:right_invariance}, there are four families of invariant metrics on the SPD Lie groups: (1) left-invariant $\triparamAIM$ w.r.t. $\odotai$; (2) bi-invariant $\biparamLEM$ w.r.t. $\odotle$ and $\theta$-LCM w.r.t. $\odotlc$; (3) right-invariant $\theta$-CRIM w.r.t. $\odotai$. Since all the above metrics are pullback metrics, the LieBN based on these metrics can be simplified and calculated in the codomain. We first show a general result on LieBN under the pullback metric. We denote \cref{alg:liebn} on the Lie group $\calM$ as
\begin{equation}
    \liebn(P_i;B,s,\epsilon,\gamma), \qquad P_i \in \{P_j\}_{j=1}^N \subset \calM.
\end{equation}
Then we can obtain the following theorem.

\begin{paristheorem} \label{thm:liebn_pullback}
    \linktoproof{thm:liebn_pullback}
    Given a Lie group $\calM_1$, a Lie group $\calM_2$ with an invariant metric $g^2$, and a map $f:\calM_1 \rightarrow \calM_2$ that is both a diffeomorphism and a Lie-group isomorphism, the map $f$ induces an invariant metric $g^1$ on $\calM_1$, denoted as $g^1=f^*g^2$.
    For a batch of activations $\{P_i\}_{i=1}^N$ in $\calM_1$, $\liebn^1(P_i;B,s,\epsilon,\gamma)$ in $\calM_1$ can be calculated in $\calM_2$ by the following process:
    \begin{align}
        &\text{Mapping data into } \calM_2: \bar P_i =f(P_i), \bar B=f(B),\\
        \label{eq:liebn_pm_codomain}
        &\text{Performing LieBN in } \calM_2 : \hat P_i = \liebn^2(\bar P_i;\bar B,s,\epsilon,\gamma),\\
        &\text{Mapping the resulting data back to } \calM_1: \tilde P_i =f^{-1}(\hat P_i),
    \end{align}
    where $\liebn^2$ is the LieBN on $\calM_2$.
\end{paristheorem}

Given a metric $g$ on $\spd{n}$, the power-deformed metric $\tilde{g}=\frac{1}{\theta^2}\pow_{\theta}^*g$ is equal to $\pow_{\theta}^*(\frac{1}{\theta^2} g)$.
\cref{thm:liebn_pullback} indicates that the LieBN under $\tilde{g}$ can be calculated by the LieBN under $\frac{1}{\theta^2} g$. Besides, as the Christoffel symbols remain the same under constant scaling, the LieBNs under $\frac{1}{\theta^2} g$ and $g$ only differ in the variance. We denote $\gbiparamai$ and $\gtriparamAI$ as the metric tensors of $\biparamAIM$ and $\triparamAIM$, respectively. Based on the above discussions, the computations of the LieBN under $\gtriparamAI$ are reduced to the LieBN under $\frac{1}{\theta^2}\gbiparamai$. Similarly, if $\gcri$ and $\gdefcri$ denote the metric tensors of CRIM and $\theta$-CRIM, respectively, then the LieBN under $\theta$-CRIM can be calculated by the one under $\frac{1}{\theta^2}\gcri$. Furthermore, as shown by \citet{chen2024spdmlr}, $\biparamLEM$ is a pullback metric from the Euclidean space $\sym{n}$ of symmetric matrices, while $\theta$-LCM is a pullback metric from the Euclidean space $\trilspace{n}$ of lower triangular matrices. As shown in \cref{prop:liebn_natural_extension_ebn}, the LieBN in the Euclidean space $\sym{n}$ or $\trilspace{n}$ is simplified to the standard Euclidean BN. Therefore, the LieBNs under $\biparamLEM$ and $\theta$-LCM can be calculated by the Euclidean BN over $\sym{n}$ and $\trilspace{n}$, respectively.

We denote the LieBN under left and right translations as LieBN-Left and LieBN-Right, respectively. Then, the LieBNs under $\triparamAIM$ and $\theta$-CRIM correspond to LieBN-Left and LieBN-Right, respectively. As $\odotlc$ and $\odotle$ are commutative, the LieBN-Left and LieBN-Right under $\biparamLEM$ and $\theta$-LCM are equivalent. We denote $P, Q, P_1$, and $P_2$ as points in the codomain, \ie, $\spd{n}$ with scaled CRIM for $\theta$-CRIM, $\spd{n}$ with scaled $\biparamAIM$ for $\triparamAIM$, $\sym{n}$ for $\biparamLEM$, and $\trilspace{n}$ for $\theta$-LCM, respectively. For CRIM, we denote $\widetilde{P}_i= (P_i)^{-1}_{\odotai}$ for $i=1,2$. We summarize all the necessary ingredients in \cref{tab:ops_liebn_spd} for calculating SPD LieBN. Note that for $\triparamAIM$, our scaling operation defined in \cref{eq:liebn_scaling} encompasses the scaling operation proposed by \citet[Equation~(9)]{kobler2022controlling} as a special case, when $(\theta,\alpha,\beta)=(1,1,0)$.

\subsection{LieBN on Rotation Matrices}
\label{subsec:liebn_son}

As the Riemannian metric on rotation matrices is bi-invariant, there are two instantiations of LieBN on this manifold, \ie, LieBN-Left based on the left translation and LieBN-Right based on the right translation. In particular, the scaling can be further simplified: $\rieexp_{I} \left(  t \rielog_{I} \left( R \right) \right) = \mexp \left( t \mlog \left(R\right) \right)$.
For $\so{3}$ in particular, the matrix exponential and logarithm can be efficiently calculated without matrix decomposition \citep[Section~3.2]{hartley2013rotation}. \cref{tab:liebn_rotations} presents the expressions of the required operators in \cref{alg:liebn}.

\subsection{LieBN on Full-Rank Correlation Matrices}
\label{subsec:liebn_cor}

\begin{table}[tbp]
  \centering
%   \resizebox{0.99\linewidth}{!}{
    \begin{tabular}{c|cccc}
        \toprule
        {Metric} & ECM  & LECM & OLM & LSM \\
        \midrule
        Invariance & \multicolumn{4}{c}{Bi-invariance} \\
        \midrule
        LieBN Type & \multicolumn{4}{c}{LieBN-Left = LieBN-Right} \\
        \midrule
        Pullback Map & $ \Theta$ & $\log \circ \Theta$ & $\offlog$ & $\logscaled$ \\
        \midrule
        Codomain & $\{\LTone{n}, \inner{\cdot}{\cdot} \}$ & $\{\LTzero{n}, \inner{\cdot}{\cdot} \}$ & $\{\hol{n}, \holinner{\cdot}{\cdot} \}$ & $\{\rzero{n}, \rzeroinner{\cdot}{\cdot} \}$\\
        \bottomrule
    \end{tabular}
    % }
    \vspace{-3mm}
  \caption{Summary of LieBN on full-rank correlation matrices. $\holinner{\cdot}{\cdot}$ and $\rzeroinner{\cdot}{\cdot}$ are permutation-invariant inner products, which are discussed in \cref{app:subsubsec:perm_metrics}.}
  \label{tab:ops_liebn_cor}
\end{table}

As discussed in \cref{subsec:cor_lie_groups}, all four correlation metrics are bi-invariant, and their associated Lie groups are commutative. Consequently, LieBN-Left is identical to LieBN-Right. Moreover, all four correlation metrics are pullback metrics from a simpler Euclidean space. Therefore, LieBN on the correlation manifold can be implemented as described in \cref{thm:liebn_pullback}: (1) map the correlation matrix into the prototype Euclidean space, (2) apply Euclidean BN, and (3) map back to the correlation matrix.

\mypara{Optimization.}
Finally, we discuss the optimization of the correlation-valued biasing parameter $B \in \cor{n}$. As shown by \citet[Section~4.1]{thanwerdas2022theoretically}, the correlation matrix can be identified by the product of hyperbolic spaces via the Cholesky decomposition. Given $C \in \cor{n}$, the $k$-th row of the Cholesky factor $L=\chol(C)$ is $\left(L_{k 1}, \ldots, L_{k, k-1}, L_{k k}, 0, \ldots, 0\right)$ with $L_{k k}>0$, which belongs to the open hemisphere model of hyperbolic space $\mathrm{HS}^{k-1} =\left\{x \in \mathbb{R}^k \mid\|x\|=1, x_k>0 \right\}$. Besides, the open hemisphere $\hs{n}$ is isometric to the Poincaré ball $\pball{n}=\left\{x \in \mathbb{R}^{n} \mid \|x\| < 1 \right\}$ by $\pi _{\hs{n} \rightarrow \pball{n}} ((x^\top, x_{n+1})^\top) = \frac{x}{1+x_{n+1}}$. Therefore, each correlation matrix can be parameterized with $n-1$ Poincaré vectors. Each Poincaré vector can be optimized using established Riemannian optimization methods \citep{becigneul2019riemannian}. The above process can be expressed as
\begin{equation}
    C {\mapsto}
    \begin{pmatrix}
    1 & 0 & \cdots & 0 \\
    L_{21} & L_{22} & \cdots & 0 \\
    \vdots & \vdots & \ddots & \vdots \\
    L_{n1} & L_{n2} & \cdots & L_{nn}
    \end{pmatrix}
    {\mapsto}
    \begin{pmatrix}
    x_1 \in \pball{1} \\
    \vdots \\
    x_{n-1} \in \pball{n-1}
    \end{pmatrix}.
\end{equation}

\section{Experiments}
\label{sec:experiments}
This section evaluates LieBN under nine invariant metrics on SPD, rotation, and correlation manifolds.

\subsection{Experiments of LieBN on the SPD Manifold}
\label{subsec:exp_spd}
Note that our LieBN layers are architecture-agnostic and can be applied to any existing SPD neural network. Following the previous work of \citet{huang2017riemannian,brooks2019riemannian,kobler2022spd}, we focus on two network architectures:
(1) SPDNet \citep{huang2017riemannian} for drone recognition on the Radar data set \citep{brooks2019riemannian}, and human action recognition on the HDM05 \citep{muller2007documentation} and FPHA \citep{garcia2018first} data sets;
(2) TSMNet \citep{kobler2022spd} for EEG classification on the Hinss2021 data set \citep{hinss_eegdata_2021}.
In the EEG application, TSMNet is endowed with SPD domain-specific momentum batch normalization (TSMNet+SPDDSMBN) \citep{kobler2022spd}, which is a domain-adaptation extension of the method proposed by \citet{kobler2022controlling}. For a fair comparison, we also implement a domain-specific momentum LieBN, referred to as DSMLieBN (detailed in \cref{app:sec:dsmliebn}). The backbone network architectures are represented as $\{d_0, d_1, \ldots, d_L\}$, where the dimension of the parameter in the $i$-th BiMap layer (\cref{app:sec:spdnet_tsmnet}) is $d_i \times d_{i-1}$. As $(\alpha,\beta)$ only affect variance calculation throughout LieBN, we simply set $(\alpha,\beta)=(1,0)$ and only tune the deformation factor $\theta$. For each family of LieBN or DSMLieBN, we report two representatives: the standard one induced by the standard metric ($\theta=1$), and the one induced by the deformed metric with proper $\theta$. \emph{If the standard one is already saturated, we only report the results of the standard one.} More details on implementation, data sets, and hyperparameters $(\theta,\alpha,\beta)$ are presented in \cref{app:subsec:exp_details_spd}.

\begin{table}[!t]
    \begin{subtable}[t]{\textwidth}
        \centering
        \resizebox{\linewidth}{!}{
        \begin{tabular}{c|cc|*{4}{>{\columncolor{HilightColor}}c}|>{\columncolor{HilightColor}}c}
            \toprule
            \multirow{3}[6]{*}{Acc} & \multirow{3}[6]{*}{SPDNet} & \multirow{3}[6]{*}{SPDNetBN} & \multicolumn{5}{c}{SPDNetLieBN} \\
            \cmidrule{4-8}          &       &       & \multicolumn{4}{c|}{$\theta=1$} & \multicolumn{1}{c}{Best $\theta$} \\
            \cmidrule{4-8}          &       &       & AIM-(1) & LEM-(1) & LCM-(1) & CRIM-(1) & LCM-(-0.5) \\
            \midrule
            \multicolumn{1}{c|}{Fit time (s)} & 0.60 & 1.19 & 1.16 & 0.96 & 0.80 & 1.41 & 1.06 \\
            Mean $\pm$ STD (\%) & $93.25 \pm 1.10$ & $94.85 \pm 0.99$ & $\textcolor{red}{\boldsymbol{95.47 \pm 0.90}}$ & $94.89 \pm 1.04$ & $93.52 \pm 1.07$ & $94.35 \pm 0.68$ & $94.80 \pm 0.71$ \\
            Max (\%) & 94.4  & 96.13 & 96.27 & \textcolor{red}{\textbf{96.8}} & 95.2  & 95.6  & 95.73 \\
            \bottomrule
        \end{tabular}%
        }
        \caption{Radar data set.}
        \label{tab:results_radar}
    \end{subtable}
    \vspace{2mm}

    \begin{subtable}[t]{\linewidth}
        \centering
        \resizebox{\linewidth}{!}{
        \begin{tabular}{c|c|c|*{4}{>{\columncolor{HilightColor}}c}|*{3}{>{\columncolor{HilightColor}}c}}
            \toprule
            \multicolumn{1}{c|}{\multirow{3}[6]{*}{Acc}} & \multirow{3}[6]{*}{SPDNet} & \multirow{3}[6]{*}{SPDNetBN} & \multicolumn{7}{c}{SPDNetLieBN} \\
            \cmidrule{4-10}    \multicolumn{1}{c|}{} &       &       & \multicolumn{4}{c|}{$\theta=1$} & \multicolumn{3}{c}{Best $\theta$} \\
            \cmidrule{4-10}    \multicolumn{1}{c|}{} &       &       & AIM-(1) & LEM-(1) & LCM-(1) & CRIM-(1) & AIM-(1.5) & LCM-(0.5) & CRIM-(0.5) \\
            \midrule
            Fit time (s) & 0.41 & 0.78 & 0.97 & 0.76 & 0.55 & 1.19 & 1.27 & 0.71 & 1.36 \\
            Mean $\pm$ STD (\%) & $59.13 \pm 0.67$ & $66.72 \pm 0.52$ & $67.79 \pm 0.65$ & $65.05 \pm 0.63$ & $66.68 \pm 0.71$ & $63.25 \pm 0.88$ & $68.16 \pm 0.68$ & $\textcolor{red}{\boldsymbol{70.84 \pm 0.92}}$ & $65.76 \pm 0.54$ \\
            Max (\%) & 60.34 & 67.66 & 68.75 & 66.05 & 68.52 & 64.94 & 69.25 & \textcolor{red}{\textbf{72.27}} & 66.96 \\
            \bottomrule
        \end{tabular}%
        }
        \caption{HDM05 data set.}
        \label{tab:results_HDM05}
     \end{subtable}

     \vspace{2mm}

    \begin{subtable}[t]{\linewidth}
        \centering
        \resizebox{\linewidth}{!}{
        \begin{tabular}{c|c|c|*{4}{>{\columncolor{HilightColor}}c}|*{3}{>{\columncolor{HilightColor}}c}}
            \toprule
            \multirow{3}[6]{*}{Acc} & \multirow{3}[6]{*}{SPDNet} & \multirow{3}[6]{*}{SPDNetBN} & \multicolumn{7}{c}{SPDNetLieBN} \\
            \cmidrule{4-10}          &       &       & \multicolumn{4}{c|}{$\theta=1$} & \multicolumn{3}{c}{Best $\theta$} \\
            \cmidrule{4-10}          &       &       & AIM-(1) & LEM-(1) & LCM-(1) & CRIM-(1) & AIM-(1.5) & LCM-(0.5) & CRIM-(-0.5) \\
            \midrule
            Fit time (s) & 0.26 & 0.54 & 0.71 & 0.50 & 0.36 & 0.87 & 0.97 & 0.49 & 1.16 \\
            Mean $\pm$ STD (\%) & $85.59 \pm 0.72$ & $89.33 \pm 0.49$ & $89.70 \pm 0.51$ & $86.56 \pm 0.79$ & $77.64 \pm 1.00$ & $84.65 \pm 1.20$ & $\textcolor{red}{\boldsymbol{90.39 \pm 0.66}}$ & $86.33 \pm 0.43$ & $86.40 \pm 0.57$ \\
            Max (\%) & 86    & 90.17 & 90.5  & 87.83 & 79    & 86.67 & \textcolor{red}{\textbf{92.17}} & 87    & 87.17 \\
            \bottomrule
        \end{tabular}%
        }
        \caption{FPHA data set.}
        \label{tab:results_FPHA}
     \end{subtable}
     \vspace{-3mm}
    \caption{Accuracy results averaged over 10 folds for SPDNet with and without SPDBN or LieBN on the Radar, HDM05, and FPHA data sets. Fit time is the average of the five fastest complete training epochs among ten controlled timing epochs. If the LieBN under the standard metric ($\theta=1$) is not saturated, the rightmost columns report the deformed LieBN.}
    \label{tab:results_spdnet}
\end{table}

\mypara{Application to SPDNet.}
As SPDNet is a canonical SPD network, we apply our LieBN to SPDNet on the Radar, HDM05, and FPHA data sets. Additionally, we compare our method with SPDNetBN, which applies the SPDBN in \cref{eq:spdnetbn_centering,eq:spdnetbn_biasing} to SPDNet. Following \citet{brooks2019riemannian,chen2024adaptive}, we use the architectures of $\{20,16,8\}$, $\{93,30\}$, and $\{63,33\}$ for the Radar, HDM05, and FPHA data sets, respectively. The 10-fold average results and controlled fit times are summarized in \cref{tab:results_spdnet}. We have three key observations regarding the choice of metrics, deformation, and training efficiency.
\begin{itemize}
    \item
    \mypara{The choice of metrics.} The metric that yields the most effective LieBN layer differs for each data set. Specifically, the optimal LieBN layers on these three data sets are the ones induced by AIM-(1), LCM-(0.5), and AIM-(1.5), respectively, \textbf{which improve the performance of SPDNet by 2.22\%, 11.71\%, and 4.8\%}. Additionally, although the LCM-based LieBN performs worse than other LieBN variants on the Radar and FPHA data sets, it exhibits the best performance on the HDM05 data set. These observations highlight the advantage of LieBN's generality.
    \item
    \mypara{The effect of deformation.}
    Deformation patterns also vary across data sets. Firstly, the standard AIM and CRIM are already saturated on the Radar data set. Secondly, the appropriate deformation $\theta$ can further enhance the performance of LieBN. Notably, even though the LieBNs induced by LCM-(1) and CRIM-(1) impede the learning of SPDNet on the FPHA data set, they can improve the performance under an appropriate deformation $\theta$. These findings highlight the efficacy of the deforming geometry on the SPD manifold.
    \item
    \mypara{Efficiency.}
    Although our LieBN involves additional computations on variance compared with SPDNetBN, our LieBN achieves efficiency comparable to or even better than that of SPDNetBN. In particular, the LieBN induced by standard LEM or LCM exhibits better efficiency than SPDNetBN. Even with deformation, the LCM-based LieBN is still comparable with SPDNetBN in terms of efficiency. This phenomenon could be attributed to the fast and simple computation of LCM and LEM.
\end{itemize}

\begin{table}[tbp]
    \centering
    \begin{subtable}[t]{0.49\linewidth}
        \centering
        \resizebox{\linewidth}{!}{
        \begin{tabular}{c|c|cc}
            \toprule
            \multicolumn{2}{c|}{Method} & Fit time (s) & Mean $\pm$ STD (\%) \\
            \midrule
            \multicolumn{2}{c|}{SPDDSMBN} & 0.16  & $54.12 \pm 9.87$ \\
            \midrule
            \multirow{5}[4]{*}{DSMLieBN} & \cellcolor{HilightColor} AIM-(1) & \cellcolor{HilightColor} 0.16  & \cellcolor{HilightColor} $\textcolor{red}{\boldsymbol{55.10 \pm 7.61}}$ \\
            & \cellcolor{HilightColor} LEM-(1) & \cellcolor{HilightColor} 0.13  & \cellcolor{HilightColor} $54.95 \pm 10.09$ \\
            & \cellcolor{HilightColor} LCM-(1) & \cellcolor{HilightColor} 0.10  & \cellcolor{HilightColor} $51.54 \pm 6.88$ \\
            & \cellcolor{HilightColor} CRIM-(1) & \cellcolor{HilightColor} 0.29  & \cellcolor{HilightColor} $51.86 \pm 9.21$ \\
            \cmidrule{2-4}          & \cellcolor{HilightColor} LCM-(0.5) & \cellcolor{HilightColor} 0.15  & \cellcolor{HilightColor} $53.11 \pm 5.65$ \\
            \bottomrule
        \end{tabular}%
        }
        \caption{Inter-session classification.}
        \label{tab:results_inter_session}
    \end{subtable}
    \hfill
    \begin{subtable}[t]{0.46\linewidth}
        \centering
        \resizebox{\linewidth}{!}{
        \begin{tabular}{c|c|cc}
            \toprule
            \multicolumn{2}{c|}{Method} & Fit time (s) & Mean $\pm$ STD (\%) \\
            \midrule
            \multicolumn{2}{c|}{SPDDSMBN} & 7.74  & $50.10 \pm 8.08$ \\
            \midrule
            \multirow{6}[4]{*}{DSMLieBN} & \cellcolor{HilightColor} AIM-(1) & \cellcolor{HilightColor} 6.94  & \cellcolor{HilightColor} $50.04 \pm 8.01$ \\
            & \cellcolor{HilightColor} LEM-(1) & \cellcolor{HilightColor} 4.71  & \cellcolor{HilightColor} $50.95 \pm 6.40$ \\
            & \cellcolor{HilightColor} LCM-(1) & \cellcolor{HilightColor} 3.59  & \cellcolor{HilightColor} $51.86 \pm 4.53$ \\
            & \cellcolor{HilightColor} CRIM-(1) & \cellcolor{HilightColor} 16.35 & \cellcolor{HilightColor} $50.71 \pm 8.1$ \\
            \cmidrule{2-4}          & \cellcolor{HilightColor} CRIM-(1.5) & \cellcolor{HilightColor} 19.51 & \cellcolor{HilightColor} $51.34 \pm 5.82$ \\
            & \cellcolor{HilightColor} AIM-(-0.5) & \cellcolor{HilightColor} 8.71  & \cellcolor{HilightColor} $\textcolor{red}{\boldsymbol{53.97 \pm 8.78}}$ \\
            \bottomrule
        \end{tabular}%
        }
        \caption{Inter-subject classification.}
        \label{tab:results_inter_subject}
     \end{subtable}
     \vspace{-3mm}
    \caption{Cross-validation results of TSMNet with SPDDSMBN and DSMLieBN on the Hinss data set. If the DSMLieBN under the standard metric ($\theta=1$) is not saturated, the bottom rows report deformed DSMLieBN.}
    \label{tab:results_TSMNet}
\end{table}

\mypara{Application to EEG classification.}
We apply our method to TSMNet under two scenarios: inter-session and inter-subject. Following \citet{kobler2022spd}, we adopt the architecture of $\{40,20\}$. Compared to SPDDSMBN, DSMLieBN-AIM obtains the highest average scores of 55.10\% and 53.97\% in these two scenarios, \emph{outperforming SPDDSMBN by 0.98\% and 3.87\%, respectively}. In the inter-subject scenario, the efficiency advantage of our LieBN over SPDDSMBN is more pronounced.
Specifically, both the LEM- and LCM-based DSMLieBN achieve performance similar to or better than that of SPDDSMBN, while requiring considerably less training time.
For example, DSMLieBN-LCM-(1) achieves better results with only half the training time of SPDDSMBN on inter-subject tasks. Interestingly, under the standard AIM, the sole difference between SPDDSMBN and our DSMLieBN is the way they perform the centering and biasing operations. SPDDSMBN applies the matrix inverse square root and matrix square root to fulfill centering and biasing, while AIM-induced LieBN uses a more efficient Cholesky decomposition. As such, the DSMLieBN induced by the standard AIM is more efficient than SPDDSMBN, particularly on the inter-subject task. On the other hand, the CRIM-based LieBN is less efficient due to the relatively complex Riemannian computation of this metric.

\begin{figure}[t]
\centering
\includegraphics[width=\linewidth,trim={1cm 10cm 1cm 3cm}]{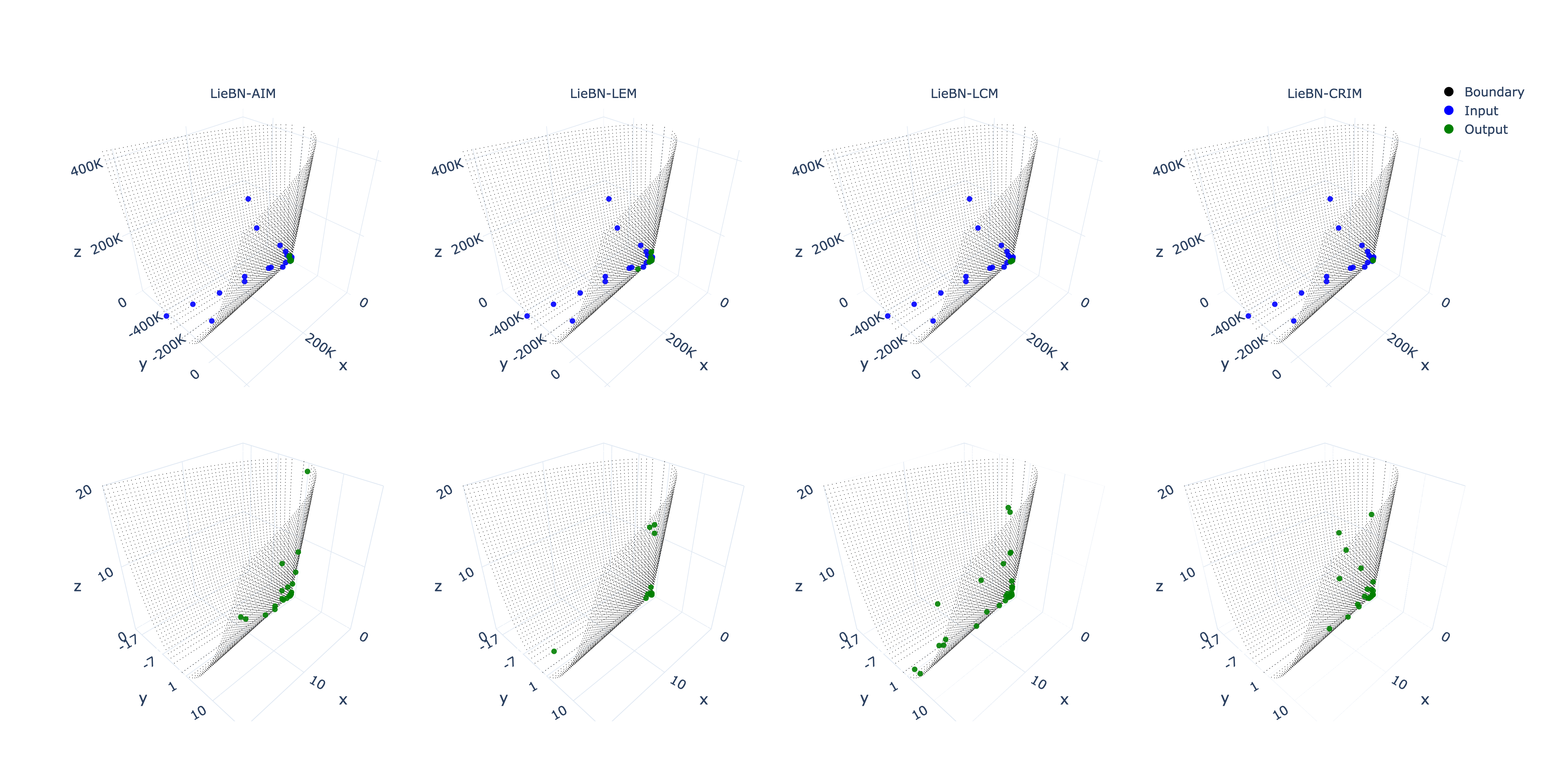}
% trim={left bottom right top}
\caption{Visualization of input and output $30 \times 30$ SPD matrices in LieBN using $2 \times 2$ Riemannian t-SNE embeddings. The first row shows the input and output under different metrics. Due to the significant difference in magnitude between the t-SNE embeddings of LieBN's input and output, the second row separately visualizes the LieBN output (at a smaller scale).}
\label{fig:visualization_hdm05}
\vspace{-4mm}
\end{figure}

\mypara{Visualization.}
We randomly select 50 samples and visualize the input and output of LieBN on the HDM05 data set. Using Riemannian t-SNE \citep{surrel2025geometryaware}, we map the $30 \times 30$ SPD matrices to $2 \times 2$ low-dimensional representations. As shown in \cref{fig:visualization_hdm05}, LieBN effectively normalizes the data distribution. Specifically, the input t-SNE embeddings are largely scattered, and their coordinates have magnitudes up to $4 \times 10^5$, whereas the coordinates of the output embeddings mostly lie within $[-20, 20]$.

\subsubsection{The Effect of \texorpdfstring{$\beta$}{beta}}

Recalling \cref{eq:oim_sym}, $\beta$ controls the importance of the trace term relative to the inner-product term. Therefore, we set the candidate values of $\beta$ to $\{1,\nicefrac{1}{n},\nicefrac{1}{n^2}, 0, -\nicefrac{1}{n} + \epsilon, -\nicefrac{1}{n^2}\}$, where $n$ is the input dimension of LieBN, and $\epsilon$ is a small positive scalar to ensure $\orth{n}$-invariance, \ie, $\alphabeta \in \bfst$. $\nicefrac{1}{n^2}$ and $\nicefrac{1}{n}$ mean averaging the trace in \cref{eq:oim_sym}, while the sign of $\beta$ denotes suppressing (-), enhancing (+), or neutralizing (0) the trace. We focus on AIM-based LieBN on the HDM05 data set. We set $\theta=1.5$, as it is the best deformation factor in this scenario. Other network settings remain unchanged. The 10-fold average results are presented in \cref{tab:beta_aim_liebn}. Note that in this setting, $n=30$. As expected, $\beta$ has minor effects on our LieBN.

\begin{table}[tbp]
    \centering
    \resizebox{\linewidth}{!}{
    \begin{tabular}{ccccccc}
    \toprule
    $\beta$  & $\nicefrac{-1}{30^2}$ & -0.03 & $\nicefrac{1}{30^2}$ & $\nicefrac{1}{30}$ & 1     & 0 \\
    \midrule
    Mean $\pm$ STD (\%) & $68.18 \pm 0.86$ & $68.12 \pm 0.74$ & $68.20 \pm 0.85$ & $68.18 \pm 0.85$ & $68.16 \pm 0.80$ & $68.16 \pm 0.68$ \\
    \bottomrule
    \end{tabular}
    }
    \caption{The effect of different values of $\beta$ for AIM-based LieBN on the HDM05 data set.}
    \label{tab:beta_aim_liebn}
\end{table}

\FloatBarrier
\subsection{Experiments of LieBN on Rotation Matrices}
In this subsection, we implement our LieBN on the special orthogonal groups $\so{n}$, whose elements are rotation matrices. As the Riemannian metrics on these groups are bi-invariant, there are two instantiations of our LieBN for these groups: LieBN-Left based on the left translation and LieBN-Right based on the right translation. We apply our LieBN to the classic LieNet backbone \citep{huang2017deep}, where the latent space is the special orthogonal group. Following \citet{huang2017deep}, we use three action recognition data sets: G3D \citep{bloom2012g3d}, HDM05 \citep{muller2007documentation}, and NTU60 \citep{shahroudy2016ntu}. We denote the LieNet models with our LieBN-Left and LieBN-Right by LieNetLieBN-Left and LieNetLieBN-Right, respectively. More implementation details are presented in \cref{app:subsec:impl_details_so3}.

\begin{figure}[t]
\centering
\includegraphics[width=0.99\linewidth,trim={0cm 0cm 0cm 0cm}]{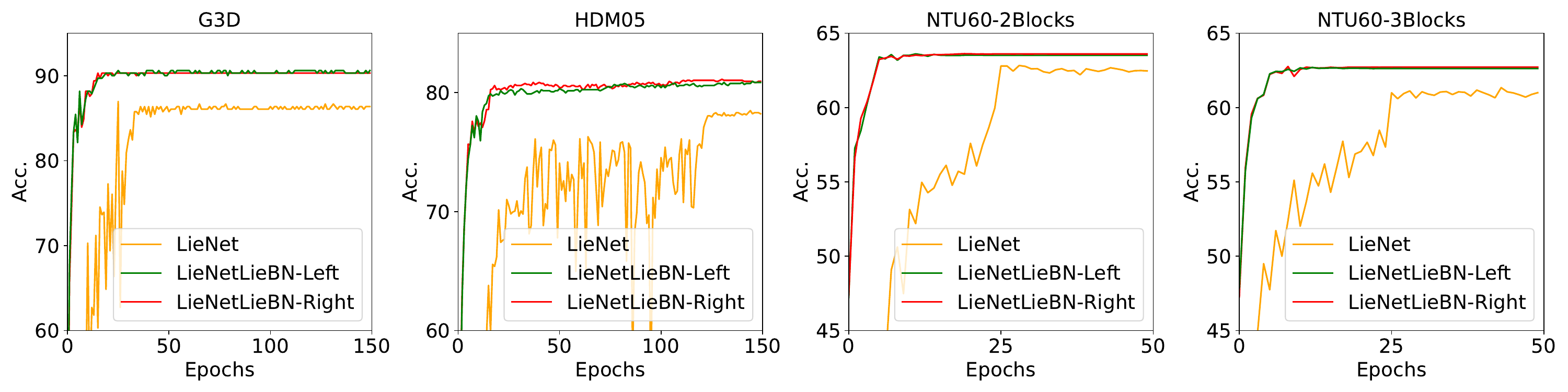}
% trim={left bottom right top}
\caption{Test accuracy curves corresponding to \cref{tab:results_liebn_so3}.}
\label{fig:acc_liebn_so3}
\vspace{-3mm}
\end{figure}

\begin{table}[tbp]
    \centering
    \footnotesize
    \setlength{\tabcolsep}{2pt}
    % \resizebox{0.99\linewidth}{!}{
    \begin{tabular}{c|cc|cc|cc}
        \toprule
        \multirow{2}[4]{*}{Method} & \multicolumn{2}{c|}{G3D} & \multicolumn{2}{c|}{HDM05} & \multicolumn{2}{c}{NTU60 (Acc.)} \\
        \cmidrule{2-7}          & Mean $\pm$ STD (\%) & Max (\%) & Mean $\pm$ STD (\%) & Max (\%) & 2-block (\%) & 3-block (\%) \\
        \midrule
        LieNet & $87.91 \pm 0.90$ & 89.73 & $76.92 \pm 1.27$ & 79.11 & 62.4  & 60.91 \\
        \midrule
        \rowcolor{HilightColor} LieNetLieBN-Left & $\textcolor{red}{\boldsymbol{88.88 \pm 1.62}}$ & \textcolor{red}{\textbf{90.67}} & $78.89 \pm 1.07$ & \textcolor{red}{\textbf{80.88}} & 63.51 & 62.62 \\
        \rowcolor{HilightColor} LieNetLieBN-Right & $88.12 \pm 1.12$ & 90.3  & $\textcolor{red}{\boldsymbol{79.39 \pm 1.13}}$ & 80.67 & \textcolor{red}{\textbf{63.6}} & \textcolor{red}{\textbf{62.72}} \\
        \bottomrule
    \end{tabular}
    % }
    % \vspace{-2mm}
    \caption{Results of LieNet with or without rotation LieBN.}
    \label{tab:results_liebn_so3}
\end{table}%

\mypara{Results.}
We conduct 10-fold experiments on the G3D and HDM05 data sets under the suggested 3-block\footnote{Each block consists of a RotMap layer followed by a RotPooling layer. For more details, please refer to \citet{huang2017deep}.} and 2-block architectures, respectively. On the NTU60 data set, we validate LieBN under the 2-block and 3-block settings. The results are presented in \cref{tab:results_liebn_so3}. Due to differences between our PyTorch implementation and the original MATLAB implementation, our reimplemented LieNet performs slightly differently from the results reported in \citet{huang2017deep}. However, we still observe a clear improvement when applying our LieBN to the vanilla LieNet backbone. Additionally, LieBN-Right performs slightly better than LieBN-Left. Although the effects of left and right translations on the sample statistics under the bi-invariant metric are identical, their transformations on each sample differ, as illustrated in \cref{fig:illustration}. This difference could slightly affect the network performance. The specific optimal choice of left or right translations depends on the data set's characteristics.

\mypara{Training dynamics.}
\cref{fig:acc_liebn_so3} presents the test accuracy curves. We have the following additional observations, which can be attributed to the mitigated covariate shift by our LieBN, as our LieBN can effectively normalize the sample statistics.
\mypara{Accelerated convergence.}
LieBN significantly accelerates the convergence of LieNet. Specifically, on the NTU60 data set---the largest data set involved---LieNet with LieBN converges by the 5th epoch, whereas the vanilla LieNet does not converge until the 25th epoch. A similar phenomenon can also be observed on the HDM05 data set.
\mypara{More stable performance.}
LieBN enhances the stability of network training. Especially on the HDM05 and G3D data sets, the initial training fluctuations are greatly mitigated by our LieBN.

\subsection{Experiments of LieBN on Correlation Matrices}
\label{subsec:exp_liebn_cor}

Our experiments focus on the SPDNet backbone using the FPHA and HDM05 data sets. LieBN-Cor is applied before the final classification layer. Specifically, SPD features are first activated by the power function, then mapped into correlation matrices via $\coropt(\cdot)$, and finally processed by LieBN-Cor. More details can be found in \cref{app:subsec:impl_details_cor}.

\begin{table}[tbp]
  \centering
%
%   \resizebox{0.99\linewidth}{!}{
    \begin{tabular}{c|c|*{4}{>{\columncolor{HilightColor}}c}}
    \toprule
    & \multicolumn{5}{c}{Mean $\pm$ STD (\%)} \\
    \cmidrule{2-6}
    \multirow{2}[4]{*}{Data Set} & \multirow{2}[4]{*}{SPDNet} & \multicolumn{4}{c}{SPDNetLieBN-Cor} \\
    \cmidrule{3-6}          &       & ECM   & LECM  & OLM   & LSM \\
    \midrule
    HDM05 & $59.13 \pm 0.67$ & $\textcolor{red}{\boldsymbol{65.37 \pm 1.07}}$ & $61.35 \pm 0.34$ & $60.33 \pm 0.12$ & $60.00 \pm 0.27$ \\
    FPHA  & $85.59 \pm 0.72$ & $\textcolor{red}{\boldsymbol{87.20 \pm 0.12}}$ & $87.03 \pm 0.32$ & $86.80 \pm 0.12$ & $86.77 \pm 0.29$ \\
    \bottomrule
    \end{tabular}%
    % }
  \caption{Results of SPDNet with or without correlation LieBN under different invariant metrics.}
  \label{tab:results_liebn_cor}
\end{table}%

\mypara{Results.}
The 5-fold average results are presented in \cref{tab:results_liebn_cor}. Although LieBN-Cor is not specifically designed for SPD networks, it still improves SPDNet's performance, demonstrating its effectiveness. Among the four invariant metrics, ECM achieves the best performance. As expected, LieBN-SPD outperforms LieBN-Cor when applied to SPDNet because SPDNet is tailored for SPD matrices. This comparison does not undermine the validity of LieBN-Cor. The consistent improvement over vanilla SPDNet highlights the potential of applying LieBN-Cor to correlation manifolds.

\begin{table}[tbp]
  \centering
%   \resizebox{0.99\linewidth}{!}{
    \begin{tabular}{c|cccc}
    \toprule
    & \multicolumn{4}{c}{Mean $\pm$ STD (\%)} \\
    \cmidrule{2-5}
    \diagbox{Optim}{Metric} & ECM   & LECM  & OLM   & LSM \\
    \midrule
    Trivialization & $63.17 \pm 1.32$ & $58.84 \pm 0.51$ & $59.84 \pm 0.45$ & $53.22 \pm 1.62$ \\
    \midrule
    \rowcolor{HilightColor} Riemannian & $\textcolor{red}{\boldsymbol{65.37 \pm 1.07}}$ & $\textcolor{red}{\boldsymbol{61.35 \pm 0.34}}$ & $\textcolor{red}{\boldsymbol{60.33 \pm 0.12}}$ & $\textcolor{red}{\boldsymbol{60.00 \pm 0.27}}$ \\
    \bottomrule
    \end{tabular}
    % }
  \caption{Ablations on optimizing the correlation parameter in LieBN under different metrics.}
  \label{tab:ablation_liebn_cor_optimization}
\end{table}%

\mypara{Ablations.}
As discussed in \cref{subsec:liebn_cor}, the correlation-valued biasing parameter $B \in \cor{n}$ is optimized via Riemannian optimization over multiple Poincar\'e vectors. Alternatively, trivialization tricks \citep{lezcano2019trivializations} can be employed. Specifically, \cref{thm:liebn_pullback} suggests that we can instead set $\bar{B} = f(B) \in V$ as the parameter, where, for each metric, $f$ denotes the corresponding isometry and $V$ its prototype space. \cref{tab:ablation_liebn_cor_optimization} presents a 5-fold comparison, demonstrating the superiority of our Poincar\'e parameterization.

\subsection{Discussion}

This discussion analyzes two practical aspects of LieBN: the numerical budget used to estimate the Fr\'echet batch mean and the distinction between normalization on a Lie group and normalization in its Lie algebra.

\subsubsection{Ablation Study on the Number of Fr\'echet Mean Iterations}
\label{subsubsec:fm_iteration_sensitivity}

\mypara{Implementation details.} LieBN computes the Fr\'echet batch mean using iterative procedures for both SPD matrices under AIM and CRIM and rotation matrices. We study the maximum number of iterations $K \in \{1,2,5,10,20\}$ while keeping all other training details unchanged. For SPD matrices, we retain the best $\theta$ values from the main experiments: $(1,1.5,1.5)$ for AIM and $(1,0.5,-0.5)$ for CRIM on Radar, HDM05, and FPHA, respectively. The solver may terminate before reaching $K$ when its convergence criterion is satisfied.

\begin{table}[t]
    \centering
    \resizebox{\linewidth}{!}{%
    \begin{tabular}{cc|cc|cc|cc}
        \toprule
        \multirow{2}{*}{Metric} & \multirow{2}{*}{$K$} & \multicolumn{2}{c|}{Radar} & \multicolumn{2}{c|}{HDM05} & \multicolumn{2}{c}{FPHA} \\
        \cmidrule{3-8}
        & & Mean $\pm$ STD (\%) & Fit time (s) & Mean $\pm$ STD (\%) & Fit time (s) & Mean $\pm$ STD (\%) & Fit time (s) \\
        \midrule
        \rowcolor{HilightColor} \multirow{5}{*}{AIM} & 1 & $95.47 \pm 0.90$ & 1.16 & $\textcolor{red}{\boldsymbol{68.16 \pm 0.68}}$ & 1.27 & $\textcolor{red}{\boldsymbol{90.39 \pm 0.66}}$ & 0.97 \\
        & 2 & $95.03 \pm 0.76$ & 1.28 & $62.30 \pm 0.90$ & 1.40 & $89.80 \pm 0.50$ & 1.09 \\
        & 5 & $94.69 \pm 0.67$ & 1.27 & $56.91 \pm 0.88$ & 1.81 & $89.87 \pm 0.41$ & 1.49 \\
        & 10 & $95.03 \pm 0.91$ & 1.27 & $53.74 \pm 1.05$ & 2.49 & $89.95 \pm 0.62$ & 2.10 \\
        & 20 & $\textcolor{red}{\boldsymbol{95.59 \pm 0.65}}$ & 1.28 & $52.88 \pm 0.85$ & 3.89 & $89.78 \pm 0.43$ & 3.38 \\
        \midrule
        \rowcolor{HilightColor} \multirow{5}{*}{CRIM} & 1 & $94.35 \pm 0.68$ & 1.41 & $\textcolor{red}{\boldsymbol{65.76 \pm 0.54}}$ & 1.36 & $86.40 \pm 0.57$ & 1.16 \\
        & 2 & $94.27 \pm 0.79$ & 1.62 & $64.55 \pm 0.75$ & 1.58 & $\textcolor{red}{\boldsymbol{86.75 \pm 0.90}}$ & 1.33 \\
        & 5 & $94.32 \pm 0.59$ & 1.86 & $63.71 \pm 0.91$ & 2.23 & $86.73 \pm 0.92$ & 1.70 \\
        & 10 & $\textcolor{red}{\boldsymbol{94.45 \pm 0.77}}$ & 1.83 & $63.84 \pm 0.77$ & 3.00 & $86.73 \pm 0.92$ & 1.70 \\
        & 20 & $94.04 \pm 0.75$ & 1.84 & $63.77 \pm 0.84$ & 3.21 & $86.73 \pm 0.92$ & 1.72 \\
        \bottomrule
    \end{tabular}
    }
    \caption{Ablation study of the number of Fréchet mean iterations for SPDNetLieBN. The shaded rows indicate $K=1$, the setting used in the main experiments.}
    \label{tab:fm_iteration_spd}
\end{table}

\begin{table}[t]
    \centering
    \resizebox{\linewidth}{!}{%
    \begin{tabular}{c|c|cc|cc|cc}
        \toprule
        \multirow{2}{*}{Method} & \multirow{2}{*}{$K$} & \multicolumn{2}{c|}{HDM05} & \multicolumn{2}{c|}{NTU60 (2-block)} & \multicolumn{2}{c}{NTU60 (3-block)} \\
        \cmidrule{3-8}
        & & Mean $\pm$ STD (\%) & Fit time (s) & Acc. (\%) & Fit time (s) & Acc. (\%) & Fit time (s) \\
        \midrule
        \rowcolor{HilightColor} \multirow{5}{*}{LieNetLieBN-Left} & 1 & $\textcolor{red}{\boldsymbol{78.89 \pm 1.07}}$ & 10.92 & 63.51 & 207.59 & 62.62 & 206.70 \\
        & 2 & $78.80 \pm 0.72$ & 11.10 & \textcolor{red}{\textbf{63.57}} & 211.25 & \textcolor{red}{\textbf{63.02}} & 208.18 \\
        & 5 & $78.80 \pm 0.65$ & 11.36 & 63.30 & 216.52 & 62.85 & 209.12 \\
        & 10 & $78.75 \pm 0.65$ & 11.62 & 63.25 & 220.17 & 62.79 & 209.83 \\
        & 20 & $78.77 \pm 0.65$ & 12.01 & 63.27 & 244.96 & 62.65 & 212.45 \\
        \midrule
        \rowcolor{HilightColor} \multirow{5}{*}{LieNetLieBN-Right} & 1 & $\textcolor{red}{\boldsymbol{79.39 \pm 1.13}}$ & 11.01 & \textcolor{red}{\textbf{63.60}} & 209.03 & \textcolor{red}{\textbf{62.72}} & 205.13 \\
        & 2 & $78.74 \pm 0.72$ & 11.09 & 63.29 & 214.24 & 62.30 & 204.89 \\
        & 5 & $78.91 \pm 0.75$ & 11.29 & 63.32 & 213.80 & 62.47 & 209.53 \\
        & 10 & $78.86 \pm 0.68$ & 11.65 & 63.22 & 219.62 & 62.37 & 213.83 \\
        & 20 & $78.85 \pm 0.61$ & 11.95 & 63.10 & 241.33 & 62.29 & 217.07 \\
        \bottomrule
    \end{tabular}%
    }
    \caption{Ablation study of the number of Fréchet mean iterations for LieNetLieBN. The shaded rows indicate $K=1$, the setting used in the main experiments.}
    \label{tab:fm_iteration_rot}
\end{table}

\mypara{Results.} \cref{tab:fm_iteration_spd,tab:fm_iteration_rot} summarize the results on SPD and rotation matrices. They show that one Fr\'echet mean iteration is generally sufficient to achieve competitive accuracy, which is also the setting used in our original LieBN experiments. Increasing the maximum iteration budget provides no systematic accuracy improvement and generally increases training time. This is reasonable because neural networks are themselves approximate models, so a more accurate estimate of an intermediate statistic, such as the Fr\'echet batch mean, does not necessarily improve end-to-end performance. In particular, increasing $K$ is detrimental to AIM on HDM05, where the accuracy decreases from $68.16\%$ at $K=1$ to $52.88\%$ at $K=20$. These observations support using $K=1$ as the default, providing a favorable balance between accuracy and efficiency.

\begin{figure}[t]
\centering
\includegraphics[width=0.8\linewidth,trim={0cm 0cm 0cm 0cm}]{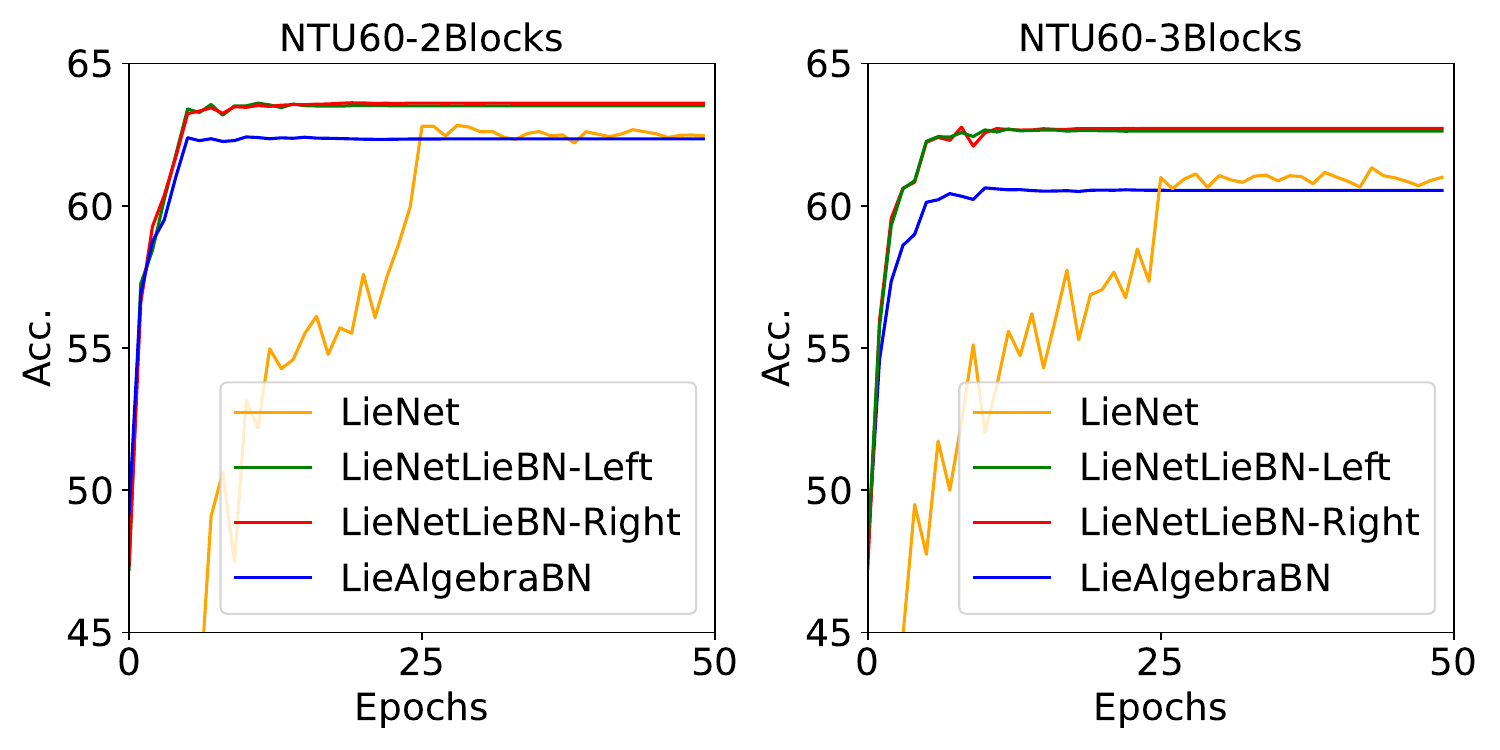}
% trim={left bottom right top}
\caption{Comparisons of LieBN and LieAlgebraBN on the LieNet backbone.}
\label{fig:acc_liebn_vs_liealgebrabn_so3}
\vspace{-3mm}
\end{figure}
\subsubsection{Lie Algebra Normalization vs. Lie Group Normalization}

\mypara{Formulation.}
A natural and direct way to construct BN on a Lie group is to perform normalization in its Lie algebra. In this subsection, we compare this Lie-algebra-based approach with our LieBN on rotation matrices. This idea corresponds to constructing a distribution on the Lie group from a Euclidean Gaussian in its Lie algebra \citep{wang2006error,de2025wrapped}. In this formulation, the associated mean and variance reduce to their Euclidean counterparts in the Lie algebra. We therefore construct \emph{Lie Algebra Batch Normalization (LieAlgebraBN)} as follows:
\begin{itemize}
    \item
    Mapping data into the Lie algebra by the Riemannian logarithm at the neutral element $\rielog_{E}$;
    \item
    Applying Euclidean BN over the Lie algebra;
    \item
    Mapping back to the Lie group by the Riemannian exponentiation at the neutral element $\rieexp_{E}$.
\end{itemize}

\mypara{Disadvantages of LieAlgebraBN.}
However, we argue that LieAlgebraBN may distort the geometry because it relies solely on a single tangent space.

\mypara{Experiments on the rotation Lie groups.}
We compare LieAlgebraBN and LieBN with the LieNet backbone on the relatively large NTU60 data set. Following the NTU60 settings used for the rotation LieBN experiments, we adopt two-block and three-block architectures. We observe that LieAlgebraBN benefits from a decreasing learning rate upon convergence. Therefore, we apply the same learning rate schedule as in LieBN while keeping all other settings identical. \cref{fig:acc_liebn_vs_liealgebrabn_so3} presents the test accuracy curves, demonstrating that LieBN outperforms LieAlgebraBN under both architectures. This may be attributed to the fact that the Lie algebra, as a local tangent space approximation, distorts the intrinsic geometry of the Lie group, leading to suboptimal normalization.

\section{Conclusions}
\label{sec:conclusions}

This paper presents a novel LieBN framework for batch normalization over Lie groups, leveraging natural Lie-group-invariant metrics. Compared to prior approaches, LieBN provides a principled method to normalize both sample and population statistics. Then, we generalize three existing Lie group structures on the SPD manifold and introduce the first non-trivial right-invariant SPD metric. By employing these parameterized invariant metrics, we instantiate our framework on the SPD manifold. Furthermore, we implement LieBN on rotation matrices using a bi-invariant metric and on the correlation manifold using four bi-invariant metrics. Extensive experiments across different manifolds validate the effectiveness of our LieBN.

\acks{This work was supported by the FIS project GUIDANCE (No. FIS2023-03251), the EU Horizon project ELLIOT (No. 101214398), a DAAD Research Grant in Germany (57811724), and an ELIZA PhD Mobility Scholarship. We acknowledge CINECA and EuroHPC for awarding high-performance computing resources. The authors declare that they have no competing interests.}

\clearpage
\appendix
\startcontents[appendices]
\printcontents[appendices]{l}{1}{\section*{Appendix Contents}}
\newpage

\section{Notations} \label{app:notations}
For clarity, we summarize the notation used throughout this paper in \cref{app:tab:sum_notaitons}.

\begin{table}[t]
    \centering
    \resizebox{0.95\linewidth}{!}{
    \begin{tabular}{cc}
        \toprule
        Notation & Explanation  \\
        \midrule
        $\{\calM, \odot, g \}$ or abbreviated as $\calM$ & Lie group with a group operation $\odot$ and an invariant metric $g$ \\
        $\gleft$ and $\gright$ & Left-invariant and right-invariant metrics\\
        $P_\odot ^{-1}$ & Group inverse of $P \in \calM$\\
        $T_P\calM$ & Tangent space at $P \in \calM$\\
        $g_p(\cdot ,\cdot)$ or $\langle \cdot, \cdot \rangle_P$ & Riemannian metric at $P \in \calM$ \\
        $\| \cdot \|_P$ & Norm induced by $\langle \cdot, \cdot \rangle_P$ on $T_P\calM$ \\
        $\dist(\cdot,\cdot)$ & Geodesic distance\\
        $\fm$ and $\wfm$ & Fréchet mean and weighted Fréchet mean\\
        $\rieexp_P$ and $\rielog_P$ & Riemannian exponentiation and logarithm at $P$\\
        $\gamma _{(P,Q)}(t)$ & Geodesic connecting $P$ and $Q$\\
        $\pt{P}{Q}$ & Riemannian parallel transportation along the geodesic connecting $P$ and $Q$\\
        $\ltrans_P$ and $\rtrans_P$ & Lie group left and right translation by $P \in \calM$ \\
        $f_{*,P}$ & Differential map of the smooth map $f$ at $P \in \calM$\\
        $f^*g$ & Pullback metric by $f$ from $g$\\
        $\bbR{n \times n}$ & Euclidean space of $n \times n$ real matrices\\
        $\spd{n}$ & SPD manifold of $n \times n$ SPD matrices \\
        $\sym{n}$ & Euclidean space of $n \times n$ symmetric matrices \\
        $\trilspace{n}$ & Euclidean space of $n \times n$ lower triangular matrices\\
        $\so{n}$ & Lie group of $n \times n$ rotation matrices\\
        $\soLieAlgebra{n}$ & Euclidean space of $n \times n$ skew-symmetric matrices\\
        $\langle \cdot, \cdot \rangle$ and $\| \cdot \|_\rmF$ & Standard Frobenius inner product and the induced norm \\
        $\langle \cdot, \cdot \rangle^{\alphabeta}$ and $\| \cdot \|^{\alphabeta}$ & $\orth{n}$-invariant Euclidean inner product and  the induced norm \\
        $\bfst$ & $\bfst = \{(\alpha,\beta) \in \mathbb{R}^2 \mid \min (\alpha, \alpha+n \beta)>0\}$ \\
        $\gbiparamai$, $\gtriparamAI$, $\gcri$ and $\gdefcri$ & Riemannian metric tensors of $\biparamAIM$, $\triparamAIM$, CRIM, and $\theta$-CRIM\\
        $\odotai$, $\odotle$, and $\odotlc$ & Group operations with respect to AIM, LEM, and LCM \\
        $\symmetrize{\cdot}$ & $\symmetrize{X}=X + X^\top, \forall X \in \bbR{n \times n}$ \\
        $(\cdot)_{\frac{1}{2}}$ & $(X)_{\frac{1}{2}}=\lfloor X \rfloor + \frac{1}{2}\bbD(X), \forall X \in \bbR{n \times n}$\\
        $\calN(M,\sigma^2)$ & Riemannian Gaussian distribution\\
        $\mlog$ and $\mexp$ & Matrix logarithm and exponentiation\\
        $\chol$ & Cholesky decomposition\\
        $\dlog$ & Diagonal element-wise logarithm \\
        $\clog$ & $\dlog \circ \chol$ \\
        $\lfloor \cdot \rfloor$ & Strictly lower triangular part of a square matrix \\
        $\pow_\theta(\cdot)$ or $(\cdot)^{\theta}$ & Matrix power function \\
        $\bbD(\cdot)$  & Returns a diagonal matrix with diagonal elements from a square matrix\\
        $\diag(\cdot)$ & Returns a diagonal matrix from an input vector \\
        $\hol{n}$ and $\rzero{n}$ & Subspaces of $\sym{n}$ with null diagonals and null row sums \\
        $\rone{n}$ & Manifold of $n \times n$ SPD matrices with unit row sum. \\
        $\cor{n}$ & Manifold of $n \times n$ full-rank correlation matrices \\
        $\LTone{n}$ and $\LTzero{n}$ & Euclidean subspaces of $\trilspace{n}$ with unit diagonals and null diagonals \\
        $\chospace{n}$ & Cholesky manifold of $n \times n$ lower triangular matrices with positive diagonals\\
        $\circledast$ & Hadamard product \\
        $\coropt$ & $\coropt: \Sigma \in \spd{n} \longmapsto \bbD(\Sigma)^{-\nicefrac{1}{2}} \Sigma \bbD(\Sigma)^{-\nicefrac{1}{2}} \in \cor{n}$ \\
        $\Theta$ & $ \Theta : C \in \cor{n} \longmapsto \bbD(\chol(C))^{-1} \chol(C) \in \LTone{n}$ \\
        $\off$ & Returns a matrix in $\hol{n}$ consisting of off-diagonal elements \\
        $\offlog \& \offexp$ & Off-log and its inverse \\
        $\logscaled \& \expscaled$ & Log-scaled and its inverse \\
        $I$ or $I_n$ \& $\bfzero$ & Identity matrix \& zero matrix\\
        $\holinner{\cdot}{\cdot}$ and $\rzeroinner{\cdot}{\cdot}$ & Permutation-invariant inner products over $\hol{n}$ and $\rzero{n}$ \\
        \bottomrule
    \end{tabular}
    }
    \caption{Summary of notations.}
    \label{app:tab:sum_notaitons}
\end{table}

\section{Riemannian Structures on the Involved Matrix Manifolds}

\subsection{Existence and Uniqueness of the Weighted Fréchet Mean}
\label{app:subsec:exist_unique_wfm}

Let $(\calM, g)$ be an orientable complete Riemannian manifold equipped with a Riemannian metric $g$. The induced distance is denoted by $\dist(\cdot,\cdot)$. We denote the supremum of the sectional curvatures of $\calM$ by $\Delta$. We recover the theorem on the existence and uniqueness of the weighted Fréchet mean (WFM) \citep{afsari2011riemannian}. We acknowledge that \citet[Appendix~A]{chakraborty2022manifoldnet} has also provided a summary of the following discussions.

\begin{parisdefinition}[Geodesic Ball]
Let $P \in \calM$ and $r > 0$. Then $B_r(P) = \{Q \in \calM \mid d(P, Q) < r\}$ is the open geodesic ball at $P$ of radius $r$.
\end{parisdefinition}

\begin{parisdefinition}[Injectivity Radius \citep{manton2004globally}]
The local injectivity radius at $P \in \calM$, $r_{\text{inj}}(P)$, is the largest radius $r$ for which $\rieexp_P : T_P \calM \supset B_r(0) \to \calM$ is a diffeomorphism onto its image. The injectivity radius of $\calM$ is defined as $r_{\text{inj}}(\calM) = \inf_{P \in \calM} \{r_{\text{inj}}(P)\}$.
\end{parisdefinition}

Within the local injectivity radius, the exponential map is invertible and we call the inverse map the Riemannian logarithmic map, $\rielog _{P} : B_{r_{\text{inj}}(P)} (P) \rightarrow B _{r_{\text{inj}}(P)}(0) \subset T _P \calM$.

\begin{parisdefinition}[Regular Geodesic Ball \citep{kendall1990probability}]
An open geodesic ball $B_r(P)$ is a regular geodesic ball if $r < r_{\text{inj}}(P)$ and $r < \pi/(2 \sqrt{\Delta})$, where $1 / \sqrt{\Delta}$ is interpreted as $\infty$ for $\Delta \leq 0$.
\end{parisdefinition}

If $P, Q$ are in a regular geodesic ball $B_r (P)$, there exists a unique geodesic $\gamma _{(P,Q)} :[0,1] \rightarrow B_r(P)$ with $\gamma(0)=P$ and $\gamma(1)=Q$ \citep{kendall1990probability}.

\begin{parisdefinition}[Strong Convexity \citep{chavel1995riemannian}]
A subset $U \subset \calM$ is strongly convex if for all $P, Q \in U$, there exists a unique length-minimizing geodesic segment between $P$ and $Q$, and the geodesic segment lies entirely in $U$.
\end{parisdefinition}

\begin{parisdefinition}[Convexity Radius \citep{groisser2004newton}]
The local convexity radius at $P \in \calM$, $r_{\text{cvx}}(P)$, is defined as
\begin{equation}
    r_{\text{cvx}}(P) = \sup \{r \le r_{\text{inj}}(P) \mid B_r(P) \text{ is strongly convex}\}.
\end{equation}
The convexity radius of $\calM$ is defined as $r_{\text{cvx}}(\calM) = \inf_{P \in \calM} \{r_{\text{cvx}}(P)\}$.
\end{parisdefinition}

\begin{paristheorem}[Existence and Uniqueness of WFM \citep{afsari2011riemannian}]
The WFM exists and is unique inside a geodesic ball of radius $r_{\text{cvx}}(\calM)$.
\end{paristheorem}

In the main paper, we always assume the involved $\rieexp$, $\rielog$, and WFM are well-defined.

\subsection{Symmetric Matrix Functions}
This subsection reviews the eigenvalue function over symmetric matrices. For more in-depth discussions, please refer to \citet[Chapter~2.7.13]{bhatia2009positive} or \citet[Chapter~V.3]{bhatia2013matrix}.

We denote $\sym{n}$ as the Euclidean space of $n \times n$ real symmetric matrices, and $\spd{n}$ as the SPD manifold of $n \times n$ SPD matrices. Let $\mathring{I}$ be an open interval of $\bbRscalar$ and $f: \mathring{I} \rightarrow \bbRscalar$ be a smooth function. The smooth map induced by $f$ for any symmetric matrix $S$ with all eigenvalues in $\mathring{I}$ is defined as
\begin{equation}
    f: S \longmapsto U f(\Sigma) U^\top \in \sym{n}, \quad \text{where } S=U \Sigma U^\top \text{ is the eigendecomposition.}
\end{equation}
Its differential is known as the Dalecki\u{\i}--Kre\u{\i}n formula:
\begin{align}
    \label{app:eq:diff_matrix_function}
    f_{*,S} (V) &=U\left(L \circledast \left(U^{\top} V U\right)\right) U^{\top}, \quad \forall V \in \sym{n}, \\
    \label{app:eq:l_matrix}
    L_{i,j}&=
    \begin{cases}
    \frac{f(\sigma_i)-f(\sigma_j)}{\sigma_i-\sigma_j}, & \text { if } \sigma_i \neq \sigma_j \\
    f^{\prime}(\sigma_i), & \text { otherwise }
    \end{cases},
\end{align}
where $L$ is called the Loewner matrix with the $(i,j)$-th element defined as \cref{app:eq:l_matrix}, and $\circledast$ denotes the Hadamard product. Two special cases are the matrix logarithm: $\log: \spd{n} \rightarrow \sym{n}$ and its inverse, the matrix exponentiation $\exp: \sym{n} \rightarrow \spd{n}$.

\subsection{SPD Geometries}
\label{app:subsec:spd_geometries}

\cref{app:tab:riem_operators_props} summarizes the Lie groups and invariant metrics on the SPD manifold with the following notation. Let $P, Q \in \spd{n}$ be SPD matrices and $V, W \in T_P\spd{n}$ be tangent vectors. We denote the matrix logarithm, exponentiation, and Cholesky decomposition by $\mlog(\cdot)$, $\mexp(\cdot)$, and $\chol(\cdot)$, respectively. The differentials at $P$ are $\mlog_{*,P}$ and $\chol_{*,P}$. The Cholesky factors of $P$ and $Q$ are denoted as $L=\chol(P)$ and $K=\chol(Q)$. The corresponding tangent vectors are $X = \chol_{*,P} (V)$ and $Y = \chol_{*,P} (W)$ for LCM. $\bbK$, $\bbL$, $\bbX$, and $\bbY$ are diagonal matrices with diagonal elements from $K$, $L$, $X$, and $Y$, respectively. $\lfloor \cdot \rfloor$ is the strictly lower part of a square matrix. The norms induced by $\langle \cdot, \cdot \rangle^{\alphabeta}$ and the standard inner product $\inner{\cdot}{\cdot}$ are denoted by $\| \cdot \|^{\alphabeta}$ and $\| \cdot \|_\rmF$, respectively.

% \cref{app:tab:riem_operators_props} summarizes all the necessary Riemannian operators under the following notations. We denote the matrix logarithm, exponentiation, and Cholesky decomposition by $\mlog(\cdot)$, $\mexp(\cdot)$, and $\chol(\cdot)$, respectively. The differentials at $P$ are $\mlog_{*,P}$ and $\chol_{*,P}$. The Cholesky factors of $P$ and $Q$ are denoted as $L=\chol(P)$ and $K=\chol(Q)$, respectively. The differentials of $\mlog$ and $\chol^{-1}$ at $P$ and $L$ are represented by $\mlog_{,P}$ and $(\chol)^{-1}_{,L}$, respectively. We denote $X = \chol_{*,P} (V)$ as the resulting tangent vector of $V$ under the differential of the Cholesky decomposition. $\bbK$, $\bbL$, and $\bbX$ are diagonal matrices with diagonal elements from $K$, $L$, and $X$, respectively. Finally, we denote the norms induced by $\langle \cdot, \cdot \rangle^{\alphabeta}$ and the standard Frobenius norm by $\| \cdot \|^{\alphabeta}$ and $\| \cdot \|_\rmF$, respectively.

\begin{table}[tbp]
    \centering
    \resizebox{0.99\linewidth}{!}{
    \begin{tabular}{c|ccc}
        \toprule
        Metric & $\biparamLEM$ & $\biparamAIM$ & LCM \\
        \midrule
        $ Q \odot P$ & $\mexp(\mlog(P)+\mlog(Q))$ & $KPK^\top$ & $\chol^{-1}(\lfloor L + K \rfloor + \bbK \bbL)$ \\
        \midrule
        $g_P(V,W)$ & $\langle \mlog_{*,P} (V), \mlog_{*,P} (W) \rangle^{\alphabeta}$ & $\langle P^{-1}V, W P^{-1} \rangle^{\alphabeta}$ & $\langle \lfloor X \rfloor, \lfloor Y \rfloor \rangle + \langle \bbX \bbL^{-1}, \bbY \bbL^{-1} \rangle$ \\
        \midrule
        $\dist(P,Q)$ & $\left \|\mlog(P)-\mlog(Q) \right\|^{\alphabeta}$ & $\left \|\mlog \left(Q^{-\frac{1}{2}}PQ^{-\frac{1}{2}} \right) \right\|^{\alphabeta}$ & $\left \| \clog (P) - \clog (Q) \right \|_\rmF$\\
        \midrule
        $\fm (\{P_i\})$ & $\mexp \left( \frac{1}{n} \sum_i \mlog{P_i} \right)$ & Karcher Flow & $\clog^{-1} \left( \frac{1}{n} \sum_i \clog(P_i) \right)$ \\
        \midrule
        $\rielog_P Q$ & $(\mlog_{*,P})^{-1} \left[ \mlog(Q) - \mlog(P) \right]$ & $P^{\frac{1}{2}} \mlog \left(P^{ -\frac{1}{2}} Q P^{-\frac{1}{2}}\right) P^{\frac{1}{2}}$ & $ (\chol^{-1})_{*, L} \left[ \lfloor K\rfloor-\lfloor L\rfloor+\bbL \dlog (\bbL^{-1} \bbK) \right]$ \\
        \midrule
        $\rieexp _P V$ & $\mexp \left( \mlog(P) + \mlog _{*,P} (V) \right)$ & $P^{\frac{1}{2}} \mexp \left(P^{ -\frac{1}{2}} V P^{-\frac{1}{2}}\right) P^{\frac{1}{2}}$ & $\chol^{-1} \left( \lfloor L \rfloor + \lfloor X \rfloor + \bbL \exp \left( \bbX \bbL^{-1} \right)  \right)$\\
        \midrule
        $\gamma _{(P,Q)} (t)$ & $\mexp \left[ \mlog(P) + t\left(\mlog(Q) - \mlog(P)\right) \right]$ & $P^{\frac{1}{2}} \left(P^{ -\frac{1}{2}} Q P^{-\frac{1}{2}}\right)^{t} P^{\frac{1}{2}}$ & $\chol^{-1} \left\{ \lfloor L \rfloor + t(\lfloor K \rfloor- \lfloor L \rfloor) + \frac{\bbK^t}{\bbL^{t-1}} \right\}$ \\
        \midrule
        Invariance & Bi-invariance & Left-invariance & Bi-invariance \\
        \midrule
        References & \makecell{\citep{arsigny2005fast} \\ \citep{thanwerdas2023n} \\ \citep{chen2024adaptive}} & \makecell{\citep{pennec2006riemannian} \\ \citep{thanwerdas2023n} \\ \citep{thanwerdas2022theoretically}} & \makecell{\citep{lin2019riemannian} \\ \citep{chen2024adaptive} \\ \citep{chen2026product}} \\
        \bottomrule
    \end{tabular}
    }
    % \vspace{-3mm}
    \caption{Lie group structures and the associated Riemannian operators on SPD manifolds.}
    \label{app:tab:riem_operators_props}
\end{table}

\subsection{Correlation Geometries}
\label{app:subsec:cor_geometries}

\begin{table}[t!]
    \centering
    \resizebox{\linewidth}{!}{
    \begin{tabular}{cccc}
    \toprule
    Metric & Prototype space & Diffeomorphisms & Properties \\
    \midrule
    \makecell{ECM \\ \citep{thanwerdas2022theoretically}} & $\LTone{n} = \LTzero{n} + I_n$ &
    \makecell{
    $ \Theta : C \in \cor{n} \longmapsto \bbD(\chol(C))^{-1} \chol(C) \in \LTone{n}$ \\
    $\Theta^{-1} = \coropt \circ \chol^{-1} : \LTone{n} \longrightarrow \cor{n}$ } & Bi-invariance \\
    \midrule
    \makecell{LECM \\ \citep{thanwerdas2022theoretically}} & $\LTzero{n}$ &
    \makecell{
    $\log \circ \Theta: \cor{n} \longrightarrow \LTzero{n}$ \\
    $(\log \circ \Theta)^{-1} = \coropt \circ \chol^{-1} \circ \exp: \LTzero{n} \longrightarrow \cor{n}$ } & Bi-invariance \\
    \midrule
    \makecell{OLM \\ \citep{thanwerdas2024permutation}} & $\hol{n}$ &
    \makecell{
    $\offlog: C \in \cor{n} \longmapsto \off \circ \log(C) \in \hol{n}$ \\
    $(\offlog) ^{-1} = \offexp: H \in \hol{n} \longmapsto \exp ( \dplus(H) + H ) \in \cor{n}$} & \makecell{Bi-invariance \\ Permutation-invariance} \\
    \midrule
    \makecell{LSM \\ \citep{thanwerdas2024permutation}} & $\rzero{n}$ & \makecell{
    $\logscaled: C \in \cor{n} \longmapsto \log ( \dstar (C) C \dstar (C)) \in \rzero{n}$ \\
    $(\logscaled)^{-1} = \expscaled: R \in \rzero{n} \longmapsto \coropt(\exp(R)) \in \cor{n}$
    } & \makecell{Bi-invariance \\ Permutation-invariance} \\
    \bottomrule
    \end{tabular}
    }
    % \vspace{-3mm}
    \caption{Riemannian metrics on the correlation manifold with the associated isometric prototype spaces and diffeomorphisms.}
    \label{tb:cor_iso_spaces_maps}
\end{table}

The four geometries on correlation matrices discussed in \cref{subsec:cor_lie_groups} can be classified into two classes: (1) non-permutation-invariant metrics, including ECM and LECM; and (2) permutation-invariant metrics, including OLM and LSM. \cref{tb:cor_iso_spaces_maps} summarizes the diffeomorphisms and prototype spaces discussed in \cref{subsec:cor_lie_groups}.

\subsubsection{Non-Permutation-Invariant Metrics}
\label{app:subsubsec:non_perm_metrics}

The non-permutation-invariant metrics \citep{thanwerdas2022theoretically}, namely ECM and LECM, are defined by isometries:
\begin{align}
    \text{ECM: }& \cor{n} \xrightleftharpoons[\Theta^{-1} = \covtocor \circ \chol^{-1}]{\Theta} \LTone{n} = I_n + \LTzero{n}, \\
    \text{LECM: }&  \cor{n} \xrightleftharpoons[(\log \circ \Theta)^{-1} = \covtocor \circ \chol^{-1} \circ \exp]{\log \circ \Theta} \LTzero{n},
\end{align}

\mypara{ECM and LECM.}
For any $C \in \cor{n}$, $V \in T_C \cor{n} \cong \hol{n}$, $K \in \LTone{n}$ and $X, \xi \in \LTzero{n}$, the involved maps and their differentials in ECM and LECM are
\begin{align}
    \Theta(C) &= \bbD(L)^{-1} L, \\
    \Theta^{-1} (K) &= \bbD\left(K K^{\top}\right)^{-\frac{1}{2}} K K^{\top} \bbD\left(K K^{\top}\right)^{-\frac{1}{2}}, \\
    \label{app:eq:log_ltone}
    \log (K) &=\sum_{k=1}^{n-1} \frac{(-1)^{k-1}}{k}\left(K-I_n\right)^k, \\
    \label{app:eq:exp_ltzero}
    \exp (\xi) &=\sum_{k=0}^{n-1} \frac{1}{k!} \xi^k, \\
    \label{app:eq:diff_Theta}
    \Theta_{*,C}(V) &= \Theta(C) \left(L^{-1} V L^{-\top}\right)_{\frac{1}{2}} -\frac{1}{2} \bbD \left(L^{-1} V L^{-\top}\right) \Theta(C),\\
    \left(\Theta_{*,C} \right)^{-1}(\xi) &=  \left( L \xi^{\top}-C \bbD\left(L \xi^{\top}\right)\right) \bbD(L) +\bbD(L)\left(\xi L^{\top}-\bbD\left(L \xi^{\top}\right) C \right),
\end{align}
\begin{equation}
    \label{app:eq:diff_log_LTone}
    \begin{aligned}
        \log_{*,K} (\xi) &= \sum_{k=1}^{n-1} \frac{(-1)^{k-1}}{k}\left[\left(K-I_n\right)^{k-1} \xi +\left(K-I_n\right)^{k-2} \xi\left(K-I_n\right)\right. \\
        &\qquad\left.{}+\cdots+\xi\left(K-I_n\right)^{k-1}\right],
    \end{aligned}
\end{equation}
\begin{align}
    \exp_{*, X} (\xi) &= \sum_{k=1}^{n-1} \frac{1}{k!}\left(X^{k-1} \xi+X^{k-2} \xi X+\cdots+\xi X^{k-1}\right), \\
    \label{app:eq:diff_logTheta}
    (\log \circ \Theta)_{*,C} (V) &=  \log_{*, \Theta(C)} \left( \Theta_{*,C}(V) \right),\\
    \label{app:eq:diff_chol}
    \chol_{*,C} (V) &= L \left(L^{-1} V L^{-\top}\right)_{\frac{1}{2}}, \\
    (\chol_{*,C})^{-1} (Z) &= L Z^\top + Z L^\top, \quad \forall Z \in T_L \chospace{n} \cong \trilspace{n},
\end{align}
where $L$ is the Cholesky factor of $C$, $I_n$ is the $n \times n$ identity matrix and $\chospace{n}$ is the Cholesky manifold of $n \times n$ Cholesky matrices. Due to the nilpotency of $\LTzero{n}$, the matrix logarithm over $\LTone{n}$ and exponentiation over $\LTzero{n}$ are free from eigendecomposition. With the above equations, \cref{app:tab:riem_operators_ecm_lecm} summarizes the Riemannian operators under ECM and LECM.

\begin{table}[t]
    \centering
    \resizebox{\linewidth}{!}{
    \begin{tabular}{ccc}
        \toprule
        Operation & ECM & LECM \\
        \midrule
        $g_{C}(V,W)$ & $\inner{\Theta_{*, C}(V)}{\Theta_{*, C}(W)}$ & $\inner{(\log \circ \Theta)_{*, C}(V)}{(\log \circ \Theta)_{*, C}(W)}$ \\
        $\rieexp_C(V)$ & $ \Theta^{-1}\left(\Theta\left(C\right) + \Theta_{*, C} \left( V\right)\right)$ & $ (\log \circ \Theta)^{-1}\left(\log \circ \Theta\left(C\right) + (\log \circ \Theta)_{*, C} \left( V \right) \right)$ \\
        $\rielog_C(C')$ & $\Theta^{-1}_{{*, \Theta(C)}}\left(\Theta\left(C'\right) - \Theta\left(C\right)\right)$ & $(\log \circ \Theta)^{-1}_{{*, \log \circ \Theta(C)}}\left(\log \circ \Theta\left(C'\right) - \log \circ \Theta\left(C\right)\right)$ \\
        $\gamma(t;C,C')$ &  $\Theta^{-1}\left(\left(1 - t\right)\Theta\left(C\right) + t\Theta\left(C'\right)\right)$ &  $(\log \circ \Theta)^{-1}\left(\left(1 - t\right)\log \circ \Theta\left(C\right) + t\log \circ \Theta\left(C'\right)\right)$ \\
        $\dist(C, C')$ & $ \fnorm{ \Theta\left(C\right) - \Theta\left(C'\right) }$ & $ \fnorm{ \log \circ \Theta\left(C\right) - \log \circ \Theta\left(C'\right) }$ \\
        Fréchet mean & $\Theta^{-1}\left(\frac{1}{k} \sum_{i=1}^{k} \Theta\left(C_i\right)\right)$ & $(\log \circ \Theta)^{-1}\left(\frac{1}{k} \sum_{i=1}^{k} \log \circ \Theta\left(C_i\right)\right)$ \\
        Curvature & $0$ & $0$ \\
        $\pt{C}{C'} (V)$ & $\left(\Theta_{*, C'} \right)^{-1}\left(\Theta_{*, C} \left(V\right)\right)$ & $\left((\log \circ \Theta)_{*, C'}\right)^{-1}\left((\log \circ \Theta)_{*, C} \left(V\right)\right)$ \\
        \bottomrule
    \end{tabular}
    }
    \caption{Riemannian operators under the non-permutation-invariant log-Euclidean metrics. Here, $C, C' \in \cor{n}$ are correlation matrices and $V,W \in T_C \cor{n} \cong \hol{n}$ are tangent vectors.}
    \label{app:tab:riem_operators_ecm_lecm}
\end{table}

\subsubsection{Permutation-Invariant Metrics}
\label{app:subsubsec:perm_metrics}

\begingroup\tolerance=1000\emergencystretch=10pt
Let $\perm{n}$ be the group of permutation matrices $P_\sigma=\left[\delta_{i, \sigma(j)}\right]_{1 \leqslant i, j \leqslant n}$ associated with the permutation $\sigma$, and $\mathcal{D}^{\pm}(n)=\left\{\diag\left( \left(\varepsilon_1, \ldots, \varepsilon_n \right) \right), \varepsilon \in\{-1,1\}^n\right\}$ be the group of diagonal matrices with coefficients in $\{-1,1\}$. \citet[Theorem~1.1]{thanwerdas2024permutation} showed that the largest congruence action on full-rank correlation matrices is the action of signed permutation matrices:\par
\endgroup
\begin{equation}
    \star:(A, C) \in \singperm{n} \times \cor{n} \longmapsto A C A^{\top} \in \cor{n},
\end{equation}
\begingroup\tolerance=1000\emergencystretch=10pt
with $\singperm{n}=\mathcal{D}^{\pm}(n) \perm{n}$.\linebreak[2] Based on this finding, \citet{thanwerdas2024permutation} proposed two permutation-invariant metrics, namely OLM and LSM, by pulling back permutation-invariant inner products via the following permutation-equivariant diffeomorphisms:\par
\endgroup
\begin{align}
    \cor{n} &\xrightleftharpoons[\offexp]{\offlog=\off \circ \log} \hol{n},\\
    \cor{n} &\xrightleftharpoons[\expscaled=\coropt \circ \exp]{\logscaled} \rzero{n}, \\
    \offexp: \hol{n} \ni H &\longmapsto \exp(\dplus(H) + H), \\
    \logscaled: \cor{n} \ni C &\longmapsto \log ( \dstar (C) C \dstar (C)) \in \rzero{n},
\end{align}
where $\log(\cdot)$ and $\exp(\cdot)$ are the symmetric matrix logarithm and exponentiation. The involved $\dplus$ and $\dstar$ can be formally expressed as $\dplus: \hol{n} \rightarrow \bbDspace{n}$ and $\dstar: \cor{n} \rightarrow \bbDplus{n}$, where $\bbDspace{n}$ denotes the Euclidean space of $n \times n$ diagonal matrices, and $\bbDplus{n}$ is the submanifold of $\bbDspace{n}$, consisting of positive diagonal matrices.

The differentials of $\offlog$ and $\logscaled$ and their inverses can be calculated using the differentials of the symmetric matrix logarithm and exponentiation \citep[Theorems~2.4 and~4.1]{thanwerdas2024permutation}. Given $C \in \cor{n}$, a tangent vector $V \in T_C\cor{n} \cong \hol{n}$, $H, W \in \hol{n}$, and $S = H+\dplus(H) = U \Delta U^{\top}$, the differentials of $\offlog$ and its inverse $\offexp$ are
\begin{align}
    \label{app:eq:diff_offlog}
    \offlog_{*,C} (V) &=\off \left(\log_{*,C} (V)\right),\\
    \offexp_{*,H}(W) & = \exp_{*, S} \left(W+\dplus_{*,H}(W)\right), \\
    \label{app:eq:diff_dplus}
    \dplus_{*,H} (W) & =-\diag \left(\left(H^0\right)^{-1} \bbD \left( \exp_{*, S} (W)\right) \vecone \right), \\
    \spd{n} \ni H_{i l}^0 &= \sum_{j, k} P_{i j} P_{i k} P_{l j} P_{l k} L_{j,k},
\end{align}
where $L$ is the Loewner matrix of $\exp_{*, S}$, and  $\vecone$ is the vector of all ones. Here, $\log_{*}$ and $\exp_{*}$ can be calculated using the Dalecki\u{\i}--Kre\u{\i}n formula of the symmetric matrix, while $\diag(\cdot): \bbR{n} \rightarrow \bbDspace{n}$ returns a diagonal matrix from an input vector. Further denoting $X, Y \in \operatorname{Row}_0(n)$ and $\Sigma=\dstar(C) C \dstar(C)$, the differentials of $\logscaled$ and its inverse $\expscaled$ are
\begin{align}
    \label{app:eq:diff_logscaled}
    \logscaled_{*,C}(V) &=\log_{*,\Sigma}  \left(\Delta V \Delta+\frac{1}{2}\left(V^0 \Sigma+\Sigma V^0\right)\right),\\
    \expscaled_{*,X}(Y) &=\Delta^{-1}\left[\exp_{*,X} (Y)-\frac{1}{2}\left(\Delta^{-2} \bbD\left(\exp_{*,X} (Y)\right) \Sigma+\Sigma \bbD\left(\exp_{*,X} (Y)\right) \Delta^{-2}\right)\right] \Delta^{-1},
\end{align}
with $\Delta=\bbD(\Sigma)^{1 / 2}$ and $V^0=-2 \diag \left(\left(I_n+\Sigma\right)^{-1} \Delta V \Delta \vecone \right)$.

As both $\logscaled_{*}$ and $\offlog_{*}$ are permutation-equivariant \citep{thanwerdas2024permutation}, permutation-invariant metrics over the correlation manifold can be induced by permutation-invariant inner products over $\hol{n}$ and $\rzero{n}$, respectively. The following two theorems review such inner products.

\begin{table}[t]
    \centering
    \resizebox{\linewidth}{!}{
    \begin{tabular}{ccc}
        \toprule
        Operation & OLM & LSM \\
        \midrule
        $g_{C}(V,W)$ & $\holinner{\offlog_{*, C}(V)}{\offlog_{*, C}(W)}$ & $\rzeroinner{\logscaled_{*, C}(V)}{\logscaled_{*, C}(W)}$ \\
        $\rieexp_C(V)$ & $ \offexp\left(\offlog\left(C\right) + \offlog_{*, C} \left( V\right)\right)$ & $ \expscaled\left(\logscaled\left(C\right) + \logscaled_{*, C} \left( V \right) \right)$ \\
        $\rielog_C(C')$ & $\offexp_{{*, \offlog(C)}}\left(\offlog\left(C'\right) - \offlog\left(C\right)\right)$ & $\expscaled_{{*, \logscaled(C)}}\left(\logscaled\left(C'\right) - \logscaled\left(C\right)\right)$ \\
        $\gamma(t;C,C')$ &  $\offexp\left(\left(1 - t\right)\offlog\left(C\right) + t\offlog\left(C'\right)\right)$ &  $\expscaled\left(\left(1 - t\right)\logscaled\left(C\right) + t\logscaled\left(C'\right)\right)$ \\
        $\dist(C, C')$ & $ \holnorm{ \offlog\left(C\right) - \offlog\left(C'\right) }$ & $ \rzeronorm{ \logscaled\left(C\right) - \logscaled\left(C'\right) }$ \\
        Fréchet mean & $\offexp\left(\frac{1}{k} \sum_{i=1}^{k} \offlog\left(C_i\right)\right)$ & $\expscaled\left(\frac{1}{k} \sum_{i=1}^{k} \logscaled\left(C_i\right)\right)$ \\
        Curvature & $0$ & $0$ \\
        $\pt{C}{C'} (V)$ & $\left(\offlog_{*, C'} \right)^{-1}\left(\offlog_{*, C} \left(V\right)\right)$ & $\left(\logscaled_{*, C'}\right)^{-1}\left(\logscaled_{*, C} \left(V\right)\right)$ \\
        Invariance & \makecell{Bi-invariance \\ Permutation-invariance \\
        Signed-permutation-invariance $(\beta=\gamma=0)$} & \makecell{Bi-invariance \\ Permutation-invariance } \\
        \bottomrule
    \end{tabular}
    }
    \caption{Riemannian geometries under the permutation-invariant log-Euclidean metrics.}
    \label{app:tab:olm_lsm}
\end{table}

\begin{paristheorem}[Permutation-Invariant Inner Products on $\hol{n}$ \citep{thanwerdas2022riemannian}]
\label{app:thm:invariant_hol_inner}
Supposing $n \geq 4$, permutation-invariant inner products on $\hol{n}$ are:
\begin{equation}
    \begin{aligned}
        \holinner{X_1}{X_2}
        &= \alpha \tr( X_1 X_2) +\beta \Sum\left(X_1 X_2\right) \\
        &\quad +\gamma \Sum(X_1) \Sum(X_2), \quad \forall X_1,X_2 \in \hol{n},
    \end{aligned}
\end{equation}
with $\alpha > 0$, $2\alpha + (n - 2)\beta > 0$, and $\alpha + (n - 1)(\beta + n\gamma) > 0$. For $n = 3$, permutation-invariant inner products have the same form with $\alpha = 0$:
\begin{equation}
\holinner{X_1}{X_2} = \beta \Sum(X_1 X_2) + \gamma \Sum(X_1)\Sum(X_2), \quad \text{with } \beta > 0  \text{ and } \beta + 3\gamma > 0.
\end{equation}
For $n = 2$, they have the same form with $\alpha = \beta = 0$:
\begin{equation}
\holinner{X_1}{X_2} = \gamma \Sum(X_1)\Sum(X_2), \quad \text{with } \gamma > 0.
\end{equation}
\end{paristheorem}

\begin{paristheorem}[Permutation-Invariant Inner Products on $\rzero{n}$ \citep{thanwerdas2024permutation}]
\label{app:thm:invariant_rzero_inner}
For $n \geq 4$, permutation-invariant inner products on $\rzero{n}$ are
\begin{equation}
\rzeroinner{Y_1}{Y_2} = \alpha \tr(Y_1 Y_2) + \delta \tr(\bbD(Y_1)\bbD(Y_2)) + \zeta  \tr(Y_1)\tr(Y_2), \quad \forall Y_1,Y_2 \in \rzero{n},
\end{equation}
with $\alpha > 0$, $n\alpha + (n-2)\delta > 0$, and $n\alpha + (n-1)(\delta + n\zeta) > 0$.
For $n = 3$, the permutation-invariant inner products have the same form with $\alpha = 0$.
For $n = 2$, they have the same form with $\alpha = \delta = 0$.
\end{paristheorem}

As shown by \citet{thanwerdas2022riemannian}, when $\beta=\gamma=0$, OLM is also signed-permutation-invariant, and its inner product reduces to the scaled canonical Euclidean inner product:
\begin{equation}
    \holinner{V}{W} = \alpha \inner{V}{W}, \quad \forall V,W \in \hol{n}.
\end{equation}
In the main paper, we assume that $\holinner{\cdot}{\cdot}$ and $\rzeroinner{\cdot}{\cdot}$ are the canonical Euclidean inner products. \cref{app:tab:olm_lsm} summarizes the Riemannian structures of OLM and LSM.
\begin{parisremark}
    $\dplus$ is also well-defined over $\sym{n}$, a surjective map $\dplus: \sym{n} \rightarrow \bbDspace{n}$. In this way, $\offexp: \sym{n} \rightarrow \cor{n}$ is no longer bijective \citep[Theorem~2.1]{thanwerdas2024permutation}. Similarly, $\dstar$ is well defined over $\spd{n}$, a surjective map $\dstar: \spd{n} \rightarrow \bbDplus{n}$. Consequently, $\logscaled: \spd{n} \rightarrow \rzero{n}$ is no longer bijective \citep[Theorem~3.5]{thanwerdas2024permutation}.
\end{parisremark}

\section{Basic Layers in SPDNet and TSMNet}
\label{app:sec:spdnet_tsmnet}
SPDNet \citep{huang2017riemannian} is a canonical SPD neural network.
SPDNet mimics the conventional densely connected feedforward network, consisting of three basic building blocks
\begin{align}
    &\text{BiMap layer: }  S^{k} = W^{k} S^{k-1} W^{k\top}, \text { with } W^k \text { semi-orthogonal,}\\
    &\text{ReEig layer: } S^{k}=U^{k-1} \max (\Sigma^{k-1}, \epsilon I_{n}) U^{k-1 \top},
    \text { with } S^{k-1}=U^{k-1} \Sigma^{k-1} U^{k-1 \top},\\
    &\text{LogEig layer: } S^{k}=\log(S^{k-1}),
\end{align}
where $\max(\cdot)$ is element-wise maximization.
BiMap and ReEig mimic transformation and non-linear activation, while LogEig maps SPD matrices into the tangent space at the identity matrix for classification.

TSMNet \citep{kobler2022spd} can be illustrated as $f_{tc} \rightarrow f_{sc} \rightarrow f_{BiMap} \rightarrow f_{ReEig} \rightarrow f_{LogEig}$, where $f_{tc}$ and $f_{sc}$ denote temporal and spatial convolution, respectively.

\section{Statistical Results of Scaling in LieBN}
\label{app:sec:result_scaling}
In this section, we will show the effect of our scaling (\cref{eq:liebn_scaling}) on the population.
We will see that while the resulting population variance generally has no closed-form expression, it becomes analytic under certain circumstances, such as SPD manifolds under LEM or LCM.
As a result, \cref{eq:liebn_scaling} can normalize and transform the latent Gaussian distribution.

To simplify, let $\phi_s(P) = \rieexp_{E} \left[ s \rielog_{E}(P) \right]$. Similar to the main paper, $\calM$ denotes a Lie group with a left-invariant metric.
First, we present a lemma on the resulting P.D.F. of a random point transformed by $\phi_s$.

\begin{parislemma}\label{lem:transform_of_pdf}
    Given a random point $X$ distributed over $\calM$ with P.D.F. $p_X$, the P.D.F. of $Y=\phi_s(X)$ is given by
    \begin{equation}
        p_Y(Q)=\Delta p_{X}(\phi_s^{(-1)}(Q) ),
    \end{equation}
    where $\Delta=\frac{|\phi^{-1}_{s*}|}{\left|L_{\phi_s^{-1}(Q)\odot Q^{-1}*} \right|}$. Here $| \cdot |$ denotes the determinant, and $\phi^{-1}_{s*}$ and $L_{\phi_s^{-1}(Q)\odot Q^{-1}*}$ are the differentials.
\end{parislemma}
\begin{proof}
    For the sake of simplicity, we will denote $\phi_s$ as $\phi$ throughout this proof.
    The volume element with respect to a left-invariant metric is the Haar measure \citep[Section~3.2]{pennec1998uniform}:
    \begin{equation}
        \diff_L \calM(P)=  \frac{d P}{|L_{P*,E}|},
    \end{equation}
    where $|L_{P*,E}|$ is the determinant\footnote{This should be understood more precisely as the determinant of the matrix representation of $L_{P*, E}$ in local coordinates.} of the differential of $L_P$ at the neutral element $E$.
    Then we have
    \begin{equation}
        \begin{aligned}
            \diff_L \calM(\phi^{-1}(Q))
            &= \frac{\diff \phi^{-1}(Q)}{|L_{\phi^{-1}(Q)*,E}|}\\
            &= |(L_{\phi^{-1}(Q) \odot Q^{-1}} \circ L_{Q})_{*,E} |^{-1} |\phi^{-1}_{*}| \diff Q\\
            &= \frac{|\phi^{-1}_{*}|}{|L_{\phi^{-1}(Q)\odot Q^{-1}*} |} \diff_{L}\calM(Q)\\
            &= \Delta \diff_{L}\calM(Q).
        \end{aligned}
    \end{equation}

    The probability of $Q=\phi(P)$ in a set $\calY \subset \calM$ is
    \begin{equation}
        \begin{aligned}
            F( \phi(P) \in \calY)
            &=F( P \in \phi^{-1} (\calY) )\\
            &=\int_{\phi^{-1} (\calY)} p_{X}(P) \cdot d_L \calM(P)\\
            &=\int_{\calY} p_{X}(\phi^{(-1)}(Q) ) d_L \calM (\phi^{(-1)}(Q))\\
            &=\int_{\calY} \Delta p_{X}(\phi^{(-1)}(Q) ) d_L \calM (Q).
        \end{aligned}
    \end{equation}
    Therefore, the density of $Y=\phi(X)$ is
    \begin{equation}
        p_Y(Q)=\Delta p_{X}(\phi^{(-1)}(Q)).
    \end{equation}
\end{proof}

The above lemma implies that when $\Delta$ is a constant, $Y$ also follows a Gaussian distribution.

\begin{pariscorollary} \label{cor:trans_gaussian}
    Following the notation in \cref{lem:transform_of_pdf}, if $\Delta=c$ is a constant and $X \sim \calN(E,\sigma^2)$, then $Y$ also follows a Gaussian distribution, \ie, $Y \sim \calN(E,s^2\sigma^2)$.
\end{pariscorollary}
\begin{proof}
    \begin{equation}
        \begin{aligned}
            p_{Y}(Q)
            &= ck(\sigma) \exp \left(-\frac{\dist(\phi_s^{-1}(Q), E)^2}{2 \sigma^2}\right)\\
            &= k'(\delta) \exp \left(-\frac{\dist(\rieexp_E \nicefrac{1}{s} \rielog_E(Q), E)^2}{2 \sigma^2}\right)\\
            &= k'(\delta) \exp \left(-\frac{\| \rielog_E(Q) \|_E^2}{2 s^2\sigma^2}\right)\\
            &= k'(\delta) \exp \left(-\frac{\dist(Q,E)^2}{2 s^2\sigma^2}\right),
        \end{aligned}
    \end{equation}
    where $\| \cdot \|_E$ is the norm of the tangent space at the neutral element $E$.
\end{proof}

\cref{cor:trans_gaussian} implies that when $\Delta=c$, $\phi_s$ can scale the population variance and further transform the Gaussian distribution.
Simple computations show that in the standard Euclidean space $\bbR{n}$, $\Delta=\nicefrac{1}{s}$.
Therefore, it is natural to expect that the pullback of $\bbR{n}$ also enjoys constant $\Delta$.

\begin{parisproposition} \label{prop:delta_pem}
    Consider an $n$-dimensional Lie group $\calM$ pulled back from the standard Euclidean space $\bbR{n}$ by the diffeomorphism $\psi:\calM \rightarrow \bbR{n}$.
    In other words, the group operations and Riemannian metric on $\calM$ are defined by $\psi$ from $\bbR{n}$.
    Then $\Delta$ remains constant on $\calM$.
\end{parisproposition}
\begin{proof}
    To simplify notation, we denote $\phi_s$ as $\phi$.
    Under the given assumption, the group addition and Riemannian metric on $\calM$ are defined as follows:
    \begin{equation}
        \begin{aligned}
            \forall P,Q \in \calM, P \odot Q &= \psi^{-1}(\psi(P)+\psi(Q))\\
            g&=\psi^{*}\geuc,
        \end{aligned}
    \end{equation}
    where $\geuc$ is the standard Euclidean metric.
    Therefore, $\phi$ can be simplified as
    \begin{equation}
        \begin{aligned}
            \phi(P)
            &=\rieexp_{E} \left [ s \rielog_{E}(P) \right]\\
            &=\psi^{-1} \left( \tilde{\rieexp}_{0} \left[\psi_{*,E} \left(s\psi^{-1}_{*,0} \tilde{\rielog}_{0}\psi(P) \right)\right] \right)\\
            &= \psi^{-1}(s\psi(P)),
        \end{aligned}
    \end{equation}
    where $\tilde{\rieexp}$ and $\tilde{\rielog}$ are the Riemannian exponential and logarithmic maps in $\bbR{n}$, which are reduced to vector addition and subtraction, respectively.
    Therefore, the inverse of $\phi$ is
    \begin{equation} \label{eq:phi_delta_simplified}
        \phi^{-1}(P) = \psi^{-1}\left(\nicefrac{1}{s}\psi(P)\right).
    \end{equation}
    Besides, $L_{\phi^{-1}(Q)\odot Q^{-1}}$ can also be further simplified:
    \begin{equation} \label{eq:lt_delta_simplified}
        L_{\phi^{-1}(Q)\odot Q^{-1}}(P) =   \psi^{-1}\left(\nicefrac{1}{s}\psi(Q) - \psi(Q) + \psi(P)\right).
    \end{equation}
    The differentials of \cref{eq:phi_delta_simplified,eq:lt_delta_simplified} at $Q$ are
    \begin{align}
        \phi^{-1}_{*,Q}&=\frac{1}{s} \psi^{-1}_{*,\nicefrac{1}{s}\psi(Q)} \circ \psi_{*,Q},\\
        L_{\phi^{-1}(Q)\odot Q^{-1}*,Q}&= \psi^{-1}_{*,\nicefrac{1}{s}\psi(Q)} \circ \psi_{*,Q}.
    \end{align}
    Therefore, $\Delta=\nicefrac{1}{s}$ for all $Q \in \calM$.
\end{proof}

By \cref{prop:delta_pem}, we can directly obtain the following corollary.
\begin{pariscorollary} \label{cor:gaussian_trans_pem}
    Given a Lie group $\calM$ pulled back from the Euclidean space, and a random point $X \sim \calN(E,\sigma^2)$ over $\calM$, $Y=\phi_s(X) \sim \calN(E,s^2\sigma^2)$.
\end{pariscorollary}

In machine learning, several Lie groups are derived by the pullback from the standard Euclidean space.
As shown by \citet{chen2024spdmlr}, $\biparamLEM$ and $\paramLCM$ are pullback metrics from the Euclidean metric.
Therefore, for the Lie groups of SPD manifolds with respect to $\biparamLEM$ and $\paramLCM$, \cref{eq:liebn_scaling} can transform the Gaussian distribution.
Specifically, given a random point $X \sim \calN(M,\sigma^2)$, \cref{eq:liebn_centering,eq:liebn_scaling,eq:liebn_biasing} transform the Gaussian distribution as
\begin{equation} \label{eq:liebn_gaussian_lem_lcm}
    \calN(M,\sigma^2) \rightarrow \calN(E,\sigma^2) \rightarrow \calN(E, s^2) \rightarrow \calN(B, s^2),
\end{equation}
where $M$ and $\sigma$ are employed to normalize $X$, and $\epsilon$ in \cref{eq:liebn_scaling} is omitted.
The above process exactly mirrors the transformation of Gaussian distributions within the framework of standard BN \citep{ioffe2015batch}.
\begin{parisremark}
    A similar result to our \cref{cor:gaussian_trans_pem} was also presented in \citet[Proposition~3]{chakraborty2020manifoldnorm}.
    However, in his proof, the author did not account for the Haar measure and only considered the P.D.F., casting doubt on the validity of their results.
    Additionally, their discussion is limited to matrix Lie groups, specifically under the distance $\dist(P,Q)= \| \mlog(P^{-1}Q)\|_{\rmF}$.
    In contrast, we rectify their proof and consider general Lie groups.
\end{parisremark}

\section{Domain-Specific Momentum LieBN for EEG Classification}
\label{app:sec:dsmliebn}

\begin{algorithm}[!t] \SetKwInOut{Input}{Input}\SetKwInOut{Output}{Output}\SetKwInOut{Parameters}{Parameters}
\caption{Momentum LieBN (MLieBN) algorithm.}
\label{alg:mliebn}
\Input{
A batch of activations $\{P_{1 \ldots N} \}$ over the Lie group $\{\calM, \odot,g\}$, and a small positive constant $\epsilon$\\
running mean $\bar{M}_r=E$, running variance $\bar{v}^2_r=1$ for training\\
running mean $\tilde{M}_r=E$, running variance $\tilde{v}^2_r=1$ for testing\\
biasing parameter $B \in \calM$, scaling parameter $s \in \bbRscalar/\{0\}$,\\
momentum for training and testing $\gamma_{train}, \gamma \in [0,1]$\\
}
\Output{Normalized activations $\{\tilde{P}_{1 \ldots N} \}$}
\BlankLine
\If{training}{
    Compute batch mean $M_b$ and variance $v_b^2$ of $\{P_{1 \ldots N}\}$;\\
    $\bar{M}_r \gets \wfm(\{1-\gamma_{train},\gamma_{train}\},\{\bar{M}_r,M_b\})$;\\
    $\bar{v}^2_r \gets (1-\gamma_{train})\bar{v}^2_r + \gamma_{train} v^2_b$;\\
    $\tilde{M}_r \gets \wfm(\{1-\gamma,\gamma\},\{\tilde{M}_r,M_b\})$;\\
    $\tilde{v}^2_r \gets (1-\gamma)\tilde{v}^2_r + \gamma v^2_b$;\\
}
\lIf{training}{$M \gets \bar{M}_r, v^2 \gets \bar{v}^2_r$} \lElse{$M \gets \tilde{M}_r, v^2 \gets \tilde{v}^2_r$}

\For{$i \gets 1$ \KwTo $N$}{
Centering to the neutral element $E$: \\
\Indp \lIf{$g$ is left-invariant}{$\bar{P}_i \gets \ltrans _{M_{\odot}^{-1}}(P_i)$} \lElse{$\bar{P}_i \gets \rtrans _{M_{\odot}^{-1}}(P_i)$} \Indm
Scaling the dispersion: \\
\hspace{1.5em}  $\hat{P}_i \gets \rieexp_{E} \left [ \frac{s}{\sqrt{v^2+\epsilon}} \rielog_{E}(\bar{P}_i) \right]$ \\
Biasing towards parameter $B$: \\
\Indp\lIf{$g$ is left-invariant}{$\tilde{P}_i \gets \ltrans _{B}(\hat{P}_i)$} \lElse{$\tilde{P}_i \gets \rtrans _{B}(\hat{P}_i)$} \Indm
}
\end{algorithm}

\begingroup\tolerance=9999\emergencystretch=4pt
\citet{kobler2022spd} proposed SPD domain-specific momentum batch normalization (SPDDSMBN) as a domain adaptation approach for EEG classification. SPDDSMBN, based on \cref{eq:kobler_rbn}, performed normalization of mean and variance on SPD manifolds under the specific AIM. Additionally, SPDDSMBN used separate momentum values for updating training and testing running statistics, inspired by the work of \citet{yong2020momentum}. Following \citet[Algorithm~1]{kobler2022spd}, we also present a momentum LieBN (MLieBN) in \cref{alg:mliebn}. Here $\gamma$ is fixed and $\gamma_{train}$ is defined as\par
\endgroup
\begin{equation}
    \gamma_{train}=1-\rho^{\frac{1}{K-1} \max (K-k, 0)}+\rho, \ \text{where}\ \rho=\frac{1}{domains\_per\_batch}.
\end{equation}

Furthermore, in line with \citet{kobler2022spd}, we adopt multi-channel mechanisms for domain-specific MLieBN (DSMLieBN), where each domain has its own MLieBN layer.
Following \citet{kobler2022spd}, we set the biasing parameter equal to the neutral element, and the scaling factor is shared across all domains.
We denote \cref{alg:mliebn} as $\operatorname{MLieBN}(P_j | M, s, \epsilon, \gamma,\gamma_{train} )$.
Then our DSMLieBN follows
\begin{equation}
    \operatorname{DSMLieBN}(P_j,i)
    =\operatorname{MLieBN}_i(P_j | E, s, \epsilon, \gamma,\gamma_{train} ), \forall P_j \in \{P_{1 \ldots N}\},
\end{equation}
where $i$ is the index of the domain. We follow the official code of SPDDSMBN\footnote{\url{https://github.com/rkobler/TSMNet}} to implement our DSMLieBN. The sole difference between DSMLieBN and SPDDSMBN is the normalization operation.

Analogous to \cref{thm:liebn_pullback}, computations for DSMLieBN under pullback metrics can also be performed by mapping, calculating, and then remapping.

\section{Backpropagation of Matrix Functions}

Our implementation of LieBN on SPD and correlation manifolds involves several matrix functions. Thus, we employ matrix backpropagation (BP) \citep{ionescu2015matrix} for gradient computation. These matrix operations can be divided into Cholesky decomposition and the functions based on eigendecomposition.

The differentiation of the Cholesky decomposition can be found in \citet[Equation~8]{murray2016differentiation} or \citet[Proposition~4]{lin2019riemannian}. Our implementation of Cholesky backpropagation produces gradients consistent with those returned by \texttt{torch.linalg.cholesky}. Therefore, during the experiments, we use \texttt{torch.linalg.cholesky}.

The second type of matrix function is based on eigendecomposition, such as matrix exponential, logarithm, and power. Although PyTorch \citep{paszke2019pytorch} supports autograd of eigendecomposition, it requires the computation of $\frac{1}{\delta_i-\delta_j}$ \citep[Proposition~1]{ionescu2015matrix}, where $\delta_i$ and $\delta_j$ denote eigenvalues. This might trigger numerical instability when $\delta_i$ and $\delta_j$ are close.
Following \citet{brooks2019riemannian}, we use the Dalecki\u{\i}--Kre\u{\i}n formula \citep[Theorem~V.3.3]{bhatia2013matrix} to calculate the BP of eigen-based matrix functions. In detail, for a matrix function defined as $X = f(S) = U f(\Sigma) U^\top$, with $S = U \Sigma U^\top$ as the eigendecomposition of an SPD matrix, its BP is expressed as
\begin{align}
    \label{eq:gradient_eigen_function} \nabla_{S} L
    &= U[K \odot(U^{T}(\nabla_{X} L) U)] U^{T},
\end{align}
where $\nabla_{X} L$ is the Euclidean gradient of the loss function $L$ with respect to $X$. Matrix $K$ is defined as
\begin{equation} \label{eq:lorenz}
    K_{i j}= \begin{cases}\frac{f\left(\sigma_{i}\right)-f\left(\sigma_{j}\right)}{\sigma_{i}-\sigma_{j}} & \text { if } \sigma_{i} \neq \sigma_{j} \\ f^{\prime}\left(\sigma_{i}\right) & \text { otherwise }\end{cases},
\end{equation}
where $\Sigma=\diag(\sigma_1,\sigma_2,\ldots,\sigma_d)$. \cref{eq:lorenz} demonstrates the numerical stability of the Dalecki\u{\i}--Kre\u{\i}n formula.

\section{Experimental Details and Additional Discussions}

\subsection{Experimental Details and Additional Discussion on the SPD Manifold}
\label{app:subsec:exp_details_spd}

\subsubsection{Data Sets and Preprocessing}
\label{app:subsubsec:spd_dataset}

The \textbf{Radar} data set \citep{brooks2019riemannian} contains 3,000 synthetic radar signals. Following the protocol of \citet{brooks2019riemannian}, each signal is split into windows of length 20, resulting in 3,000 $20 \times 20$ covariance matrices that are equally distributed across three classes.

The \textbf{HDM05} data set \citep{muller2007documentation} consists of 2,273 skeleton-based motion capture sequences executed by different actors. Each frame consists of 3D coordinates of 31 joints, allowing the representation of each sequence as a $93 \times 93$ covariance matrix. In line with \citet{brooks2019riemannian}, we trim the data set to 2,086 instances distributed across 117 classes by removing some under-represented clips.

The \textbf{FPHA} data set \citep{garcia2018first} includes 1,175 skeleton-based first-person hand gesture videos of 45 different categories with 600 clips for training and 575 for testing. Following \citet{wang2022symnet}, we represent each sequence as a $63 \times 63$ covariance matrix.

The \textbf{Hinss2021} data set \citep{hinss_eegdata_2021} is a recently released competition data set containing EEG signals for mental workload estimation. The data set is employed for two tasks, inter-session and inter-subject, which are treated as domain adaptation problems. Geometry-aware methods \citep{yair2019parallel,kobler2022spd} have demonstrated promising performance in EEG classification. We follow \citet{kobler2022spd} for data preprocessing. In detail, the Python package MOABB \citep{jayaram_moabb_2018} and MNE \citep{gramfort_meg_2013} are used to preprocess the data sets. The applied steps include resampling the EEG signals to 250/256 Hz, applying temporal filters to extract oscillatory EEG activity in the 4--36 Hz range, extracting short segments ($\leq 3$ s) associated with a class label, and finally obtaining $40 \times 40$ SPD covariance matrices.

\subsubsection{Implementation Details}
\label{app:subsubsec:impl_details}
We use the official code of SPDNetBN\footnote{\url{https://proceedings.neurips.cc/paper_files/paper/2019/file/6e69ebbfad976d4637bb4b39de261bf7-Supplemental.zip}} \citep{brooks2019riemannian} and TSMNet\footnote{\url{https://github.com/rkobler/TSMNet}} \citep{kobler2022spd} to implement our experiments on the SPDNet and TSMNet backbones. For the SPDNet architecture, we compare our LieBN with SPDNetBN \citep{brooks2019riemannian}, which applies the SPDBN (\cref{eq:spdnetbn_centering,eq:spdnetbn_biasing}) to SPDNet. Similar to SPDNetBN, we apply our LieBN after each transformation layer (BiMap layer in \cref{app:sec:spdnet_tsmnet}). In the EEG application, one of the state-of-the-art methods is TSMNet with SPD domain-specific momentum batch normalization (TSMNet+SPDDSMBN) \citep{kobler2022spd}, which is a domain adaptation version of the approach proposed by \citet{kobler2022controlling}. For a fair comparison, we also implement a domain-specific momentum LieBN, referred to as DSMLieBN (detailed in \cref{app:sec:dsmliebn}). Following \citet{kobler2022spd}, we apply our DSMLieBN before the LogEig layer (detailed in \cref{app:sec:spdnet_tsmnet}) in TSMNet. We use the standard cross-entropy loss and optimize the parameters with the Riemannian AMSGrad optimizer \citep{becigneul2019riemannian}. The network architectures are represented as $\{d_0, d_1, \ldots, d_L\}$, where the dimension of the parameter in the $i$-th BiMap layer is $d_i \times d_{i-1}$. The experiments are conducted with a learning rate of $5 \times 10^{-3}$, a batch size of 30, and 200 training epochs on the Radar, HDM05, and FPHA data sets. For the Hinss2021 data set, following \citet{kobler2022spd}, we use a learning rate of $1 \times 10^{-3}$ with a weight decay of $1 \times 10^{-4}$, a batch size of 50, and 50 training epochs.

In line with the previous work of \citet{brooks2019riemannian,kobler2022spd}, we use accuracy as the scoring metric for the Radar, HDM05, and FPHA data sets, and balanced accuracy (\ie, the average recall across classes) for the Hinss2021 data set. Ten-fold experiments on the Radar, HDM05, and FPHA data sets are carried out with randomized initialization and split (split is officially fixed for the FPHA data set), while on the Hinss2021 data set, models are fit and evaluated with a randomized leave 5\% of the sessions (inter-session) or subjects (inter-subject) out cross-validation scheme.

\subsubsection{Candidate Values of Hyperparameters}
% \textbf{Hyper-parameters: }
We implement the SPD LieBN and DSMLieBN induced by four standard invariant metrics, namely AIM, LEM, LCM, and CRIM, along with their deformed metrics.
Therefore, our method has a maximum of three hyperparameters, \ie, $(\theta,\alpha,\beta)$. As $(\alpha,\beta)$ only affect the variance calculation in the LieBN framework, we set $(\alpha,\beta)=(1,0)$ and only tune the deformation factor $\theta$ from the candidate values of $\pm 0.5$, $\pm 1$, and $\pm 1.5$. We denote [Baseline]+[BN\_Type]+[Metric]-[$\theta$] as the baseline endowed with a specific LieBN, such as SPDNet+LieBN-AIM-(1) and TSMNet+DSMLieBN-LCM-(1).

\subsubsection{Empirical Insights on the Hyperparameters}

Our SPD LieBN has at most three types of hyperparameters: Riemannian metric, deformation factor $\theta$, and $\orth{n}$-invariance parameters $\alphabeta$. The general order of importance should be Riemannian metric $>$ $\theta$ $>$ $\alphabeta$.

The most significant parameter is the choice of Riemannian metric, as all the geometric properties are sourced from a metric. A safe choice would start with AIM, and then decide whether to explore other metrics further. The most important reason is the property of affine invariance of AIM, which is a natural characteristic of covariance matrices. In our experiments, the LieBN-AIM generally achieves the best performance. However, AIM is not always the best metric. As shown in \cref{tab:results_spdnet}, the best result on the HDM05 data set is achieved by LCM-based LieBN, which improves the vanilla SPDNet by 11.71\%. Therefore, when choosing Riemannian metrics on SPD manifolds, a safe choice would start with AIM and extend to other metrics. Besides, if efficiency is an important factor, one should first consider LCM, as it is the most efficient one.

The second one is the deformation factor $\theta$. As we discussed in \cref{subsubsec:spd_param_lie_groups}, $\theta$ interpolates between different types of metrics ($\theta=1$ and $\theta \rightarrow 0$). Inspired by this, we select $\theta$ around its deformation boundaries (1 and 0). In this paper, we roughly select $\theta$ from $\{ \pm 0.5, \pm 1,\pm 1.5 \}$.

The less important parameters are $(\alpha,\beta)$. Recalling \cref{alg:liebn} and \cref{app:tab:riem_operators_props}, $(\alpha,\beta)$ only affects the calculation of variance, which should have a smaller effect than the preceding two parameters. Therefore, we simply set $(\alpha,\beta)=(1,0)$ during experiments.

\subsection{Implementation Details on the Rotation Matrix}
\label{app:subsec:impl_details_so3}

\subsubsection{Data Sets and Preprocessing}
\label{app:subsubsec:datasets_so3}
The \textbf{G3D} data set \citep{bloom2012g3d} consists of 663 sequences of 20 different gaming actions. Each sequence records the 3D locations of 20 joints (\ie, 19 bones). Following \citet{huang2017deep}, we use the cross-subject setting, where half of the subjects are used for training, and the other half for testing, respectively.

The \textbf{HDM05} data set \citep{muller2007documentation} has been discussed in \cref{app:subsubsec:spd_dataset}.

The \textbf{NTU60} data set \citep{shahroudy2016ntu} has 56,880 sequences of 3D skeleton data classified into 60 classes, where each frame contains the 3D coordinates of 25 or 50 body joints. We focus on the cross-view protocol setting \citep{shahroudy2016ntu}.

Following \citet{huang2017deep}, we use the code\footnote{\url{https://ravitejav.weebly.com/kbac.html}} of \citet{vemulapalli2014human} to represent each skeleton sequence as a point on the Lie group $\soprod{N \times T}{3}$, where $N$ and $T$ denote spatial and temporal dimensions. Following the preprocessing of \citet{huang2017deep}, we set $T$ to 100, 64, and 16 for the three data sets, respectively.

\subsubsection{LieNet}
The LieNet consists of three basic layers: RotMap, RotPooling, and LogMap layers.
The RotMap mimics the convolutional layer, while the RotPooling extends the pooling layers to rotation matrices. The LogMap layer maps the rotation matrix into the tangent space at the identity for classification. Note that the official code of LieNet\footnote{\url{https://github.com/zhiwu-huang/LieNet}} is developed in MATLAB. We use the open-source PyTorch code\footnote{\url{https://github.com/hjf1997/LieNet}} to implement our experiments.
To reproduce LieNet more faithfully, we made the following modifications to this PyTorch code. We recoded the LogMap and RotPooling layers to make them consistent with the official MATLAB implementation. In addition, we also extend the Riemannian computation of geoopt \citep{becigneul2019riemannian} into $\so{3}$ to allow for a Riemannian optimizer on $\so{3}$, which is missing in the current package. We apply our LieBN before the LogMap layer. Note that the dimension of input features in LieNet is $B \times N \times T \times 3 \times 3$. We calculate Lie group statistics along the batch and temporal dimensions ($B \times T$). We denote the LieNet models with our LieBN-Left and LieBN-Right as LieNetLieBN-Left and LieNetLieBN-Right, respectively.

\subsubsection{Training Details}
We find that SGD is the most effective optimizer for LieNet, and thus, we adopt it for our experiments. The learning rate is set to $1 \times 10^{-2}$. The batch sizes are 30, 30, and 256 for the G3D, HDM05, and NTU60 data sets, respectively. On the NTU60 data set, the learning rate is reduced by a factor of 10 upon model convergence, specifically at the 5th and 25th epochs for LieNetLieBN and LieNet, respectively. For each model, we apply \texttt{torch.nn.utils.clip\_grad\_norm\_} with \texttt{max\_norm=5} to the transformation matrix in the final FC layer.

\subsection{Implementation Details on the Correlation Matrix}
\label{app:subsec:impl_details_cor}

We follow the same settings as the experiments on the SPD manifold with respect to the backbone architecture, batch size, number of training epochs, optimizer, and learning rate. The network architecture can be denoted as BiMap-[Power-Cov2Cor-LieBN-Cor]-LogEig, where Power denotes the matrix power and Cov2Cor is $\coropt(\cdot): \spd{n} \to \cor{n}$. The matrix powers used for each data set are presented in \cref{app:tab:liebncor_power}. A single iteration for computing $\dplus$ and $\dstar$ is sufficient to achieve saturated network performance in OLM and LSM, except for $\dstar$ on the HDM05 data set, which requires up to 20 iterations to converge.

\begin{table}[tbp]
  \centering
    \begin{tabular}{c|cccc}
    \toprule
    \diagbox{Data Set}{Metric}      & ECM   & LECM  & OLM   & LSM \\
    \midrule
    HDM05 & 0.75  & 0.5   & 0.5   & -0.5 \\
    FPHA  & -0.5  & -0.25 & -0.25 & -0.25 \\
    \bottomrule
    \end{tabular}%
  \caption{Matrix powers in LieBN-Cor under different metrics on each data set.}
  \label{app:tab:liebncor_power}
\end{table}%

\subsection{Hardware}
All experiments use an Intel Core i9-7960X CPU with 32 GB RAM and an NVIDIA GeForce RTX 2080 Ti GPU.

\section{Proofs} \label{app:sec:proofs}

\linkofproof{props:invariance_correlation}
\subsection{Proof of \texorpdfstring{\cref{props:invariance_correlation}}{}}
\label{prf:props:invariance_correlation}

\begin{proof}
    \mypara{Invariance.}
    First, we note that all four Lie groups are commutative. Secondly, the construction of all four metrics is similar. It suffices to show the left-invariance of ECM.

    We only need to show that $L_{Q}: \cor{n} \to \cor{n}$ for any $Q \in \cor{n}$ is a Riemannian isometry. $L_{Q}$ can be rewritten as
    \begin{equation}
        \begin{aligned}
            L_{Q} = \Theta^{-1} \circ \widetilde{L}_{\Theta(Q)} \circ \Theta,
        \end{aligned}
    \end{equation}
    where $\widetilde{L}_{\Theta(Q)}$ is the left translation over $\LTone{n}$, which is an isometry over $\LTone{n}$. As $\Theta^{-1}$, $\widetilde{L}_{\Theta(Q)}$, and $\Theta$ are all isometries, their composition is an isometry as well.

    \mypara{WFM.}
    The WFM in the Euclidean space is reduced to an arithmetic weighted average. By the isometry of $f$, one can directly obtain the results.
\end{proof}

\linkofproof{props:population_gaussian}
\subsection{Proof of \texorpdfstring{\cref{props:population_gaussian}}{}}
\label{prf:population_gaussian}

\begin{proof}
    \mypara{\cref{pro:mle_m}.}
    The MLE of $M$ is
    \begin{equation}
        \begin{aligned}
            M_{\mle}
            &= \argmax \log(K(v)) - \sum_{i=1}^N \frac{\dist(P_i,M)^2}{2v^2}\\
            &= \argmin \sum_{i=1}^N \dist(P_i,M)^2.
        \end{aligned}
    \end{equation}

    \mypara{\cref{pro:hom}.}
    We denote $Y=\ltrans_B(X)$, and $p_X$ and $p_Y$ as the densities of $X$ and $Y$, respectively.
    The density of  $Y$ is
    \begin{equation}
        \begin{aligned}
            p_{Y}(Q)
            &\stackrel{(1)}{=}p_X(\ltrans _{B^{-1}_{\odot}}(Q))\\
            &= k(\sigma) \exp \left(-\frac{\dist(\ltrans_{B^{-1}_{\odot}}(Q), M)^2}{2 \sigma^2}\right)\\
            &\stackrel{(2)}{=} k(\sigma) \exp \left(-\frac{\dist(Q, \ltrans_{B}(M))^2}{2 \sigma^2}\right).
    \end{aligned}
    \end{equation}
    The above comes from:
    \begin{enumerate}[label=(\arabic*)]
        \item
        \citet[Theorem~7]{pennec2004probabilities};
        \item
        The isometry of the left translation.
    \end{enumerate}
\end{proof}

\linkofproof{props:samples}
\subsection{Proof of \texorpdfstring{\cref{props:samples}}{}}
\label{prf:samples}

\begin{proof}
    The isometry of $\ltrans _B$ directly implies the homogeneity of the sample mean.
    Now let us focus on \cref{eq:variance_lie_group}.
    We have the following:
    \begin{equation}
        \begin{aligned}
            \sum\nolimits_{i=1}^N  w_i \dist^2(\phi_{s}(P_i), E)
            &= \sum\nolimits_{i=1}^N w_i \|s\rielog_E P_i\|_E^2\\
            &= s^2\sum\nolimits_{i=1}^N w_i \|\rielog_E P_i\|^2_E\\
            &= s^2\sum\nolimits_{i=1}^N w_i \dist^2(P_i, E),
    \end{aligned}
    \end{equation}
    where $\| \cdot \|_{E}$ is the norm on $T_E\calM$.
\end{proof}

\linkofproof{props:core_operation_right}
\subsection{Proof of \texorpdfstring{\cref{props:core_operation_right}}{}}
\label{prf:core_operation_right}

\begin{proof}
    As the right-invariant metric is similar to the left-invariant one, this proof follows logic similar to that of the above two proofs.

    \mypara{Gaussian homogeneity.}
    We denote $Y=\rtrans_B(X)$, and $p_X$ and $p_Y$ as the densities of $X$ and $Y$, respectively.
    The density of  $Y$ is
    \begin{equation}
        \begin{aligned}
            p_{Y}(Q)
            &\stackrel{(1)}{=}p_X(\rtrans _{B^{-1}_{\odot}}(Q))\\
            &= k(\sigma) \exp \left(-\frac{\dist(\rtrans_{B^{-1}_{\odot}}(Q), M)^2}{2 \sigma^2}\right)\\
            &\stackrel{(2)}{=} k(\sigma) \exp \left(-\frac{\dist(Q, \rtrans_{B}(M))^2}{2 \sigma^2}\right).
    \end{aligned}
    \end{equation}
    The above comes from:
    \begin{enumerate}[label=(\arabic*)]
        \item
        \citet[Theorem~13]{pennec2004probabilities};
        \item
        The isometry of the right translation.
    \end{enumerate}

    \mypara{Sample mean homogeneity.} This is a direct corollary of the isometry of right translation.
\end{proof}

\linkofproof{prop:liebn_natural_extension_ebn}
\subsection{Proof of \texorpdfstring{\cref{prop:liebn_natural_extension_ebn}}{}}
\label{prf:liebn_natural_extension_ebn}

\begin{proof}
    As $\bbR{n}$ is an abelian group and the Euclidean inner product is bi-invariant, we focus on left-translation in the following.
    The core of this proof lies in the fact that on $\bbR{n}$,
    (1) the Fréchet mean and variance are reduced to the familiar Euclidean statistics.
    (2) the calculation of the running mean becomes the weighted arithmetic mean.
    (3) \cref{eq:liebn_centering,eq:liebn_scaling,eq:liebn_biasing} become \cref{eq:ebn}.
    We prove these three points one by one.

    As stated by \citet[Proposition~G.1 and Corollary~G.2]{lou2020differentiating}, from the view of the product manifold, the elementwise Fréchet mean and variance on $\bbR{n}$ are equivalent to the vector-valued Euclidean variance and mean.

    Besides, by a similar proof to \citet[Proposition~G.1]{lou2020differentiating}, the weighted Fréchet mean on $\bbR{n}$ is simplified as the weighted arithmetic average.
    Therefore, on $\bbR{n}$, the calculation of running statistics in our \cref{alg:liebn} becomes the familiar moving average.

    Thirdly, on $\bbR{n}$, we know that $\ltrans _x(y)=x+y$, $\rieexp_x v=x+v$, $\rielog_x y=y-x$, and the neutral element is $0$.
    Since statistics, as well as the Euclidean BN, are calculated elementwise, we can safely consider a single element, \ie, $\bbR{n}=\bbRscalar$.
    For a batch of activations $\{x_i\}_{i=1}^N \subset \bbRscalar$, where the batch mean and batch variance are denoted as $\mu_b$ and $v^2_b$, \cref{eq:liebn_centering,eq:liebn_scaling,eq:liebn_biasing} can be rewritten as
    \begin{equation}
        L_{\beta}\left( \rieexp_0 \left[\frac{\gamma}{\sqrt{v^2_{b}+\epsilon}}\rielog_0(\ltrans_{-\mu_b}(x_i)) \right]\right)
        = \gamma \frac{x_i-\mu_{b}}{\sqrt{v^2_{b}+\epsilon}} + \beta.
    \end{equation}
    The above equation is the exact core computation of the standard Euclidean BN.
\end{proof}

\linkofproof{prop:spd_param_lem_lcm_deformation}
\subsection{Proof of \texorpdfstring{\cref{prop:spd_param_lem_lcm_deformation}}{}}
\label{prf:spd_param_lem_lcm_deformation}

\begin{proof}
    We first prove the case of $\triparamLEM$, and then proceed to the case of $\paramLCM$.

    \mypara{$\triparamLEM$.}
    For clarity, we denote the metric tensor of $\triparamLEM$ as
    \begin{equation}
        \gtriparamLE = \frac{1}{\theta^2}\pow_\theta^*\gbiparamLE,
    \end{equation}
    where $\gbiparamLE$ is the metric tensor of $\biparamLEM$.
    Let $P \in \spd{n}$ and $V,W \in T_P\spd{n}$, then we have
    \begin{equation}
        \begin{aligned}
            \gtriparamLE_P(V,W)
            &= \frac{1}{\theta^2} \gbiparamLE_{\pow_{\theta}(P)} \left(\pow_{\theta*,P}(V),\pow_{\theta*,P}(W) \right)\\
            &= \frac{1}{\theta^2} \langle \left( \mlog \circ \pow_{\theta} \right)_{*,P} (V), \left( \mlog \circ \pow_{\theta} \right)_{*,P} (W) \rangle^{\alphabeta}\\
            &= \langle \mlog_{*,P} (V), \mlog_{*,P} (W) \rangle^{\alphabeta}\\
            &= \gbiparamLE_P(V,W).
        \end{aligned}
    \end{equation}
% $\gdefLC = \frac{1}{\theta^2}\pow_\theta^*\gLC$

    \mypara{$\paramLCM$.}
    Let us first review a well-known fact of deformed metrics \citep{thanwerdas2022geometry}.
    Let $\tilde{g}=\frac{1}{\theta^2}\pow_{\theta}^*g$ be the power-deformed metric on the SPD manifold.
    Then when $\theta$ tends to 0, for all $P \in \spd{n}$ and all $V \in T_P\spd{n}$, we have
    \begin{equation} \label{eq:deformed_g_lim}
        \tilde{g}_P(V,V) \to g_{I}(\log_{*,P}(V),\log_{*,P}(V)).
    \end{equation}

    By \cref{eq:deformed_g_lim}, we can readily obtain the results.
\end{proof}

\linkofproof{prop:spd_param_invariance}
\subsection{Proof of \texorpdfstring{\cref{prop:spd_param_invariance}}{}}
\label{prf:spd_param_invariance}

\begin{proof}
    $\biparamAIM$ is left-invariant \citep{thanwerdas2022theoretically}.
    As the pullback of $\biparamAIM$, $\triparamAIM$ is left-invariant as well.
    Besides, \citet{chen2024adaptive} shows that LCM is the pullback metric from the Euclidean space of $\trilspace{n}$.
    Therefore, $\theta$-LCM is bi-invariant.
\end{proof}

\linkofproof{thm:crim}
\subsection{Proof of \texorpdfstring{\cref{thm:crim}}{}}
\label{prf:crim}

\begin{proof}
    In the following, we denote $\dist^{\mrL}$, $\rielog^{\mrL}$, and $\rieexp^{\mrL}$ as the Riemannian operators under the left-invariant metric, \ie, AIM. Note that the Cholesky decomposition pulls back the group operation of matrix product from the Cholesky manifold $\chospace{n}$ \citep{thanwerdas2022theoretically}. For simplicity, we abbreviate $\odotai$ as $\odot$.

    Let us first review the differential map of Cholesky decomposition and its inverse \citep[Proposition~4]{lin2019riemannian}. Following the notation in this proposition and further denoting $X \in T_L\chospace{n}$, we have the following
    \begin{align}
        \label{eq:diff_chol}
        \chol_{*,P} (V) &= L\left(L^{-1} V L^{-\top}\right)_{\frac{1}{2}},\\
        \label{eq:diff_chol_inv}
        (\chol^{-1})_{*,L} (X) &= L X^{\top}+X L^{\top}.
    \end{align}
    Specifically, for the differential map at $I$, we have
    \begin{align}
        \label{eq:diff_chol_at_i}
         \chol_{*,I} (V) &= \left( V \right) _{\frac{1}{2}}, \forall V \in T_I\spd{n}.\\
         \label{eq:diff_chol_inv_at_i}
         (\chol^{-1})_{*,I} (X) &= \symmetrize{X}, \forall X \in T_I\chospace{n}.
    \end{align}
    Denoting $\widetilde{\ltrans}$ and $\widetilde{\rtrans}$ as the group translations on the Cholesky manifold $\chospace{n}$, we have the following with respect to the differential maps of left and right translation:
    \begin{equation}
        \label{eq:diff_ltrans}
        \begin{aligned}
            \left(\ltrans_{P,* \inv{P}} \right)^{-1}
            & \stackrel{(1)}{=} \left( \left( \chol^{-1} \right)_{*, I}
            \widetilde{\ltrans}_{L *, L^{-1}} \circ \chol_{*, \inv{P}} \right)^{-1}\\
            & = \left( \chol_{*, \inv{P}} \right)^{-1} \left( \widetilde{\ltrans}_{L *, L^{-1}} \right)^{-1} \circ \chol_{*, I}.
        \end{aligned}
    \end{equation}
    \begin{equation}
        \label{eq:diff_rtrans}
        \rtrans_{\inv{P} *,P}
        \stackrel{(2)}{=} \left(\chol^{-1}\right)_{*, I} \circ \widetilde{\rtrans}_{L^{-1} *, L} \circ \chol_{*, P}.
    \end{equation}
    The above derivation comes from the following:
    \begin{enumerate}[label=(\arabic*)]
        \item
        $\ltrans _P =  \chol^{-1} \circ \widetilde{\ltrans}_L \circ \chol$;
        \item
        $\rtrans _{\inv{P}} =  \chol^{-1} \circ \widetilde{\rtrans}_{L^{-1}} \circ \chol$.
    \end{enumerate}
    \mypara{Riemannian metric.}
    For the differential of right translation, we have the following
    \begin{equation}
        \label{eq:diff_rtrans_spd}
        \begin{aligned}
            \rtrans_{\inv{P} *,P} (V)
            & = \left(\chol^{-1}\right)_{*, I} \circ \widetilde{\rtrans}_{L^{-1} *, L} \circ \chol_{*, P} (V) \\
            & = \symmetrize{ L(L^{-1} V L^{-\top})_{\frac{1}{2}} L^{-1}}. \\
        \end{aligned}
    \end{equation}
    By \cref{eq:diff_rtrans_spd}, one can obtain the expression for the Riemannian metric tensor.

    \mypara{Riemannian geodesic and exponential map.}
    According to \citet{zacur2014left}, we have the following for the operators between left- and right-invariant metrics:
    \begin{align}
        \label{eq:exp_l2r_start}
        \rieexp _{P} (V) &= \left\{ \rieexp^{\mrL}_ {P^{-1}_{\odot}} \left( -\left(\ltrans_{P,*P^{-1}_{\odot}} \right)^{-1} \circ \rtrans _{P^{-1}_{\odot}*,P} (V) \right) \right\}^{-1}_{\odot},\\
        \label{eq:dist_l2r}
        \dist(P,Q) &= \dist^{\mrL} \left( \inv{P}, \inv{Q}\right).
    \end{align}
    Putting the AIM-based geodesic distance into the RHS of \cref{eq:dist_l2r}, one can obtain the geodesic distance under CRIM.

    Now, we simplify \cref{eq:exp_l2r_start}.
    Putting \cref{eq:diff_ltrans,eq:diff_rtrans} into \cref{eq:exp_l2r_start}, we have the following:
    \begin{equation} \label{eq:exp_crim_derivation}
        \begin{aligned}
        \rieexp_{P}(V)
        & = \left\{ \rieexp^{\mrL}_{\inv{P}} \left(- \left( \chol_{*, \inv{P}} \right)^{-1} \circ \left( \widetilde{\ltrans}_{L *, L^{-1}} \right)^{-1} \circ \widetilde{\rtrans}_{L^{-1} *, L} \circ \chol_{*, P} (V) \right) \right\}_{\odot} ^{-1} \\
        & = \left\{ \rieexp^{\mrL}_{\inv{P}} \left(- \left( \chol_{*, \inv{P}} \right)^{-1} \left[ L^{-1}  \chol_{*, P} (V) L^{-1} \right] \right) \right\}_{\odot} ^{-1} \\
        & \stackrel{(1)}{=} \left\{ \rieexp^{\mrL}_{\inv{P}} \left(- \left( \chol_{*, \inv{P}} \right)^{-1} \left[ \left(L^{-1} V L^{-\top}\right)_{\frac{1}{2}} L^{-1} \right] \right) \right\}_{\odot} ^{-1} \\
        & \stackrel{(2)}{=} \left\{ \rieexp^{\mrL}_{\inv{P}} \left(- \symmetrize{ \left(L^{-1} V L^{-\top}\right)_{\frac{1}{2}} L^{-1} L^{-\top}} \right) \right\}_{\odot} ^{-1}.\\
        \end{aligned}
    \end{equation}
    The above comes from the following:
    \begin{enumerate}[label=(\arabic*)]
        \item
        \cref{eq:diff_chol};
        \item
        \cref{eq:diff_chol_inv}.
    \end{enumerate}

    \mypara{Riemannian logarithm.}
    From the second equality in \cref{eq:exp_crim_derivation}, we have the following
    \begin{equation} \label{eq:log_crim_derivation}
        \begin{aligned}
            \rielog _P (Q)
            &= - \chol_{*, P}^{-1} \left\{ L\chol_{*, \inv{P}} \left(\rielog ^\mrL _{\inv{P}} \left(\inv{Q} \right) \right) L \right\}\\
            &\stackrel{(1)}{=} - \chol_{*, P}^{-1} \left\{ \left(L \widetilde{V} L^\top \right)_{\frac{1}{2}} L \right\}\\
            &= -\symmetrize{L L^{\top} \left( L \widetilde{V}L^{\top} \right)_{\frac{1}{2}}^{\top}}.
        \end{aligned}
    \end{equation}
    The above comes from the following:
    \begin{enumerate}[label=(\arabic*)]
        \item
        $\chol_{*,\inv{P}} (V) = L^{-1} \left(L V L^\top\right)_{\frac{1}{2}}, \forall V \in T_{\inv{P}}\spd{n}$.
    \end{enumerate}
\end{proof}

\linkofproof{cor:crim_geodesic}
\subsection{Proof of \texorpdfstring{\cref{cor:crim_geodesic}}{}}
\label{prf:crim_geodesic}

\begin{proof}
    \mypara{Completeness.}
    \cref{eq:exp_crim} indicates that $\rieexp _{I}$ is defined over the whole $T _I \spd{n}$. Besides, the SPD manifold is connected \citep{pennec2006riemannian}. By \citet[Corollary~6.20]{lee2018introduction}, CRIM is complete.

    \mypara{Geodesic.}
    For simplicity, we abbreviate $\odotai$ as $\odot$. The geodesic connecting $P$ and $Q$ can be obtained by the following:
    \begin{equation}
        \begin{aligned}
             \gamma{(t; P,Q)}
             &= \rieexp _{P} \left( t\rielog _{P} (Q)\right)\\
             &\stackrel{(1)}{=} \left\{ \rieexp^{\mrL}_{\inv{P}} \left(- \left( \chol_{*, \inv{P}} \right)^{-1} \left[ L^{-1}  \chol_{*, P} \left(t \rielog _P (Q) \right) L^{-1} \right] \right) \right\}_{\odot} ^{-1} \\
             &\stackrel{(2)}{=} \left\{ \rieexp^{\mrL}_{\inv{P}} \left( t\rielog ^\mrL _{\inv{P}} \left(\inv{Q} \right) \right) \right\}_{\odot} ^{-1} \\
             &= \left\{ \gamma^{\mathrm{AI}}(t; \widetilde{P},\widetilde{Q}) \right\}_{\odot} ^{-1}. \\
        \end{aligned}
    \end{equation}
    The above comes from the following:
    \begin{enumerate}[label=(\arabic*)]
        \item
        The second equality in \cref{eq:exp_crim_derivation};
        \item
        The first equality in \cref{eq:log_crim_derivation}.
    \end{enumerate}
\end{proof}

\linkofproof{thm:liebn_pullback}
\subsection{Proof of \texorpdfstring{\cref{thm:liebn_pullback}}{}}
\label{prf:liebn_pullback}

\begin{proof}
    Without loss of generality, we focus on the case of the left-invariant metric. The results for the right-invariant metric can be proven similarly.

     We denote \cref{eq:liebn_centering,eq:liebn_scaling,eq:liebn_biasing} on $\calM_k$, $k \in \{1,2\}$, as the mapping $\xi^k(\cdot \mid M,v^2,B,s,\epsilon)$.
     Let $\calB=\{P_i\}_{i=1}^N$ and $f(\calB)=\{f(P_i)\}_{i=1}^N$. Throughout the proof, $f$ acts pointwise on finite collections.

    The core of this proof lies in three points:
    \begin{enumerate}
        \item
        The Fréchet mean and variance of $\calB$ in $\calM_1$ correspond to the counterparts of $f(\calB)$ in $\calM_2$.
        \item
        $\xi^1(P_i \mid M,v^2,B,s,\epsilon)$ in $\calM_1$ is equal to $f^{-1} (\xi^2(f(P_i) \mid f(M),v^2,f(B),s,\epsilon))$.
        \item
        The updates of running statistics in $\calM_1$ correspond to the counterparts in $\calM_2$.
    \end{enumerate}

    We denote $M$ as the Fréchet mean of $\calB$, and $v^2$ as the Fréchet variance of $\calB$.
    Then, by the isometry of $f$, the Fréchet mean and variance of $f(\calB)$ are $f(M)$ and $v^2$, respectively.

    On $\calM_k$, $k \in \{1,2\}$, we denote $\ltrans^k, \odot^k, \rieexp^k,\rielog^k$ as the Lie group and Riemannian operators, $E^k$ as the neutral element, and the scaling map in \cref{eq:liebn_scaling} as $\phi^k_a(\cdot)$ with $a=s/\sqrt{v^2+\epsilon}$.
    With the isometry and Lie group isomorphism of $f$, we have the following equations:
    \begin{equation}
        \ltrans^1_{M_{\odot^1}^{-1}}
        =f^{-1} \circ \ltrans^2_{\left(f(M)\right)_{\odot^2}^{-1}} \circ f.
    \end{equation}
    \begin{equation}
        \begin{aligned}
            \phi^1_{a} &=\rieexp^1_{E^1}\left[ a \rielog^1_{E^1}(\cdot) \right]\\
            &= f^{-1} \left( \rieexp^{2}_{E^2}\left[ a \rielog^2_{E^2}(f(\cdot))\right]\right)\\
            &= f^{-1} \circ \phi^2_a \circ f.
        \end{aligned}
    \end{equation}
    \begin{equation}
        \ltrans^1_{B}
        =f^{-1} \circ \ltrans^2_{f(B)} \circ f.
    \end{equation}
    Then we have
    \begin{equation} \label{eq:core_ops_pm}
        \xi^1(P_i \mid M,v^2,B,s,\epsilon)=f^{-1}(\xi^2(f(P_i) \mid f(M),v^2,f(B),s,\epsilon)).
    \end{equation}

    Lastly, we show the correspondence between running statistics.
    Since the Fréchet variance is the same for both $\calB$ and $f(\calB)$, we focus on the running mean.
    Let $M_r$ and $f(M_r)$ denote the initial values of the running means in $\calM_1$ and $\calM_2$, respectively, and let $\wfm^k$ represent the weighted Fréchet mean in $\calM_k$ for $k \in \{1,2\}$.
    Then the updated running mean in $\calM_1$ is
    \begin{equation} \label{eq:wfm_pm}
        \wfm^1(\{1-\gamma,\gamma\},\{M_r,M\})=
        f^{-1}(\wfm^2(\{1-\gamma,\gamma\},\{f(M_r),f(M)\})).
    \end{equation}

    Denoting $\liebn^k$ as the LieBN algorithm on $\calM_k$ for $k \in \{1,2\}$, \cref{eq:core_ops_pm} and \cref{eq:wfm_pm} imply that
    \begin{equation}
    \liebn^1(P_i;B,s,\epsilon,\gamma)=f^{-1} \left[\liebn^2(f(P_i);f(B),s,\epsilon,\gamma) \right].
    \end{equation}
\end{proof}

\bibliography{ref}

@inproceedings{chen2025understanding,
    title={Understanding Matrix Function Normalizations in Covariance Pooling through the Lens of {Riemannian} Geometry},
    author={Ziheng Chen and Yue Song and Xiaojun Wu and Gaowen Liu and Nicu Sebe},
    booktitle={ICLR},
    year={2025}
}

@inproceedings{chen2026product,
  title={Fast and Stable {Riemannian} Metrics on {SPD} Manifolds via {Cholesky} Product Geometry},
  author={Ziheng Chen and Yue Song and Xiaojun Wu and Nicu Sebe},
  booktitle={ICLR},
  year={2026}
}

@inproceedings{hu2026riemannian,
  title={Riemannian High-Order Pooling for Brain Foundation Models},
  author={Chen Hu and Ziheng Chen and Rui Wang and Yefeng Zheng and Nicu Sebe},
  booktitle={ICLR},
  year={2026}
}

@inproceedings{chen2024rmlr,
    title={{RMLR}: Extending Multinomial Logistic Regression into General Geometries},
    author={Chen, Ziheng and Song, Yue and Wang, Rui and Wu, Xiao-Jun and Sebe, Nicu},
    booktitle={NeurIPS},
    year={2024}
}

@inproceedings{chen2024liebn,
    title={A {Lie} Group Approach to {Riemannian} Batch Normalization},
    author={Ziheng Chen and Yue Song and Yunmei Liu and Nicu Sebe},
    booktitle={ICLR},
    year={2024}
}

@inproceedings{chen2024spdmlr,
    title={Riemannian Multiclass Logistics Regression for {SPD} Neural Networks},
    author={Chen, Ziheng and Song, Yue and Liu, Gaowen and Kompella, Ramana Rao and Wu, Xiaojun and Sebe, Nicu},
    booktitle={CVPR},
    year={2024}
}

@article{chen2024adaptive,
  title={Adaptive {Log-Euclidean} metrics for {SPD} matrix learning},
  author={Chen, Ziheng and Song, Yue and Xu, Tianyang and Huang, Zhiwu and Wu, Xiao-Jun and Sebe, Nicu},
  journal={IEEE TIP},
  year={2024}
}

@inproceedings{chen2023riemannian,
  title={Riemannian Local Mechanism for {SPD} Neural Networks},
  author={Chen, Ziheng and Xu, Tianyang and Wu, Xiao-Jun and Wang, Rui and Huang, Zhiwu and Kittler, Josef},
  booktitle={AAAI},
    year={2023}
}

@article{chen2023hybrid,
  title={Hybrid {R}iemannian Graph-Embedding Metric Learning for Image Set Classification},
  author={Chen, Ziheng and Xu, Tianyang and Wu, Xiao-Jun and Wang, Rui and Kittler, Josef},
  journal={IEEE Transactions on Big Data},
  volume={9},
  number={1},
  pages={75--92},
  year={2023},
  doi={10.1109/TBDATA.2021.3113084}
}

@inproceedings{wang2022dreamnet,
  title={{DreamNet}: A Deep {R}iemannian Manifold Network for {SPD} Matrix Learning},
  author={Wang, Rui and Wu, Xiao-Jun and Chen, Ziheng and Xu, Tianyang and Kittler, Josef},
  booktitle={ACCV},
    year={2022}
}

@article{wang2022learning,
  title={Learning a discriminative {SPD} manifold neural network for image set classification},
  author={Wang, Rui and Wu, Xiao-Jun and Chen, Ziheng and Xu, Tianyang and Kittler, Josef},
  journal={Neural Networks},
  volume={151},
  pages={94--110},
  year={2022},
  publisher={Elsevier},
  }

@article{wangspdmetric2024,
  author={Wang, Rui and Wu, Xiao-Jun and Chen, Ziheng and Hu, Cong and Kittler, Josef},
  journal={IEEE TNNLS},
  title={{SPD} Manifold Deep Metric Learning for Image Set Classification},
  year={2024}
  }

@inproceedings{wang2024grassatt,
  title = {A {Grassmannian} Manifold Self-Attention Network for Signal Classification},
  author = {Wang, Rui and Hu, Chen and Chen, Ziheng and Wu, Xiao-Jun and Song, Xiaoning},
  booktitle = {IJCAI},
  year = {2024},
}

@inproceedings{yim2023se,
  title={{SE} (3) diffusion model with application to protein backbone generation},
  author={Yim, Jason and Trippe, Brian L and De Bortoli, Valentin and Mathieu, Emile and Doucet, Arnaud and Barzilay, Regina and Jaakkola, Tommi},
  booktitle={ICML},
  year={2023}
}

@inproceedings{de2022riemannian,
  title={Riemannian score-based generative modelling},
  author={De Bortoli, Valentin and Mathieu, Emile and Hutchinson, Michael and Thornton, James and Teh, Yee Whye and Doucet, Arnaud},
  booktitle={NeurIPS},
  year={2022}
}

@inproceedings{manton2004globally,
  title={A globally convergent numerical algorithm for computing the centre of mass on compact {Lie} groups},
  author={Manton, Jonathan H},
  booktitle={The 8th Control, Automation, Robotics and Vision Conference, 2004.},
  volume={3},
  pages={2211--2216},
  year={2004},
  organization={IEEE},
  }

@inproceedings{barbaresco2021gaussian,
  title={Gaussian distributions on the space of symmetric positive definite matrices from {Souriau’s Gibbs} state for {Siegel} domains by coadjoint orbit and moment map},
  author={Barbaresco, Fr{\'e}d{\'e}ric},
  booktitle={Geometric Science of Information: 5th International Conference},
  year={2021}
  }

@inproceedings{thanwerdas2019affine,
  title={Is affine-invariance well defined on {SPD} matrices? A principled continuum of metrics},
  author={Thanwerdas, Yann and Pennec, Xavier},
  booktitle={Geometric Science of Information: 4th International Conference},
  year={2019}
}

@inproceedings{thanwerdas2019exploration,
  title={Exploration of balanced metrics on symmetric positive definite matrices},
  author={Thanwerdas, Yann and Pennec, Xavier},
  booktitle={Geometric Science of Information: 4th International Conference, GSI 2019, Toulouse, France, August 27--29, 2019, Proceedings 4},
  pages={484--493},
  year={2019},
    organization={Springer}
}

@inproceedings{vemulapalli2014human,
  title={Human action recognition by representing {3D} skeletons as points in a {Lie} group},
  author={Vemulapalli, Raviteja and Arrate, Felipe and Chellappa, Rama},
  booktitle={CVPR},
    year={2014}
}

@inproceedings{huang2017deep,
  title={Deep learning on {Lie} groups for skeleton-based action recognition},
  author={Huang, Zhiwu and Wan, Chengde and Probst, Thomas and Van Gool, Luc},
  booktitle={CVPR},
    year={2017}
}

@inproceedings{garcia2018first,
  title={First-person Hand Action Benchmark with {RGB-D} Videos and 3{D} Hand Pose Annotations},
  author={Garcia-Hernando, Guillermo and Yuan, Shanxin and Baek, Seungryul and Kim, Tae-Kyun},
  booktitle={CVPR},
    year={2018}
}

@inproceedings{ionescu2015matrix,
  title={Matrix Backpropagation for Deep Networks with Structured Layers},
  author={Ionescu, Catalin and Vantzos, Orestis and Sminchisescu, Cristian},
  booktitle={ICCV},
    year={2015}
}

@inproceedings{ioffe2015batch,
  title={Batch normalization: Accelerating deep network training by reducing internal covariate shift},
  author={Ioffe, Sergey and Szegedy, Christian},
  booktitle={ICML},
  year={2015}
  }

@inproceedings{he2016deep,
  title={Deep Residual Learning for Image Recognition},
  author={He, Kaiming and Zhang, Xiangyu and Ren, Shaoqing and Sun, Jian},
  booktitle={CVPR},
  year={2016}
}

@inproceedings{huang2017riemannian,
  title={A {R}iemannian Network for {SPD} Matrix Learning},
  author={Huang, Zhiwu and Van Gool, Luc},
  booktitle={AAAI},
  year={2017}
}

@inproceedings{becigneul2019riemannian,
  title={Riemannian adaptive optimization methods},
  author={B{\'e}cigneul, Gary and Ganea, Octavian-Eugen},
  booktitle={ICLR},
    year={2019}
}

@inproceedings{bloom2012g3d,
  title={{G3D}: A gaming action dataset and real time action recognition evaluation framework},
  author={Bloom, Victoria and Makris, Dimitrios and Argyriou, Vasileios},
  booktitle={CVPR Workshops},
  year={2012},
  }

@inproceedings{ganea2018hyperbolic,
  title={Hyperbolic neural networks},
  author={Ganea, Octavian and B{\'e}cigneul, Gary and Hofmann, Thomas},
  booktitle={NeurIPS},
  year={2018}
}

@inproceedings{huang2018building,
  title={Building deep networks on {G}rassmann manifolds},
  author={Huang, Zhiwu and Wu, Jiqing and Van Gool, Luc},
  booktitle={AAAI},
    year={2018}
}

@inproceedings{yong2020momentum,
  title={Momentum batch normalization for deep learning with small batch size},
  author={Yong, Hongwei and Huang, Jianqiang and Meng, Deyu and Hua, Xiansheng and Zhang, Lei},
  booktitle={ECCV},
  year={2020}
}

@inproceedings{wu2018group,
  title={Group normalization},
  author={Wu, Yuxin and He, Kaiming},
  booktitle={ECCV},
  year={2018},
  }

@article{frechet1948elements,
  title={Les {\'e}l{\'e}ments al{\'e}atoires de nature quelconque dans un espace distanci{\'e}},
  author={Fr{\'e}chet, Maurice},
  journal={Annales de l'Institut Henri Poincar{\'e}},
  volume={10},
  number={4},
  pages={215--310},
  year={1948}
}

@inproceedings{shahroudy2016ntu,
  title={{NTU RGB+ D}: A large scale dataset for {3D} human activity analysis},
  author={Shahroudy, Amir and Liu, Jun and Ng, Tian-Tsong and Wang, Gang},
  booktitle={CVPR},
  year={2016}
}

@inproceedings{chakraborty2018statistical,
  title={A Statistical Recurrent Model on the Manifold of Symmetric Positive Definite Matrices},
  author={Chakraborty, Rudrasis and Yang, Chun-Hao and Zhen, Xingjian and Banerjee, Monami and Archer, Derek and Vaillancourt, David and Singh, Vikas and Vemuri, Baba},
  booktitle={NeurIPS},
    year={2018}
}

@inproceedings{kobler2022spd,
  title={{SPD} domain-specific batch normalization to crack interpretable unsupervised domain adaptation in {EEG}},
  author={Kobler, Reinmar and Hirayama, Jun-ichiro and Zhao, Qibin and Kawanabe, Motoaki},
  booktitle={NeurIPS},
    year={2022}
}

@article{archakov2021new,
  title={A new parametrization of correlation matrices},
  author={Archakov, Ilya and Hansen, Peter Reinhard},
  journal={Econometrica},
  volume={89},
  number={4},
  pages={1699--1715},
  year={2021},
  publisher={Wiley Online Library}
}

@article{david2019riemannian,
  title={A {Riemannian} structure for correlation matrices},
  author={David, Paul and Gu, Weiqing},
  journal={Operators and Matrices},
  volume={13},
  number={3},
  pages={607--627},
  year={2019}
}

@article{thanwerdas2024permutation,
  title={Permutation-invariant {log-Euclidean} geometries on full-rank correlation matrices},
  author={Thanwerdas, Yann},
  journal={SIMAX},
  year={2024}
}

@article{hartley2013rotation,
  title={Rotation averaging},
  author={Hartley, Richard and Trumpf, Jochen and Dai, Yuchao and Li, Hongdong},
  journal={IJCV},
  year={2013}
}

@article{zacur2014left,
  title={Left-invariant {Riemannian} geodesics on spatial transformation groups},
  author={Zacur, Ernesto and Bossa, Matias and Olmos, Salvador},
  journal={SIMAX},
  year={2014},
}

@article{afsari2011riemannian,
  title={{Riemannian} {Lp} center of mass: Existence, uniqueness, and convexity},
  author={Afsari, Bijan},
  journal={Proceedings of the American Mathematical Society},
  volume={139},
  number={2},
  pages={655--673},
  year={2011}
}

@article{boumal2011discrete,
  title={A discrete regression method on manifolds and its application to data on SO (n)},
  author={Boumal, Nicolas and Absil, P-A},
  journal={IFAC Proceedings Volumes},
  volume={44},
  number={1},
  pages={2284--2289},
  year={2011},
  publisher={Elsevier}
}

@article{ulyanov2016instance,
  title={Instance normalization: the missing ingredient for fast stylization},
  author={Ulyanov, Dmitry and Vedaldi, Andrea and Lempitsky, Victor},
  journal={arXiv preprint arXiv:1607.08022},
  year={2016},
  }

@article{ba2016layer,
  title={Layer normalization},
  author={Ba, Jimmy Lei and Kiros, Jamie Ryan and Hinton, Geoffrey E},
  journal={arXiv preprint arXiv:1607.06450},
  year={2016},
  }

@article{thanwerdas2022theoretically,
  title={Theoretically and computationally convenient geometries on full-rank correlation matrices},
  author={Thanwerdas, Yann and Pennec, Xavier},
  journal={SIMAX},
  year={2022}
}

@article{thanwerdas2022geometry,
  title={The geometry of mixed-{Euclidean} metrics on symmetric positive definite matrices},
  author={Thanwerdas, Yann and Pennec, Xavier},
  journal={Differential Geometry and its Applications},
  volume={81},
  pages={101867},
  year={2022},
    publisher={Elsevier}
}

@article{wang2022symnet,
  title={{SymNet}: A simple symmetric positive definite manifold deep learning method for image set classification},
  author={Wang, Rui and Wu, Xiao-Jun and Kittler, Josef},
  journal={IEEE TNNLS},
  year={2022}
}

@article{yair2019parallel,
  title={Parallel transport on the cone manifold of {SPD} matrices for domain adaptation},
  author={Yair, Or and Ben-Chen, Mirela and Talmon, Ronen},
  journal={IEEE TIP},
  year={2019}
}

@article{chakraborty2022manifoldnet,
    title={{ManifoldNet}: A deep neural network for manifold-valued data with applications},
    author={Chakraborty, Rudrasis and Bouza, Jose and Manton, Jonathan H and Vemuri, Baba C},
    journal={IEEE TPAMI},
    year={2022}
}

@inproceedings{lopez2021vector,
  title={Vector-valued distance and {Gyrocalculus} on the space of symmetric positive definite matrices},
  author={L{\'o}pez, Federico and Pozzetti, Beatrice and Trettel, Steve and Strube, Michael and Wienhard, Anna},
  booktitle={NeurIPS},
    year={2021}
}

@inproceedings{kobler2022controlling,
  title={Controlling the {Fr{\'e}chet} variance improves batch normalization on the symmetric positive definite manifold},
  author={Kobler, Reinmar J and Hirayama, Jun-ichiro and Kawanabe, Motoaki},
  booktitle={ICASSP},
  year={2022},
  }

@inproceedings{vaswani2017attention,
  title={Attention Is All You Need},
  author={Vaswani, Ashish and Shazeer, Noam and Parmar, Niki and Uszkoreit, Jakob and Jones, Llion and Gomez, Aidan N and Kaiser, {\L}ukasz and Polosukhin, Illia},
  booktitle ={NeurIPS},
    year={2017}
}

@inproceedings{krizhevsky2012imagenet,
  title={Imagenet Classification with Deep Convolutional Neural Networks},
  author={Krizhevsky, Alex and Sutskever, Ilya and Hinton, Geoffrey E},
  booktitle={NeurIPS},
  year={2012}
}

@inproceedings{paszke2019pytorch,
  title={{PyTorch}: An imperative style, high-performance deep learning library},
  author={Paszke, Adam and Gross, Sam and Massa, Francisco and Lerer, Adam and Bradbury, James and Chanan, Gregory and Killeen, Trevor and Lin, Zeming and Gimelshein, Natalia and Antiga, Luca and Desmaison, Alban and Kopf, Andreas and Yang, Edward and DeVito, Zachary and Raison, Martin and Tejani, Alykhan and Chilamkurthy, Sasank and Steiner, Benoit and Fang, Lu and Bai, Junjie and Chintala, Soumith},
  booktitle={NeurIPS},
  year={2019}
  }

@inproceedings{brooks2019riemannian,
  title={Riemannian Batch Normalization for {SPD} Neural Networks},
  author={Brooks, Daniel and Schwander, Olivier and Barbaresco, Fr{\'e}d{\'e}ric and Schneider, Jean-Yves and Cord, Matthieu},
  booktitle={NeurIPS},
    year={2019}
}

@inproceedings{lou2020differentiating,
  title={Differentiating through the {Fr{\'e}chet} mean},
  author={Lou, Aaron and Katsman, Isay and Jiang, Qingxuan and Belongie, Serge and Lim, Ser-Nam and De Sa, Christopher},
  booktitle={ICML},
  year={2020}
  }

@article{de2025wrapped,
  title={Wrapped Gaussian on the manifold of Symmetric Positive Definite Matrices},
  author={de Surrel, Thibault and Lotte, Fabien and Chevallier, Sylvain and Yger, Florian},
  journal={arXiv preprint arXiv:2502.01512},
  year={2025}
}

@article{surrel2025geometryaware,
title={Geometry-Aware visualization of high dimensional Symmetric Positive Definite matrices},
author={Thibault de Surrel and Sylvain Chevallier and Fabien Lotte and Florian Yger},
journal={TMLR},
year={2025},

}

@article{groisser2004newton,
  title={Newton's method, zeroes of vector fields, and the Riemannian center of mass},
  author={Groisser, David},
  journal={Advances in Applied Mathematics},
  volume={33},
  number={1},
  pages={95--135},
  year={2004},
  publisher={Elsevier}
}

@article{kendall1990probability,
  title={Probability, convexity, and harmonic maps with small image I: uniqueness and fine existence},
  author={Kendall, Wilfrid S},
  journal={Proceedings of the London Mathematical Society},
  volume={3},
  number={2},
  pages={371--406},
  year={1990},
  publisher={Wiley Online Library}
}

@article{chen2023distribution,
  title={Distribution Knowledge Embedding for Graph Pooling},
  author={Chen, Kaixuan and Song, Jie and Liu, Shunyu and Yu, Na and Feng, Zunlei and Han, Gengshi and Song, Mingli},
  journal={IEEE TKDE},
  year={2023},
  }

@article{pennec1998uniform,
  title={Uniform distribution, distance and expectation problems for geometric features processing},
  author={Pennec, Xavier and Ayache, Nicholas},
  journal={Journal of Mathematical Imaging and Vision},
  volume={9},
  pages={49--67},
  year={1998},
  publisher={Springer},
  }

@article{chakraborty2019statistics,
  title={Statistics on the {Stiefel} manifold: theory and applications},
  author={Chakraborty, Rudrasis and Vemuri, Baba C},
  journal={The Annals of Statistics},
  volume={47},
  number={1},
  pages={415--438},
  year={2019},
  }

@article{wang2006error,
  title={Error propagation on the Euclidean group with applications to manipulator kinematics},
  author={Wang, Yunfeng and Chirikjian, Gregory S},
  journal={IEEE Transactions on Robotics},
  volume={22},
  number={4},
  pages={591--602},
  year={2006},
  publisher={IEEE}
}

@inproceedings{lezcano2019trivializations,
  title={Trivializations for gradient-based optimization on manifolds},
  author={Lezcano Casado, Mario},
  booktitle={NeurIPS},
  year={2019}
}

@article{said2017riemannian,
  title={Riemannian {Gaussian} distributions on the space of symmetric positive definite matrices},
  author={Said, Salem and Bombrun, Lionel and Berthoumieu, Yannick and Manton, Jonathan H},
  journal={IEEE TIT},
  year={2017}
  }

@article{gramfort_meg_2013,
    title = {{MEG} and {EEG} data analysis with {MNE}-{Python}},
    volume = {7},
    journal = {Frontiers in Neuroscience},
    author = {Gramfort, Alexandre},
        year = {2013}
}

@article{jayaram_moabb_2018,
    title = {{MOABB}: trustworthy algorithm benchmarking for {BCIs}},
    volume = {15},
    shorttitle = {{MOABB}},
        number = {6},
    journal = {Journal of Neural Engineering},
    author = {Jayaram, Vinay and Barachant, Alexandre},
    year = {2018},
    pages = {066011}
}

@article{pennec2006riemannian,
  title={A {R}iemannian Framework for Tensor Computing},
  author={Pennec, Xavier and Fillard, Pierre and Ayache, Nicholas},
  journal={IJCV},
  year={2006},
}

@article{lin2019riemannian,
  title={Riemannian Geometry of Symmetric Positive Definite Matrices via {C}holesky Decomposition},
  author={Lin, Zhenhua},
  journal={SIMAX},
  year={2019},
}

@article{hochreiter1997long,
  title={Long Short-Term Memory},
  author={Hochreiter, Sepp and Schmidhuber, J{\"u}rgen},
  journal={Neural Computation},
  volume={9},
  number={8},
  pages={1735--1780},
  year={1997},
  publisher={MIT Press}
}

@article{wang2020deep,
  title={Deep {CNN}s Meet Global Covariance Pooling: better Representation and Generalization},
  author={Wang, Qilong and Xie, Jiangtao and Zuo, Wangmeng and Zhang, Lei and Li, Peihua},
  journal={IEEE TPAMI},
  year={2020},
}

@article{murray2016differentiation,
  title={Differentiation of the {Cholesky} decomposition},
  author={Murray, Iain},
  journal={arXiv preprint arXiv:1602.07527},
  year={2016},
  }

@article{thanwerdas2023n,
  title={O (n)-invariant {Riemannian} metrics on {SPD} matrices},
  author={Thanwerdas, Yann and Pennec, Xavier},
  journal={Linear Algebra and its Applications},
  volume={661},
  pages={163--201},
  year={2023},
    publisher={Elsevier}
}

@article{bronstein2017geometric,
  title={Geometric deep learning: going beyond {Euclidean} data},
  author={Bronstein, Michael M and Bruna, Joan and LeCun, Yann and Szlam, Arthur and Vandergheynst, Pierre},
  journal={IEEE Signal Processing Magazine},
  volume={34},
  number={4},
  pages={18--42},
  year={2017},
    publisher={IEEE}
}

@article{chakraborty2020manifoldnorm,
  title={{ManifoldNorm}: Extending normalizations on {R}iemannian manifolds},
  author={Chakraborty, Rudrasis},
  journal={arXiv preprint arXiv:2003.13869},
  year={2020},
  }

@article{karcher1977riemannian,
  title={Riemannian center of mass and mollifier smoothing},
  author={Karcher, Hermann},
  journal={Communications on Pure and Applied Mathematics},
  volume={30},
  number={5},
  pages={509--541},
  year={1977},
  publisher={Wiley Online Library},
  }

@techreport{muller2007documentation,
  author={M{\"u}ller, Meinard and R{\"o}der, Tido and Clausen, Michael and Eberhardt, Bernhard and Kr{\"u}ger, Bj{\"o}rn and Weber, Andreas},
  year={2007},
  title={Documentation Mocap Database {HDM}05},
  type = {Technical Report},
    institution = {Universit{\"a}t Bonn}
}

@techreport{arsigny2005fast,
  author={Arsigny, Vincent and Fillard, Pierre and Pennec, Xavier and Ayache, Nicholas},
  year={2005},
  title={Fast and simple computations on tensors with log-{Euclidean} metrics},
  institution={INRIA Sophia Antipolis},
  number={RR-5584},
  type={Research Report}
}

@book{berger2003panoramic,
  title={A panoramic view of {Riemannian} geometry},
  author={Berger, Marcel},
  year={2003},
  publisher={Springer},
  }

@book{do1992riemannian,
  title={Riemannian Geometry},
  author={do Carmo, Manfredo P.},
  translator={Flaherty, Francis},
  series={Mathematics: Theory \& Applications},
  year={1992},
  publisher={Birkh{\"a}user}
}

@book{loring2011introduction,
  title={An Introduction to Manifolds},
  author={Tu, Loring W.},
  year={2011},
  publisher={Springer}
}

@book{bhatia2009positive,
  title={Positive Definite Matrices},
  author={Bhatia, Rajendra},
  year={2009},
    publisher={Princeton University Press}
}

@book{bhatia2013matrix,
  title={Matrix analysis},
  author={Bhatia, Rajendra},
  volume={169},
  year={2013},
  publisher={Springer Science \& Business Media}
}

@book{lee2018introduction,
  title={Introduction to {Riemannian} manifolds},
  author={Lee, John M},
  series={Graduate Texts in Mathematics},
  volume={176},
  edition={2nd},
  year={2018},
  publisher={Springer}
}

@book{hall2015lie,
  title={{Lie} Groups, {Lie} Algebras, and Representations: An Elementary Introduction},
  author={Hall, Brian C.},
  series={Graduate Texts in Mathematics},
  volume={222},
  edition={2nd},
  year={2015},
  publisher={Springer Cham},
  doi={10.1007/978-3-319-13467-3}
}

@misc{hinss_eegdata_2021,
    title = {An {EEG} dataset for cross-session mental workload estimation: {Passive} {BCI} competition of the {Neuroergonomics} {Conference} 2021},
    publisher = {Zenodo},
    author = {Hinss, Marcel F. and Darmet, Ludovic and Somon, Bertille and Jahanpour, Emilie and Lotte, Fabien and Ladouce, Simon and Roy, Raphaëlle N.},
    year = {2021},
    doi = {10.5281/ZENODO.5055046},
    }

@techreport{pennec2004probabilities,
  title={Probabilities and statistics on {Riemannian} manifolds: A geometric approach},
  author={Pennec, Xavier},
  institution={INRIA},
  number={RR-5093},
  type={Research Report},
  month={January},
  year={2004},
  }

@phdthesis{thanwerdas2022riemannian,
  title={Riemannian and stratified geometries on covariance and correlation matrices},
  author={Thanwerdas, Yann},
  year={2022},
  school={Universit{\'e} C{\^o}te d'Azur}
}

@book{chavel1995riemannian,
  title={Riemannian Geometry: A Modern Introduction},
  author={Chavel, Isaac},
  number={108},
  series={Cambridge Tracts in Mathematics},
  year={1995},
  publisher={Cambridge University Press}
}

\end{document}